\documentclass[journal]{IEEEtran}
\usepackage{amsmath,amsfonts}
\usepackage{algorithmic}
\usepackage{algorithm}
\usepackage{array}
\usepackage{color}
\usepackage[caption=false,font=normalsize,labelfont=sf,textfont=sf]{subfig}
\usepackage{textcomp}
\usepackage{stfloats}
\usepackage{url}
\usepackage{verbatim}
\usepackage{graphicx}
\usepackage{cite}
\usepackage{amsthm,amssymb}
\usepackage{mathrsfs}
\usepackage{float}
\usepackage{subfig}
\usepackage{booktabs}
\usepackage{multirow}
\usepackage{bm}
\newtheorem{assumption}{Assumption}

\newtheorem{lemma}{Lemma}

\newtheorem{theorem}{Theorem}
\newtheorem{definition}{Definition}
\newtheorem{remark}{Remark}
\newtheorem{corollary}{Corollary}
\begin{document}

\title{Multi-agent Policy Reciprocity with \\ Theoretical Guarantee}

% \author{Michael~Shell,~\IEEEmembership{Member,~IEEE,}
%         John~Doe,~\IEEEmembership{Fellow,~OSA,}
%         and~Jane~Doe,~\IEEEmembership{Life~Fellow,~IEEE}% <-this % stops a space
% \thanks{M. Shell was with the Department
% of Electrical and Computer Engineering, Georgia Institute of Technology, Atlanta,
% GA, 30332 USA e-mail: (see http://www.michaelshell.org/contact.html).}% <-this % stops a space
% \thanks{J. Doe and J. Doe are with Anonymous University.}% <-this % stops a space
% \thanks{Manuscript received April 19, 2005; revised August 26, 2015.}}

\author{Haozhi Wang, Yinchuan Li,~\IEEEmembership{Member,~IEEE}, Qing Wang,~\IEEEmembership{Member,~IEEE},Yunfeng Shao, Jianye Hao 
    %<-this % stops a space
\thanks{This work is supported by the National Natural Science Foundation of China (Grant No. 61871282 and U20A20162), the Key Research and Development Program of Tibet Autonomous Region, and the Science and Technology Major Project of Tibetan Autonomous Region of China (Grant No. XZ202201ZD0006G03).
(\emph{Haozhi Wang
and Yinchuan Li contributed equally to this work.}) 
 (\emph{Corresponding author: Qing Wang (email: wangq@tju.edu.cn)).}}
\thanks{Haozhi Wang and Qing Wang are with School of Electrical and Information Engineering, Tianjin University, Tianjin, China. Jianye Hao is with the College of Intelligence and Computing, Tianjin University, Tianjin, China.
This work was completed while Haozhi Wang was an intern of the Huawei Noah’s Ark Lab for Advanced Study.
}
\thanks{Yinchuan Li and Yunfeng Shao are with Huawei Noah's Ark Lab, Beijing, China.}
        % <-this % stops a space
}

% This work was completed while Haozhi Wang was an intern of the Huawei Noah's Ark Lab.
% (e-mail: wanghaozhi@tju.edu.cn, wangq@tju.edu.cn, jianye.hao@tju.edu.cn)
% (e-mail: shaoyunfeng@huawei.com, lidong106@huawei.com, liyinchuan@huawei.com)

% \markboth{Journal of \LaTeX\ Class Files,~Vol.~14, No.~8, August~2015}%
% {Shell \MakeLowercase{\textit{et al.}}: Bare Demo of IEEEtran.cls for IEEE Journals}

\maketitle

% \begin{abstract}
% The abstract goes here.
% \end{abstract}

% % Note that keywords are not normally used for peerreview papers.
% \begin{IEEEkeywords}
% IEEE, IEEEtran, journal, \LaTeX, paper, template.
% \end{IEEEkeywords}

\begin{abstract}
Modern multi-agent reinforcement learning (RL) algorithms hold great potential for solving a variety of real-world problems.
However, they do not fully exploit cross-agent knowledge to reduce sample complexity and improve performance. Although transfer RL supports knowledge sharing, it is hyperparameter sensitive and complex.
To solve this problem, we propose a novel multi-agent policy reciprocity (PR) framework, where each agent can fully exploit cross-agent policies even in mismatched states.
We then define an adjacency space for mismatched states and design a plug-and-play module for value iteration, which enables agents to infer more precise returns. To improve the scalability of PR, deep PR is proposed for continuous control tasks. Moreover, theoretical analysis shows that agents can asymptotically reach consensus through individual perceived rewards and converge to an optimal value function, which implies the stability and effectiveness of PR, respectively. Experimental results on discrete and continuous environments demonstrate that PR outperforms various existing RL and transfer RL methods.
\end{abstract}


% \fi
\IEEEpeerreviewmaketitle

\section{Introduction}
\label{submission}
Multi-agent reinforcement learning (MARL) have demonstrated extraordinary capabilities in solving practical applications, such as the coordination of robotic swarms \cite{huttenrauch2017guided} and autonomous vehicles \cite{cao2012overview}.
The most popular method is the centralized training and decentralized execution (CTDE) paradigm \cite{rashid2018qmix, wang2020qplex,yu2021surprising}, which avoids both the combinatorial complexity of centralized training \cite{yang2019alpha} and the non-stationary problem of independent learning \cite{sunehag2017value}. 
Unfortunately, these algorithms pay more attention to the credit assignment among agents, they do not utilize other agents' policies to improve performance further and reduce sample complexity, which limits practical applications \cite{wang2021sample, christianos2021scaling, zhang2020model}. 
Another approach is to introduce differences based on the sharing of policy parameters during decentralized execution, thereby enhancing multi-agent exploration capabilities and learning complex cooperative policies \cite{wang2020roma, chenghao2021celebrating}.
Instead, our approach explores how to learn from other agents in terms of value functions.

Transfer learning (TL) based RL algorithms enable agents to quickly adapt to new tasks or transfer between agents to improve performance through knowledge sharing.
Most of the current policy transfer works are mainly applied to the single-agent setting, performing policy transfer based on the similarity of tasks \cite{taylor2007transfer, brys2015policy,song2016measuring}, or leveraging other agent's policy for a compelling exploration of new tasks \cite{li2018optimal, li2018context}. 
% These algorithms cannot be directly applied to the multi-agent system since the non-stationary environment. 
In the MARL setting, existing works transfer knowledge via policy distillation \cite{wadhwania2019policy, xue2020transfer} or adaptively exploit a suitable policy based on the option learning \cite{yang2021efficient}. 
However, these methods are usually complex to model and sensitive to hyperparameters.

In this paper, we propose a novel policy reciprocity (PR) framework in the collaborative setting \cite{panait2005cooperative} to solve the above problems, where each agent can share knowledge with other agents while learning to achieve mutually beneficial symbiosis. 
In particular, we define an adjacency space for mismatched states, then each agent performs modified value iteration by exploiting these adjacency states. Under this setting, agents can learn more information from the policies of other agents even with observations of different dimensions, enhancing performance and accelerating the learning process. Moreover, theoretical analysis shows that multiple agents can achieve consensus with each other and gradually converge to the optimal value function by trading off the reciprocity potential, which provides guidance for empirical analysis. Finally, we extend PR with deep neural networks that can improve the performance of various existing MARL and transfer RL algorithms.

\subsection{Main Contributions}
First, we propose a novel policy reciprocity framework to reduce the sample complexity of existing multi-agent RL algorithms, enabling policies to interact under different dimension observations. In particular, we give the definition of adjacent states among agents and propose a protocol that each agent use the value function of these adjacent states to update its policy. This framework allows RL to exploit more information in each iteration,  speeding up the learning process and improving accuracy.

Second, we theoretically analyze the multi-agent policy reciprocity framework in the tabular setting. We found that the agents can achieve consensus gradually even though their individual rewards may be inconsistent, which ensures the stability of the learning process of the agents.
In addition, we also prove that this iteration method can converge to the optimal state-action value function after enough iterations, which validates the effectiveness of our algorithm.

Third, we conduct experiments on multiple discrete environments to validate the consensus theorem and the convergence of our policy reciprocity approach. 
Moreover, we extend this iteration process to a plug-and-play deep PR module with deep neural networks to improve scalability. Extensive continuous experiments show that our deep PR can improve the performance of many existing MARL and transfer RL algorithms, such as MADDPG \cite{lowe2017multi}, QMIX \cite{rashid2018qmix}, MAPPO \cite{yu2021surprising} and MAPTF \cite{yang2021efficient}.
\section{Related Works}
Multi-agent policy reciprocity is closely related to the following works.
\subsection{Cooperative Multi-agent Reinforcement Learning}
A whole suite of MARL algorithms has been developed to solve collaborative tasks, two extreme approaches are independent learning \cite{tan1993multi} and centralized training.
The independent training method treats the effect of other agents as part of the environments, resulting in agents facing non-stationary environments and spurious rewards \cite{sunehag2017value, d2019sharing}.
Centralized training treats the multi-agent problem as a single-agent counterpart. Unfortunately, this method exhibits combinatorial complexity and is difficult to scale beyond dozens of agents \cite{yang2019alpha}. 
Therefore, the most popular paradigm is centralized training and decentralized execution (CTDE), including value-based \cite{sunehag2017value, rashid2018qmix, son2019qtran, wang2020qplex} and policy-based \cite{lowe2017multi, yu2021surprising, kuba2021trust} methods. 
The goal of value-based methods is to decompose the joint value function among agents for decentralized execution, which requires that local maxima on the value function per every agent should amount to the global maximum on the joint value function.
VDN \cite{sunehag2017value} and QMIX \cite{rashid2018qmix} propose two classic and efficient factorization structures, additivity and monotonicity, respectively, despite the strict factorization method.
The policy-based methods extend the single-agent TRPO \cite{schulman2015trust} and PPO \cite{schulman2017proximal} into the multi-agent setting, such as MAPPO \cite{yu2021surprising}, which has shown the surprising effectiveness in cooperative, multi-agent games.
However, these methods do not fully exploit the existing knowledge of other agents to accelerate convergence further and improve performance.

\subsection{Multi-agent Transfer Learning} 
The transfer learning on single-agent has been widely studied. 
A major approach to transfer learning in RL focuses on measuring the similarity between two tasks through the state space \cite{taylor2007transfer, brys2015policy} or computing the similarity of two Markov Decision Processes (MDP) \cite{song2016measuring}. 
Then the agent can directly transfer the source policy to the target task according to the similarity to accelerate the convergence.Another direction of policy transfer methods is to select an appropriate source policy based on the performance of the source policy on the target task and guide the agent to explore during the learning process \cite{li2018optimal, li2018context}. 
CAPS is proposed in \cite{li2018context}, which requires the optimality of source policies since it only learns an intra-option policy over these source policies.
\cite{yang2020efficient} selects a suitable source policy during target task learning and use it as a complementary optimization objective of the target policy, which can avoid manually adding source policies.
Some works apply the idea of TL to the multi-agent setting \cite{omidshafiei2019learning, kim2019learning, wadhwania2019policy, xue2020transfer, yang2021efficient}.
The authors propose a teacher-student framework that enables each agent to learn when to transfer other agents' policies to other agents. However, this approach is limited to two-agent scenarios in \cite{omidshafiei2019learning, kim2019learning}.
DVM \cite{wadhwania2019policy} and LTCR \cite{xue2020transfer} transfer knowledge among multiple agents through policy distillation, but serve in a coarse-grained manner since the two-stage training. MAPTF \cite{yang2021efficient} models the policy transfer among multiple agents as the option learning problem, which can adaptively select a suitable policy for each agent to exploit. 
However, these methods usually require complex modeling and are sensitive to hyperparameters, and most works lack theoretical guarantees.
% \begin{figure*}[!htbp]
% 	\centering
% 	\subfloat{
% 	\includegraphics[width=7in]{fig/system/pMeta-RL2.pdf}
% 	}
% 	\caption{(a) Several distinct tasks on MuJoCo. The tasks vary in either the transition function (e.g., robots with different dynamics) or the reward function (e.g., different goal location).
% 	(b) Visualization of 2D Sparse-Point-Robot. 
% 	The blue circles represent the different tasks.
% 	We highlight two distinct goals in dark blue and visualize the policies leading the agent to reach them.
% 	$x$ and $y$ axes of the heatmap represent the two dimensions of action, while $z$ axis represents the state-action value on the initial state. Red represents higher values.
% 	For goals 1 and 2, the personalized policies prefer left and right actions, respectively. In contrast, traditional gradient aggregation-based meta-policy have similar propensities for left and right actions.
% 	(c) Performance comparison of our algorithm and PEARL at 50000 training steps. 
% 	The agent using the personalized policy can navigate to more right goals.}
% 	\label{fig_system}
% \end{figure*}

\section{Multi-agent Policy Reciprocity}
\subsection{Background}
A fully cooperative multi-agent task usually can be formalized as a multi-agent MDP, which is characterized by a tuple $\mathcal{G}=\left \langle \mathcal{N}, \mathcal{S}, \{\mathcal{A}_i\}_{i\in \mathcal{N}}, \mathcal{P}, \{\mathcal{R}_i\}_{i\in \mathcal{N}}, \gamma \right \rangle$, where $\mathcal{N}:=\left\{1,2,\cdots,N\right\}$ is a finite set of agents, $s_{\text{G}} \in \mathcal{S}$ is a finite set of global states, $\mathcal{A}_i$ is the action space of agent $i$, $\mathcal{A}=\prod_{i=1}^{N} \mathcal{A}_i$ is the joint action space, 
and $r_i \in \mathcal{R}_i: \mathcal{S} \times \mathcal{A} \rightarrow \mathbb{R}$ is the reward function. 
At each discrete time step, each agent makes action decision $a_i \in \mathcal{A}_i$ based on its state $s_i \in \mathcal{S}_i$ and its individual policy $\pi_i(a|s_i)$. After performing a joint action $\boldsymbol{a} \in \mathcal{A}$, the environment $\mathcal{E}$ produces a reward and transitions the agents to a new global state. 
For the partially observable setting, each agent may receive an individual
partial observation $s_i \in \mathcal{S}_i$, which constitutes the global state space $\mathcal{S}:=\times_{i \in \mathcal{N}} \mathcal{S}_{i}$
The general objective function is to find a joint policy $\boldsymbol{\pi} = \langle \pi_1, \cdots, \pi_n\rangle$ by maximizing a joint state value function $V^{\boldsymbol{\pi}}(s_{\text{G}}) = \mathbb{E}[\sum_{t=0}^{\infty}\gamma^t r^t|s_0=s_{\text{G}}, \boldsymbol{\pi}]$, or optionally optimizing the joint state-action value function $Q^{\boldsymbol{\pi}}(s_{\text{G}}, \boldsymbol{a}) = r(s_{\text{G}}, \boldsymbol{a}) + \gamma \mathbb{E}_{s_{\text{G}}^{\prime}\sim\mathcal{P}}[V^{\boldsymbol{\pi}}(s_{\text{G}}^{\prime})]$.

One common method in multi-agent learning is independent $Q$-learning (IQL) \cite{tan1993multi}, which decomposes a multi-agent problem into a collection of simultaneous single-agent problems that share the same environment. 
Under the tabular setting, each agent exploits the $Q$-learning algorithm to find the optimal action value function of each agent via temporal differences:
\begin{equation}\label{eq_q_learning}
\begin{split}
    & Q_i(s_i,a_i) = Q_i(s_i,a_i) \\
    &~~~+ \alpha \Big[ r_i(s_i,a_i) + \gamma \max_{a_i^{\prime}}Q_i(s_i^{\prime},a_i^{\prime}) - Q_i(s_i,a_i) \Big],
\end{split}
\end{equation}
where $\alpha$ is the learning rate.

\begin{definition}[Adjacency States] \label{def1}
For any two agents $i,j \in \mathcal{N}$, 
we say that the two agents have adjacency states if $\exists s_i \in \mathcal{S}_i, s_j \in \mathcal{S}_j$, such that $\| \Pi_{s_i}^T s_i - \Pi_{s_j}^T s_j\|_0 \leq \rho$, 
% where $\|\cdot\|_0$ and $(\cdot)^T$ denote the $\ell_0$-norm and matrix transpose, respectively, and 
where $\rho$ is a small positive constant named adjacency level. The set of agents with adjacency states to agent $i$ is defined as $\mathcal{N}_i$.
\end{definition}

\begin{definition}[Adjacency Space] \label{def2}
Given a state $s$, its adjacency space $\mathcal{S}^{\sharp}$ is defined as $\forall s^{\sharp} \in \mathcal{S}^{\sharp}, \rho_{s^{\sharp}} = \|\Pi^{T}_s s-\Pi_{s^\sharp}^{T} s^{\sharp}\|_0 \leq \rho_{\sharp}$\footnote{This $\ell_0$-norm can be generalized to other similar states metric methods such as Euclidean distance, cosine similarity or Kantorovich distance \cite{song2016measuring}.}, where $\Pi_s$ and $\Pi_{s^{\sharp}}$ are the corresponding observation matrix of $s$ and $s^{\sharp}$, respectively, and $\rho_{\sharp}$ represents the the adjacency level of $\mathcal{S}^{\sharp}$.  
\end{definition}

\subsection{Multi-agent Policy Reciprocity with Tabular $Q$-values}
We consider a decentralized multi-agent system in a tabular setting \cite{kar2013cal}, and introduce observation matrices to account for adjacency conditions for different states.
Assume that each agent observes all or part of the global state $s_{\text{G}} \in \mathbb{R}^d$ with $d$ being the state dimension. 
We define a general state representation $s \in \cup_{i \in \mathcal{N}}\mathcal{S}_i$, including all possible states observed by the agents. 
Define $\Pi_s \in \mathbb{R}^{d_s \times d}$ as the observation matrix related to state $s$, where only one element in each row is $1$ and the other elements are $0$, such that $s = \Pi_s s_{\text{G}} \in \mathbb{R}^{d_s}$ with $d_s$ being the state dimension of $s$.
Then, the adjacency state is described in Definition~\ref{def1}.

\begin{figure}[!h]
\vspace{-0.4cm} 
\setlength{\abovecaptionskip}{0.2cm} 
\setlength{\belowcaptionskip}{-0.1cm} 
	\centering
% 	\hspace{-0.3cm}

	\includegraphics[width=3.5in]{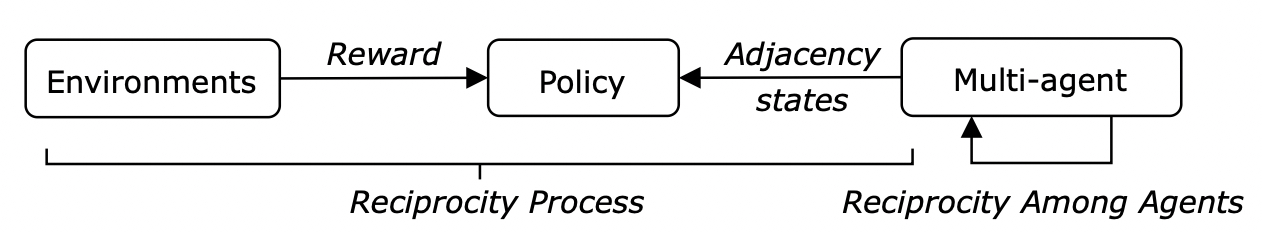}
% system_paper_4.pdf
	\caption{An illustration of multi-agent policy reciprocity.}
	\label{fig_example}
\end{figure}

\begin{remark}
For ease of understanding, we give an example to further illustrate Definition~\ref{def2}.
% Figure~\ref{fig_example} depicts an practical example of these similar states. 
Assume that the global state of the environment is a four-dimensional vector $s_{\text{G}}=[a,b,c,d]^{T}$.
Each agent obtains its observation by the observation matrix.
% \begin{wrapfigure}[5]{r}{0.45\textwidth}
% \vspace{-14pt}
%     \centering
%  \includegraphics[height=0.6in]{system_fig/adjacency2.pdf}
%  \caption{}
%  \label{fig_example}
% \end{wrapfigure}
Consider two states $s$ and $s^{\sharp}$ observed by 
$\Pi_s$ and $\Pi_{s^{\sharp}}$, which can be written as
\begin{equation}
\begin{split}
\left[\begin{array}{l}
a  \\
b  \\
c
\end{array}\right]=\underbrace{\left[\begin{array}{llll}
1 & 0 & 0 & 0 \\
0 & 1 & 0 & 0 \\
0 & 0 & 1 & 0
\end{array}\right]}_{\Pi_s } \cdot \left[\begin{array}{l}
a  \\
b  \\
c  \\
d
\end{array}\right], \\ \left[\begin{array}{l}
a  \\
b  \\
d  
\end{array}\right]=\underbrace{\left[\begin{array}{llll}
1 & 0 & 0 & 0 \\
0 & 1 & 0 & 0 \\
0 & 0 & 0 & 1 
\end{array}\right]}_{\Pi_{s^{\sharp}}} \cdot \left[\begin{array}{l}
a  \\
b  \\
c  \\
d
\end{array}\right],
\end{split}
\end{equation}
therefore, if an agent observes state $s=\Pi_s s_{\text{G}} = [a,b,c]^{T}$, then $s^{\sharp}=[a,b,d]^{T}$ can be seen as an adjacency state of $s$ since $\|\Pi_{s}^{T}s-\Pi_{s^{\sharp}}^{T}s^{\sharp}\|_0=2$. These adjacency states to form an adjacency space based on the adjacency level $\rho_{\sharp}$.
\end{remark}

Figure~\ref{fig_example} shows that PR can exploit the value function of adjacency space performing policy improvement.
% All these adjacency states can form an adjacency space.
In this case, each agent can transform its own $Q$-table for policy reciprocity based on the adjacency space through unified coordination, thereby achieving the improvements of policy performance and reducing the sample complexity. Next, we show the multi-agent policy reciprocity framework, where each agent can
interacts with other agents for policy learning together. 
In this setting, every agent eventually learns its optimal value function $Q^{*}_i(s,a)$ based on the stochastic processes $\{s^t\}, \{a^t\}, \{r_i^t\}$ and the value function $\{Q_\ell^{t'}\}_{\ell \in \mathcal{N}_i, t' \leq t}$ on $\mathcal{S}^{\sharp}$ from other agents. 
To do this, we first give the measurability of the aforementioned stochastic process, which requires characterizing locally accessible agent information over time for decision-making.

\begin{lemma}[Measurability and Moments]\label{def3}
Let $(\Omega, \mathcal{F}, \mathcal{P})$ be a complete probability space with filtration $\left\{\mathcal{F}_t\right\}$, which induced by the following the processes
\begin{equation}
    \mathcal{F}_t = \sigma\big(\{s^{t'},a^{t'}\}_{t'\leq t},\{r_i(s^{t'},a^{t'})\}_{t' \leq t},
    \{Q_{\ell}^{t'}\}_{\ell \in \mathcal{N}_i, t' \leq t} \big)
\end{equation}
where $\sigma(\mathcal{J})$ is the smallest $\sigma$-algebra for the random variables collection $\mathcal{J}$. 
The second term in $\mathcal{F}_t$ records the sensed reward at all time, while the latter involves all the information of other agents. 
% Each agent can update its policy by this information and locally sensed reward.
The state and action processes $\left\{s^t\right\}$ and $\left\{a^t\right\}$, are adapted to $\mathcal{F}_t$. 
The transition probability and reward of a specific state $s^t=s$ satisfy that $\mathcal{P}(s^{t+1}=s^{\prime}|\mathcal{F}_t) = p_{s,s^{\prime}}^a$ and   $\mathbb{E}[r_i(s,a)|\mathcal{F}_t] = \mathbb{E}[r_i(s,a)|s,a]$, respectively. For the adjacency space of $\mathcal{S}^{\sharp}$ of state $s$, we have 
$$\mathbb{E}[Q_i(s^{\sharp},a)|\mathcal{F}_t] = \mathbb{E}[Q_i(s^{\sharp},a)|s^{\sharp},a], \forall s^{\sharp} \in \mathcal{S}^{\sharp}.$$
\end{lemma}
Lemma~\ref{def3} states that for any $s^{\sharp}$ in the adjacency space of state $s$, its value function $Q_i(s^{\sharp},a)$ is only conditioned on the current state-action pair and irrelevant to its history.
Based on Lemma~\ref{def3}, we can make fully advantage of the same and similar state for policy reciprocity.
\begin{definition}[Policy Reciprocity]
For any greedy policy, i.e., $\pi(s) = \arg\max_a Q(s,a)$, the policy reciprocity enables agent to
estimate $Q(s,a)$ by the sum-average state value function $Q_{\sharp}(s,a)$ in the adjacency space of state $s$, which can be written as 
\begin{equation}\label{eq_average}
    Q_{\sharp}(s,a) = \sum_{s^{\sharp} \in \mathcal{S}^{\sharp}}\frac{\exp(d-\rho_{s^{\sharp}})}{Z(\rho_{s^{\sharp}})} Q(s^{\sharp},a),
\end{equation}
where $Z(\rho_{s^{\sharp}})$ is a partition function calculated by $Z(\rho_{s^{\sharp}}) = \sum_{s^{\sharp} \in \mathcal{S}^{\sharp}}\exp(d-\rho_{s^{\sharp}})$.
\end{definition}

To utilize policy reciprocity, each agent updates its $Q$-values based on its exploration and interacts with other agents, i.e., the $Q$-values of each agent is updated as
\begin{equation}\label{e2}
\begin{split}
    Q_i^{t+1}(s,a) =&~ [1-\alpha_t(s,a)]Q_i^{t}(s,a) +\\ 
    & \underbrace{\alpha_t(s,a)\big[ r_i(s,a) + \gamma \max_{a^{\prime} \in \mathcal{A}_i} Q_i^{t}(s^{\prime},a^{\prime})\big]}_{\emph{\text{local~autonomy}}} \\
    &~~~~ + \underbrace{\beta_t(s,a) (Q_{\star,i}^t(s,a)-Q_i^{t}(s,a))}_{\emph{\text{reciprocity~potential}}},
\end{split}
\end{equation}
where $\beta_t$ is a weight factor controlling the degree of interaction with other agents and
\begin{equation}\label{e3}
    Q_{\star,i}^t(s,a) = \frac{\kappa}{|\mathcal{N}_i|} \sum_{i \in \mathcal{N}_i} Q_i^t(s,a) + \frac{1-\kappa}{|\mathcal{N}_i|} \sum_{i \in \mathcal{N}_i} Q_{\sharp,i}^t(s,a)
\end{equation}
with $\kappa$ being a weight factor. 
When $\kappa=1$, it means that the agent only uses the $Q$-values with the exact same state-action for policy reciprocity. 
The iterative process in~\eqref{e2} involves \emph{local autonomy} of each agent (update by its reward $r_i$) and \emph{reciprocity potential} with other agents (update by $Q_{\star,i}(s,a)$). By trading off the weight factor, we can theoretically show the convergence process of this algorithm, i.e., each agent can iterate to the optimal policy, while all the agents can asymptotically achieve consensus on the value function and the corresponding optimal stationary policy.
We summarize the tabular multi-agent PR algorithm in Algorithm~\ref{alg_PR} for clarification.

\begin{algorithm}[t]\small\caption{Multi-agent Tabular PR} 
% \hspace*{0.02in} % \hspace*{0.02in}用来控制位置，同时利用 \\ 进行换行
\begin{algorithmic}[1] %[1] enables line numbers
\REQUIRE Training Epoch $T$; weight Factor $\alpha_0, \beta_0, \kappa$
\STATE Initial the value function $Q_{i}^0, \forall i=1,\cdots,N$
\FOR[\emph{parallel do with $N$ agents}]{$t = 1, \cdots, T$}
% \For{$i = 1, \cdots, N$}
\STATE Update $Q_{\star,i}^t$ based on \eqref{e3} or \eqref{eq_global}
\FOR{$e= 1, \cdots, E$}
\STATE Interact with the environment using current $\epsilon$-greedy policy
\STATE Sample a batch of transition sample $(s,a,r,s^{\prime})$
\STATE Update $Q_{i}^{t}(s,a)$ based on \eqref{e2} 
% \EndFor
\ENDFOR
\ENDFOR
\ENSURE
    Tabular value function $Q_i, \forall i=1,\cdots,N$
\end{algorithmic}
\label{alg_PR}
\end{algorithm}

\subsection{Theoretical Analysis}
In this subsection, we provide theoretical analysis for consensus and convergence of the proposed method. Before that, we present some useful assumptions.

\begin{assumption}[Reward boundedness]
For each agent $i$, there exists a constant $\varepsilon_1 >0$ such that $\mathbb{E}\left[r_i^{2+\varepsilon_1}(s,a)\right] < \infty, \forall s, a.$
\label{assumption_mm}
\end{assumption}
\begin{assumption}[Finite stopping times] 
For each state-action pair $(s,a)$ of agent $i$, we introduce a random time sequence $\left\{T_{s,a}(k)\right\}$, for each positive integer $k$, it can be written as
\begin{equation}
    T_{s,a}(k) = \inf\big\{t \geq 0 \big| \sum_{{t'}=0}^t \mathbb{I}(s^{ t'},a^{t'}=s,a)=k\big\},
\end{equation}
where $\mathbb{I}(\cdot)$ denotes the indicator function. $T_{s,a}(k)$ means that the $k$-th sampling instant of the $(s,a)$ pair, and is almost sure finite, i.e., $\mathbb{P}(T_{s,a}(k)<\infty)=1$. 
\label{assumption_stop}
\end{assumption}
\begin{assumption}[Connectivity]
The connection relation of agents can be defined as a Laplacian matrix $L_t$, which is independent of the instantaneous reward $r_i(s,a)$. We assume the second smallest eigenvalue of the average matrix $\bar{L}_t$ is greater than 0, i.e., $\lambda_2(\mathbb{E}\big[\bar{L}_t\big])>0$.
\label{assumption_matrix}
\end{assumption}
\begin{remark}
{\rm{{[Condition (A1)]:}}} Assumption~\ref{assumption_mm} indicates that the reward is bounded, which is reasonable since the reward function can be set artificially.
{\rm{{[Condition (A2)]:}}} 
Assumption~\ref{assumption_stop} presents a requirement that ensures each state-action pair can be simulated or observed infinitely. It is a common assumption in some real-time control approaches for desired convergence guarantee \cite{kar2013cal, kar2013distributed}.
{\rm{{[Condition (A3)]:}}} Assumption~\ref{assumption_matrix} means that all agents are fully connected to each other in average. 
In this setting, each agent can communicate with other agents, i.e., $Q_{\star,i}$ contains the information of all agents.
\end{remark}

Next, we present the following boundedness lemma based on the above assumptions.

\begin{theorem}\label{Lemma_1}
Let Assumptions~\ref{assumption_mm}, \ref{assumption_stop} and \ref{assumption_matrix} hold.
For all the state-action pairs, 
if the weight sequences $\left\{\alpha_t(s,a)\right\}$ and $\{\beta_t(s,a)\}$ satisfy
\begin{equation}\label{eq_weight}
\begin{split}
\beta_t(s, a)&=\left\{\begin{array}{cl}
\frac{b}{(k+1)^{\tau_{2}}} & \text { if } t=T_{s, a}(k) \text { for some } k \geq 0 \\
0 & \text { otherwise }
\end{array}\right. \\
\alpha_t(s, a)&=\left\{\begin{array}{cl}
\frac{a}{(k+1)^{\tau_{1}}} & \text { if } t=T_{s, a}(k) \text { for some } k \geq 0 \\
0 & \text { otherwise }
\end{array}\right.
\end{split},
\end{equation}
where $a$ and $b$ are positive constants, $\tau_1 \in \left(1/2,1\right]$, and $0 < \tau_2 <\tau_1-1/(2+\varepsilon_1)$, then the successive refinement sequence $\left\{Q_i^t\right\}$ obtained based on policy reciprocity in \eqref{e2} satisfies
\begin{equation*}
    \sup_{t \geq 0}\left\|Q_i^t\right\| < \infty, \forall i.
\end{equation*}
\end{theorem}

Lemma~\ref{Lemma_1} shows the boundedness of agents' state-action value under the policy reciprocity. Based on Lemma~\ref{Lemma_1}, we have the following main theorems.

\begin{theorem}\label{Theorem_1_1}
Let Assumptions~\ref{assumption_mm}, \ref{assumption_stop} and \ref{assumption_matrix} hold. 
The weight sequences $\{\beta_t(s,a)\}$ and $\{\alpha_t(s,a)\}$ satisfy the condition in \eqref{eq_weight}, 
then we can conclude that the agents' Q-table can reach consensus with each other as $t \rightarrow \infty$, such that
\begin{equation}\label{theorem_1}
    \mathbb{P}\left(\lim_{t\rightarrow\infty}\left|Q_i^t(s,a)-\bar{Q}^t(s,a)\right|=0\right)=1, \forall i, s, a,
\end{equation}
where $\bar{Q}^t(s,a)=\frac{1}{N}\sum_i Q_i^t(s,a)$. 
\end{theorem}

Theorem~\ref{Theorem_1_1} shows that the agents reach consensus asymptotically with their individual sensed rewards, which illustrates that each agent will make similar decision under same observation after policy reciprocity. 
Usually, the individual rewards are difficult to reflect the effective information of the whole system. 
In the case of fully decentralization, the agent cannot optimize the global return only based on its own reward. 
The consensus potential guarantees the stability of the agent as it learns independently \cite{kar2013cal, zhang2018fully}.

\begin{corollary}\label{Theorem_1_2}
Let Assumption~\ref{assumption_mm}, \ref{assumption_stop} and \ref{assumption_matrix} hold and the weight sequences $\{\beta_t(s,a)\big\}$ and $\{\alpha_t(s,a)\}$ satisfy the condition in \eqref{eq_weight}. Consider a particular case, i.e., $\kappa=1$, which means that $Q_{\star}$ only aggregate the Q-values with same state. 
Let this stochastic process can be represented by $\{\tilde{Q}_{i}^t(s,a)\}$, if $2 \tau_2 \geq \tau_1-1/(2+\varepsilon_1)$, then we have
\begin{equation}\label{theorem_2}
    \mathbb{P}\left(\lim_{t\rightarrow\infty}\left|Q_i^t(s,a)-\tilde{Q}_{i}^t(s,a)\right|=0\right)=1, \forall i, s, a.
\end{equation}
\end{corollary}

\begin{remark}
Corollary~\ref{Theorem_1_2} illustrates that if $\tau_2$ is chosen appropriately, the iterative process of PR can be consensus with the case when $\kappa=1$, which guides how well the PR algorithm utilizes the value function of adjacency states.
% the effective of our multi-agent policy reciprocity algorithm. 
% Although the Q-values under similar value may introduce error for learning process, 
% the agent's model can reach consensus with the theme of Q-value iteration using same states.
\end{remark}

\begin{theorem}\label{Theorem_2}
    Let $Q^{*}(s,a)$ be the optimal $Q$-value defined in \cite{kar2013cal}. If the weight sequences $\{\beta_t(s,a)\}$ and $\{\alpha_t(s,a)\}$ satisfy the condition in \eqref{eq_weight}, for each agent $i$, we have
    \begin{equation}
        \mathbb{P}\left(\lim_{t\rightarrow\infty}Q_i(s,a)=Q^{*}(s,a)\right)=1, \forall i,s,a.
    \end{equation}
\end{theorem}

Theorem~\ref{Theorem_2} illustrates that the iterative process in \eqref{e2} and \eqref{e3} can converge to the unique fixed point of Q-leaning operator. 
This means that the algorithm can iterate asymptotically to the optimal policy, which instructs us to design more efficient algorithms that take advantage of these similar states.

\section{Multi-agent Deep Policy Reciprocity}
% Since the tabular setting is difficult to adapt to large-scale problems, existing methods use function approximators to solve multi-agent training.
In this section, we extend our PR into the deep RL scheme.
Existing model-free RL methods can be divided as value-based, policy-based and actor-critic methods \cite{franccois2018introduction}. 
Although these algorithms are different, they all need to estimate state value function or its variants $Z$, i.e., $Z$ can be the state-action value $Q(s,a)$ or advantage function $A(s,a)=Q(s,a)-V(s)$.
For example, MADDPG and MAPPO use $Q(s,a)$ and $A(s,a)$ to calculate the gradient of policy network, respectively.
Therefore, a deep PR scheme is defined as a scheme that can perform value or advantage function interaction between agents for joint training, as shown in Definition~\ref{def4} below.

% Therefore, the deep PR scheme can perform value or advantage function interaction among agents for joint training, which can be defined as follows.
\begin{definition}[Deep Policy Reciprocity]
For any agent $i$, deep policy reciprocity enables each agent to estimate the value $Z$ based on other agent's information, i.e., 
\begin{equation}\label{eq_global}
    Z_{\star,i}^{w_i}\big(s, a\big) = (1-\kappa)Z_i^{w_i}\big(s, a\big)+\kappa \sum_{j \in \mathcal{N}_i}\sum_{s^{\sharp}\in\mathcal{S}^{\sharp}}Z_j^{w_j}\big(s^{\sharp}, a\big), \forall i,s,a,
\end{equation}
where $Z \in \{Q, V, A\}$ parameterized by $w$,  $\mathcal{N}_i$ is the set of adjacency agents of the $i$-th agent.% and $\kappa$ is a trade-off factor. 
\label{def4}
\end{definition}

% Since the actor-critic based algorithms, such as MADDPG, are a hybrid architecture of value-based and policy-based approaches, we explore the role of PR based on these kind of algorithms, which can be extended to other algorithms straightforwardly. 

Since actor-critic-based algorithms, such as MADDPG, are hybrid architectures of value-based and policy-based approaches, we demonstrate the power of PR based on such more general algorithms, which can be extended to other algorithms straightforwardly, such as MAPPO \cite{yu2021surprising} and TD3 \cite{fujimoto2018addressing}.
More specifically, consider a decentralized partially observable MDP with $N$ agents. Let $\boldsymbol{\mu} = \left\{\mu_1, \cdots, \mu_N\right\}$ be the set of all agents' deterministic policies parameterized by $\boldsymbol{\theta}=\left\{\theta_1, \cdots, \theta_N\right\}$. Define $\boldsymbol{w} = \left\{w_1, \cdots, w_N\right\}$ as the parameters set of centralized critic functions. Each agent $i$ maintains a centralized critic $Q_i^{w_i}(s_{\text{G}},\bm{a})$ to estimate joint action values, and its goal is to directly adjust the parameters $\theta_i$ of the policy to maximize the objective $J(\mu_i)=\mathbb{E}_{s \sim p^{\bm{\mu}}} R_i(s ,a )$, where $p^{\bm{\mu}}$ is the state distribution caused by the policy $\bm{\mu}$, and $R_i$ is the total expected return $R_i = \sum_ {t =0}^ T \gamma^t r_i^t$. Then, the gradient of the objective $J(\mu_i)$ is given by
\begin{equation}\label{eq_actor_loss}
    \nabla_{\theta_{i}} J\left(\mu_{i}\right)=\mathbb{E}_{ \mathcal{D}}\left[\left.\nabla_{\theta_{i}} \mu_{i}\left(a_{i} \mid s\right) \nabla_{a_{i}} Q_{i}^{w_i}\left(s_{\text{G}}, \bm{a}\right)\right|_{a_{i}=\mu_{i}\left(s\right)}\right],
\end{equation}
where the buffer $\mathcal{D}$ records experiences of all agents. Note that this gradient relies on $\nabla_{a_{i}} Q_{i}^{w_i}(s_{\text{G}}, \bm{a})$, a good value function estimate can provide more accurate gradients for policy learning. Traditional methods of training value functions are usually based on the TD-loss:
% Note that this gradient depends on $\nabla_{a_{i}} Q_{i}^{w_i}(s_{\text{G}}, a_{1}, \ldots, a_{N})$, a good The value function estimate of may be able to provide more accurate gradients for policy learning. Traditional methods of training value functions are usually based on TD-loss:
\begin{equation}\label{eq_TD_loss}
    \mathcal{L}(w_i) = \mathbb{E}_{\mathcal{D}}\left[\left(Q_{i}^{w_i}\left(s_{\text{G}}, \bm{a}\right)-y_i\right)^2\right],
\end{equation}
where $y_i$ is the target value function, calculated by $y_i =r_i+\gamma Q_i^{w^{\prime}_i}(s^{\prime}, a^{\prime}_1, \cdots, a^{\prime}_N)|a^{\prime}_i= \mu_{i}^{\prime}(s^{\prime})$ with $\mu_{i}^{\prime}$ being the target policy with delayed parameters, and $Q_i^{w_i^{\prime}}$ being the target value function with parameter $w_i^{\prime}$. 
Since each $w_i^{\prime}$ is learned separately, the proposed deep PR scheme can be used to design more efficient objective value functions $Q_{\star,i}^{w^{\prime}_i}$ based on Definition~\ref{def4}, which is given by
$$Q_{\star,i}^{w^{\prime}_i}\big(s^{\prime}_{\text{G}}, \bm{a}^{\prime}\big) = (1-\kappa)Q_i^{w^{\prime}_i}\big(s^{\prime}_{\text{G}}, \bm{a}^{\prime}\big)+\kappa \sum_{j \in \mathcal{N}_i}\sum_{s^{\sharp, \prime}_{\text{G}} \in \mathcal{S}^{\sharp}}Q_j^{w^{\prime}_j}\big(s^{\sharp, \prime}_{\text{G}}, \bm{a}^{\prime}\big).$$
Then the target value function $y_i$ in~\eqref{eq_TD_loss} can be written as 
$y_i =r_i+\gamma Q_{\star,i}^{w^{\prime}_i}(s^{\prime}_{\text{G}}, \bm{a}^{\prime})$. 
After enough iterations, we obtain a well-trained model $(w_i, \theta_i), \forall i \in \mathcal{N}$. Finally, we summarize the deep multi-agent PR algorithm in Algorithm~\ref{alg_PR2} for clarification.

% \subsection{Policy Reciprocity on MAPPO}
% MAPPO is 
% After iteration, we obtain a well-trained model $(w_i, \theta_i), \forall i \in \mathcal{N}$. 
% We summarize the tabular multi-agent PR and deep multi-agent PR algorithms in Algorithm \ref{alg_PR} for clarification. 
% From Definition~\ref{def4}, since the proposed deep PR scheme also can share $V(s)$ and $A(s,a)$, it can
% be pluged into other existing multi-agent RL algorithms straightforwardly, such as MAPPO \cite{yu2021surprising}. Moreover, this theme can also be combined with some independent RL algorithms, such as TD3 \cite{fujimoto2018addressing}, to speed up agent learning in a decentralized method.

% {teal}
\begin{algorithm}[t]\small\caption{Deep PR Algorithms} %算法的名字
% \hspace*{0.02in} % \hspace*{0.02in}用来控制位置，同时利用 \\ 进行换行
\begin{algorithmic}[1] %[1] enables line numbers
\REQUIRE Training Epoch $T$; weight Factor $\alpha_0, \beta_0, \kappa$
\STATE Initialize the actor and critic parameters $w_i, \theta_i, \forall i=1,\cdots,N$
\FOR[\emph{parallel do with $N$ agents}]{$t = 1, \cdots, T$}
\STATE Update $Q_{\star,i}^t$ based on \eqref{eq_global}
\FOR{$e= 1, \cdots, E$}
\STATE Interact with the environment using behavior policy $\mu_i$
\STATE Sample a batch of transition sample $(s,a,r,s^{\prime})$
\STATE  Update the critic model parameters $w_i$ by \eqref{eq_TD_loss}
\STATE Update the actor model parameters $\theta_i$ by \eqref{eq_actor_loss}
% \EndFor
\ENDFOR
\ENDFOR
\ENSURE
Actor and critic parameters $w_i, \theta_i, \forall i=1,\cdots,N$
\end{algorithmic}
\label{alg_PR2}
\end{algorithm}

\section{Experiments Results}
In this section, we first evaluate multi-agent PR on discrete control tasks based on independent $Q$-learning \cite{tan1993multi}, an efficient method in the tabular setting.
We then conduct experiments on continuous control tasks to evaluate deep PR. In each iteration, we generate adjacency states for a mini-batch of transition data based on adjacency conditions to estimate the value function, which is simple and effective to be plugged in different algorithms.
To verify that PR can improve many existing RL methods, we consider a series of classical algorithms such as TD3 \cite{fujimoto2018addressing}, DDPG \cite{lillicrap2015continuous}, MADDPG \cite{lowe2017multi}, QMIX \cite{rashid2018qmix}, MAPPO \cite{yu2021surprising}, and MAPTF \cite{yang2021efficient}, they show amazing results in many RL and transfer RL settings.
For a fair comparison, except for the PR module, other details are set the same.

% In this section, we first evaluate multi-agent PR on discrete control tasks based on independent $Q$-learning \cite{tan1993multi}, which is an effective method under the tabular setting.
% Then we conduct experiments on the continuous control tasks to evaluate deep PR. In order to verify that PR can improve many existing RL methods, we considered a series of classical algorithms, such as TD3 \cite{fujimoto2018addressing}, DDPG \cite{lillicrap2015continuous}, MADDPG \cite{lowe2017multi}, QMIX \cite{rashid2018qmix} and MAPPO \cite{yu2021surprising}, which have shown the surprising effectiveness in many RL environments.

\begin{figure}[!tbp]
	\centering
% 	\hspace{-0.3cm}
    \subfloat{
	\includegraphics[width=2.5in]{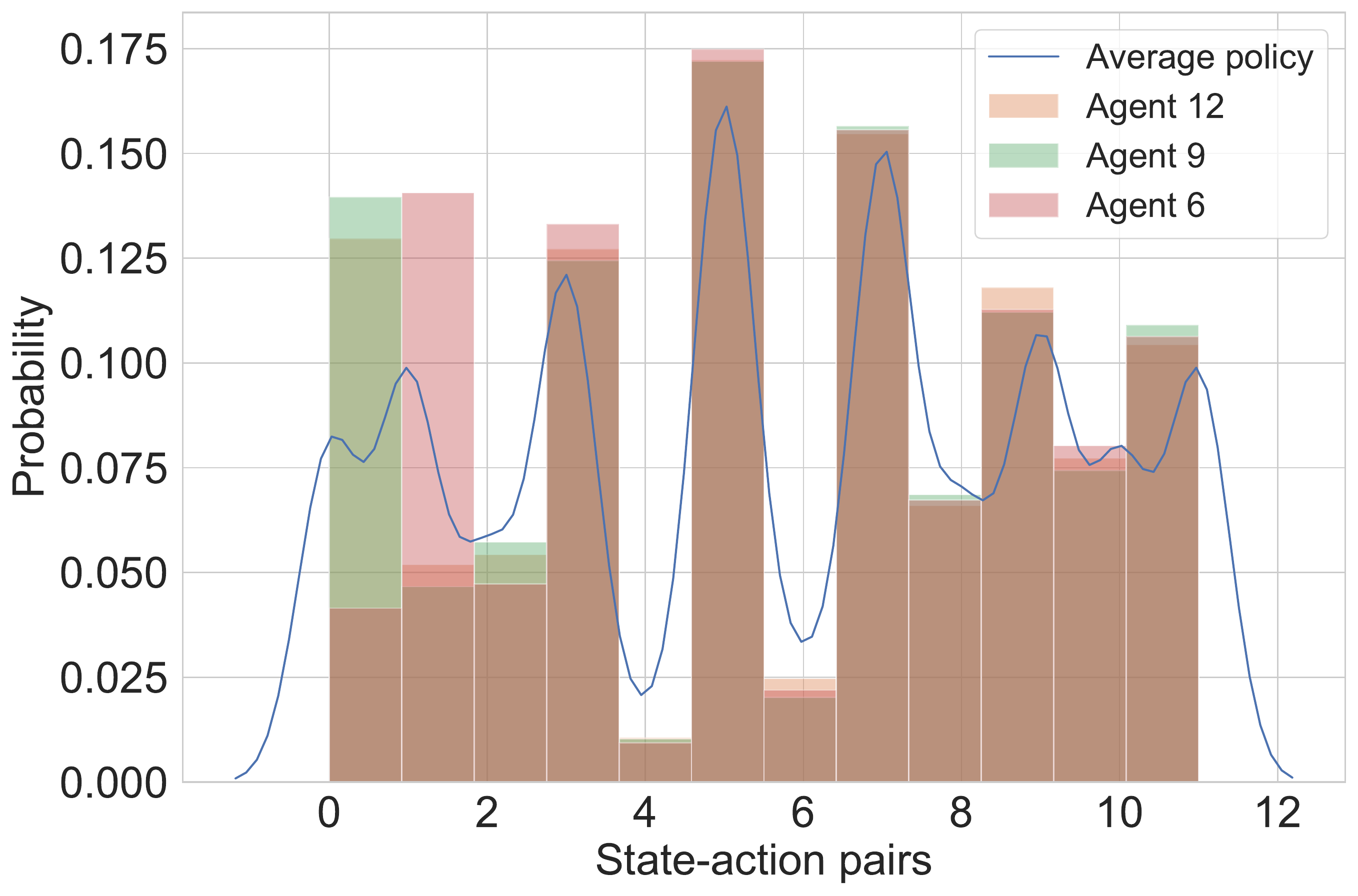}
	}
	
	\subfloat{
	\includegraphics[width=2.5in]{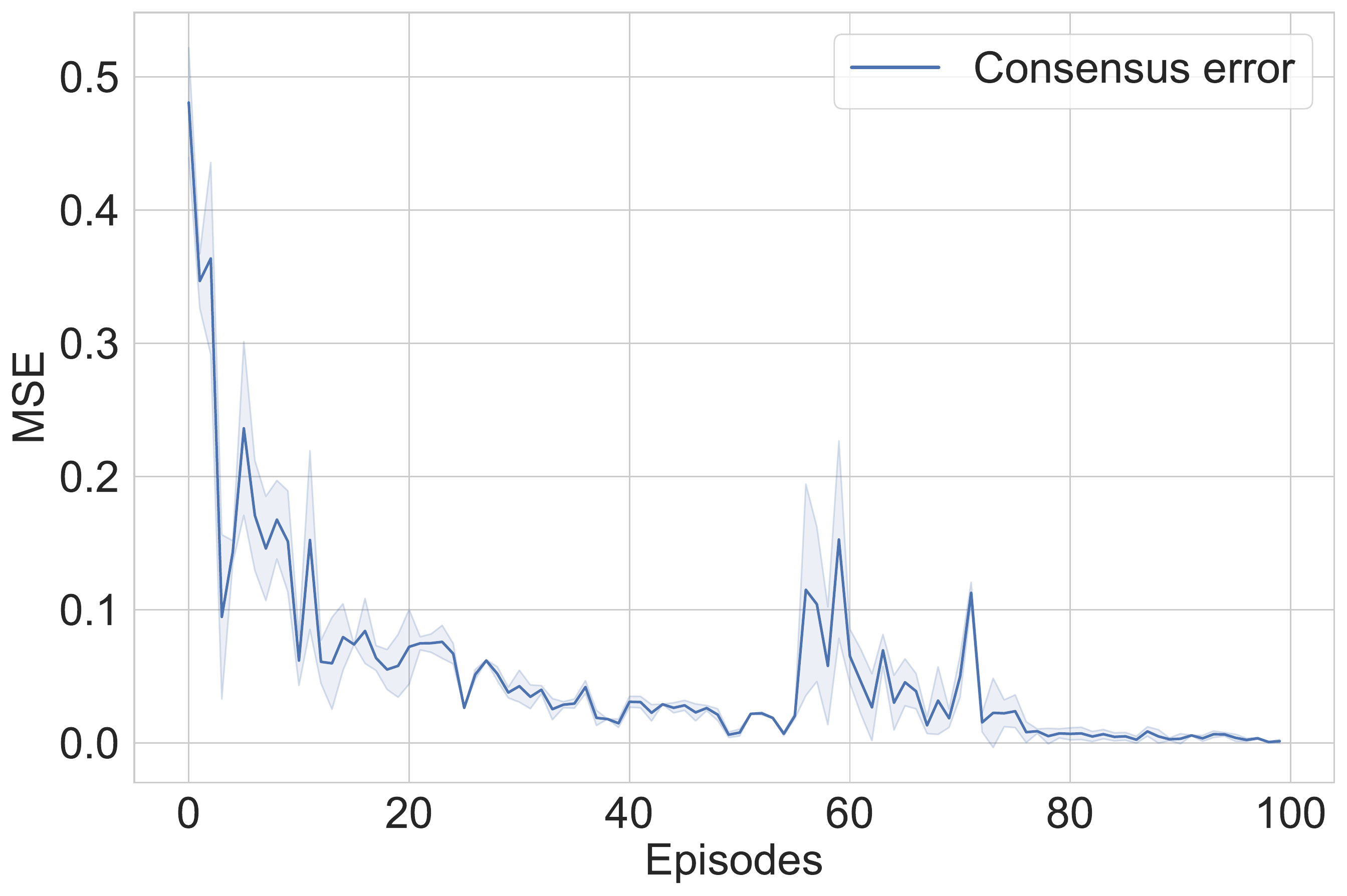}
	}

	\caption{The consensus analysis on the Digital environment. \textbf{Upper:} normalized policy distribution. Agents 6, 9 and 12 can reach consensus with the average policy. \textbf{Low:} average consensus error.}
	\label{fig_ql}
\end{figure}

\subsection{Tabular PR Experiment Results}

\begin{figure*}[!ht]
	\centering
% 	\hspace{-0.3cm}
    \subfloat{
	\includegraphics[width=2.2in]{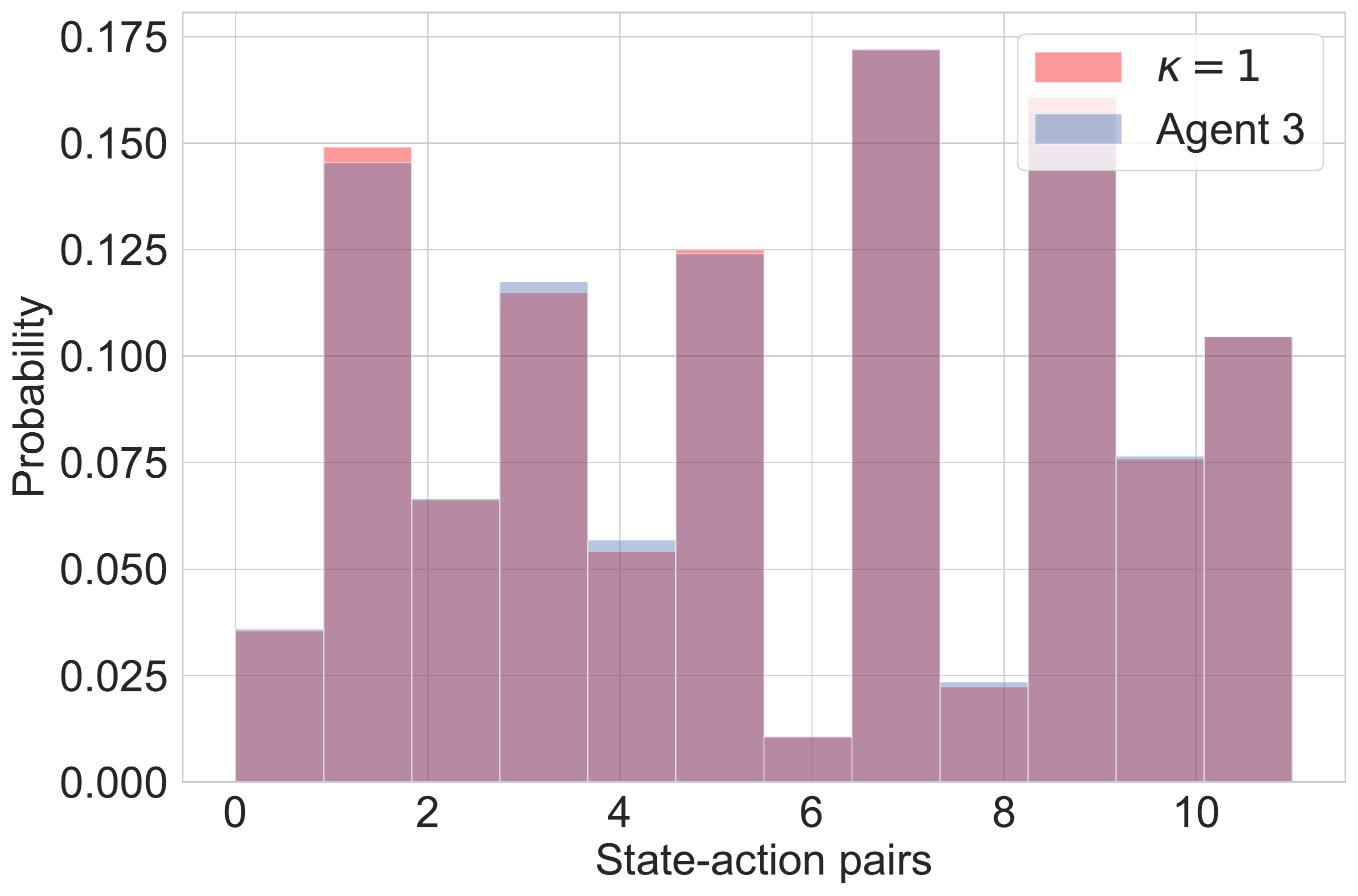}
	}
    \subfloat{
	\includegraphics[width=2.2in]{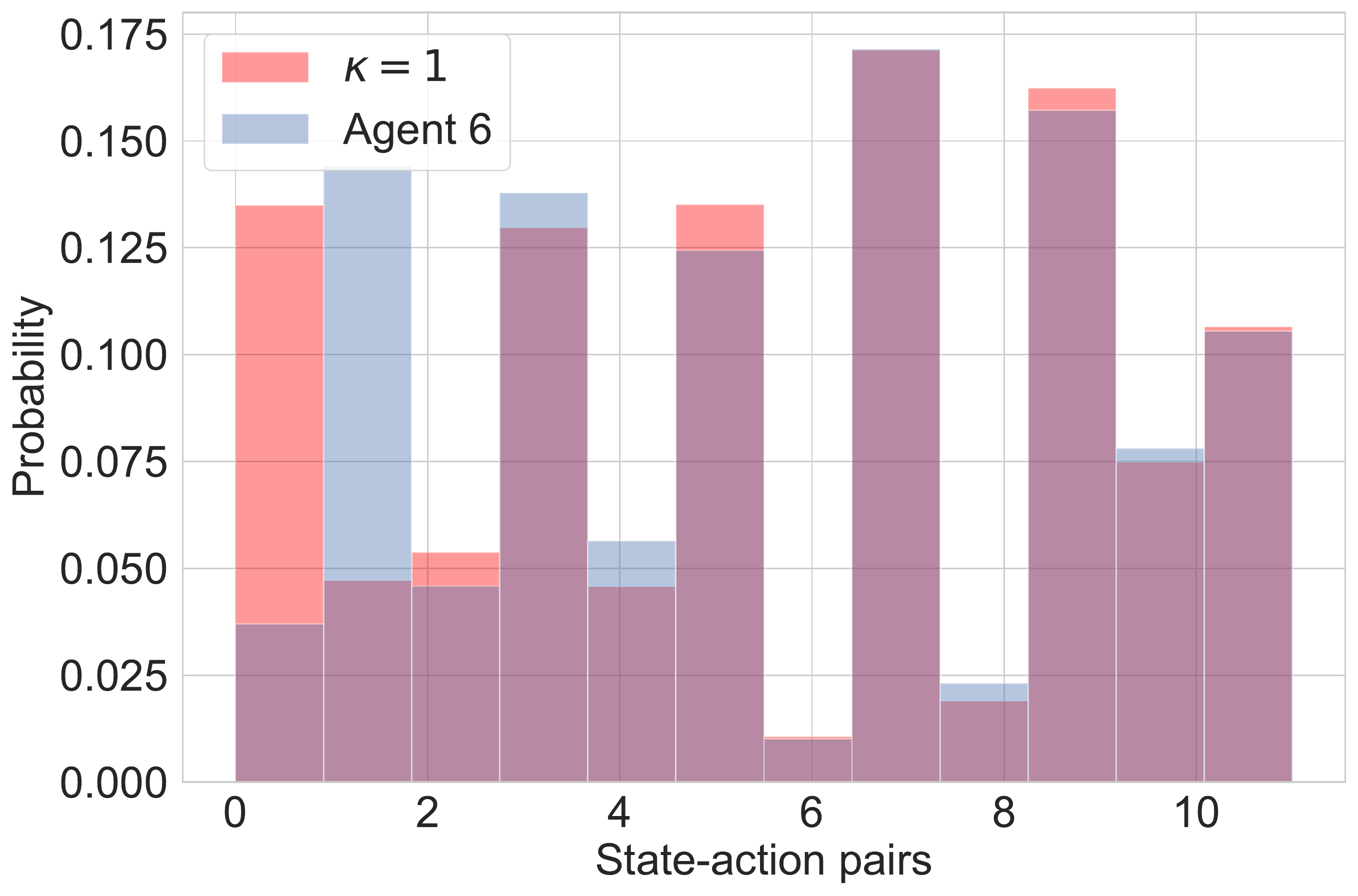}
	}
    \subfloat{
	\includegraphics[width=2.2in]{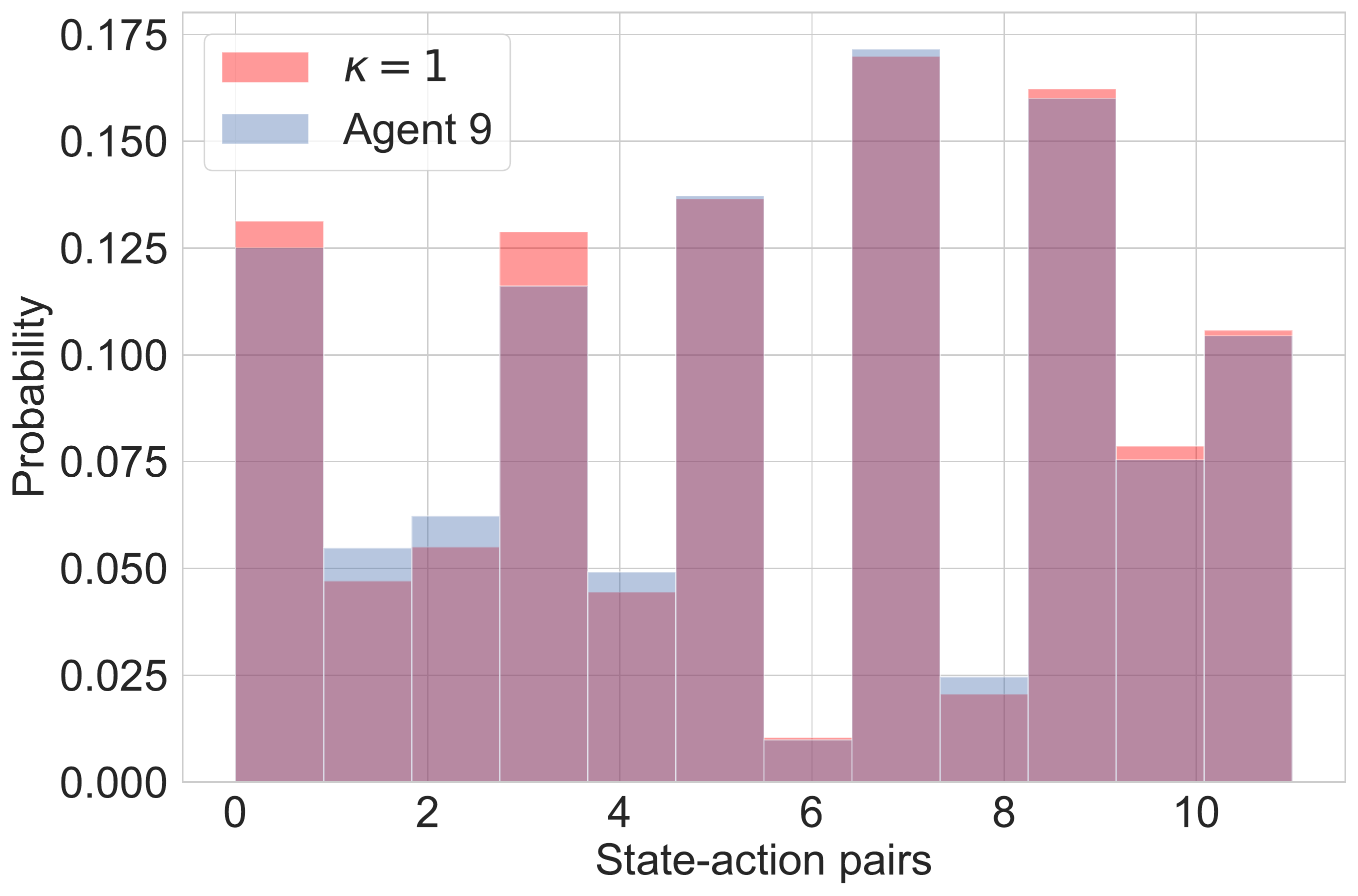}
	}
	
	\subfloat{
	\includegraphics[width=2.2in]{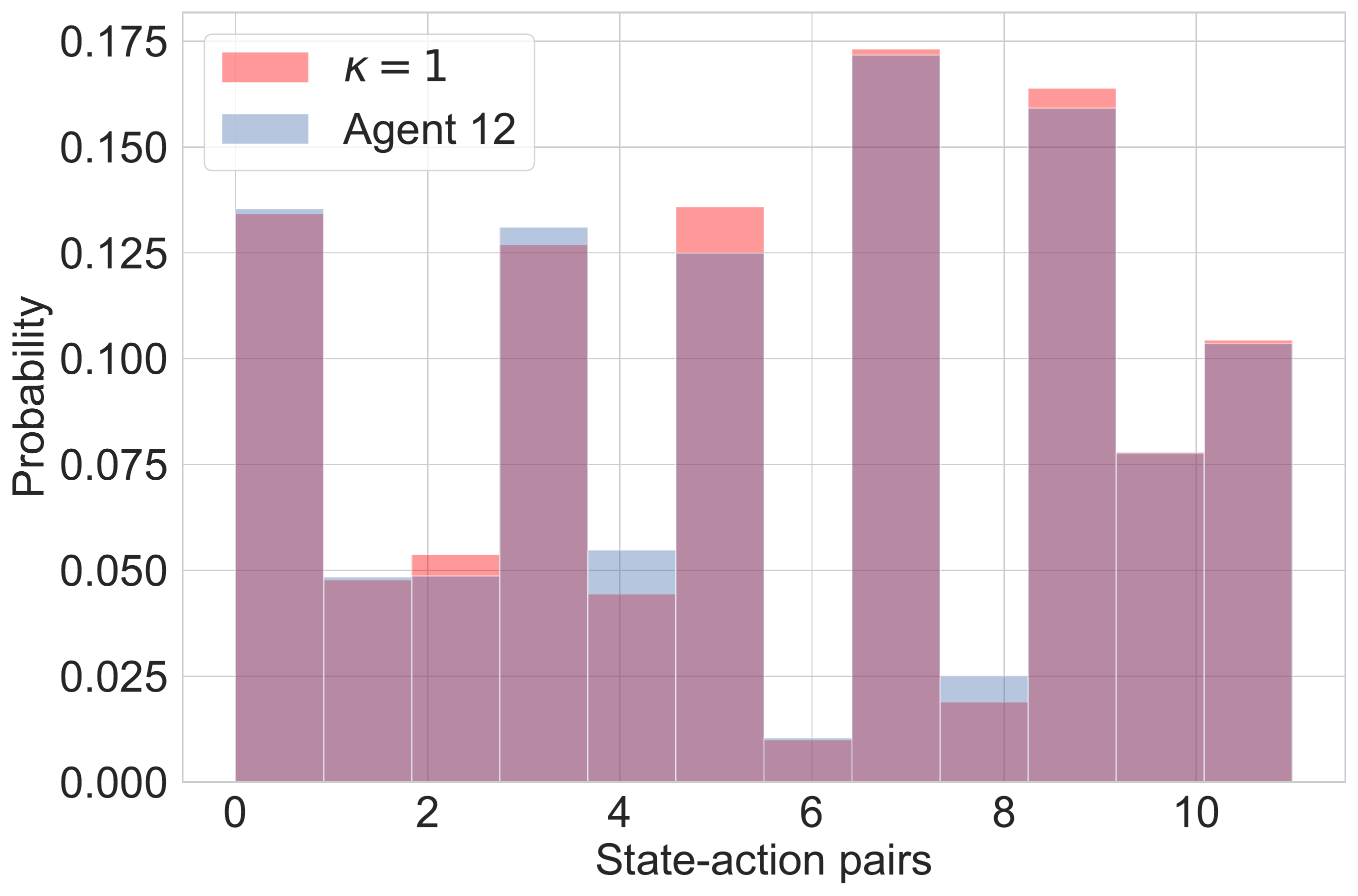}
	}
    \subfloat{
	\includegraphics[width=2.2in]{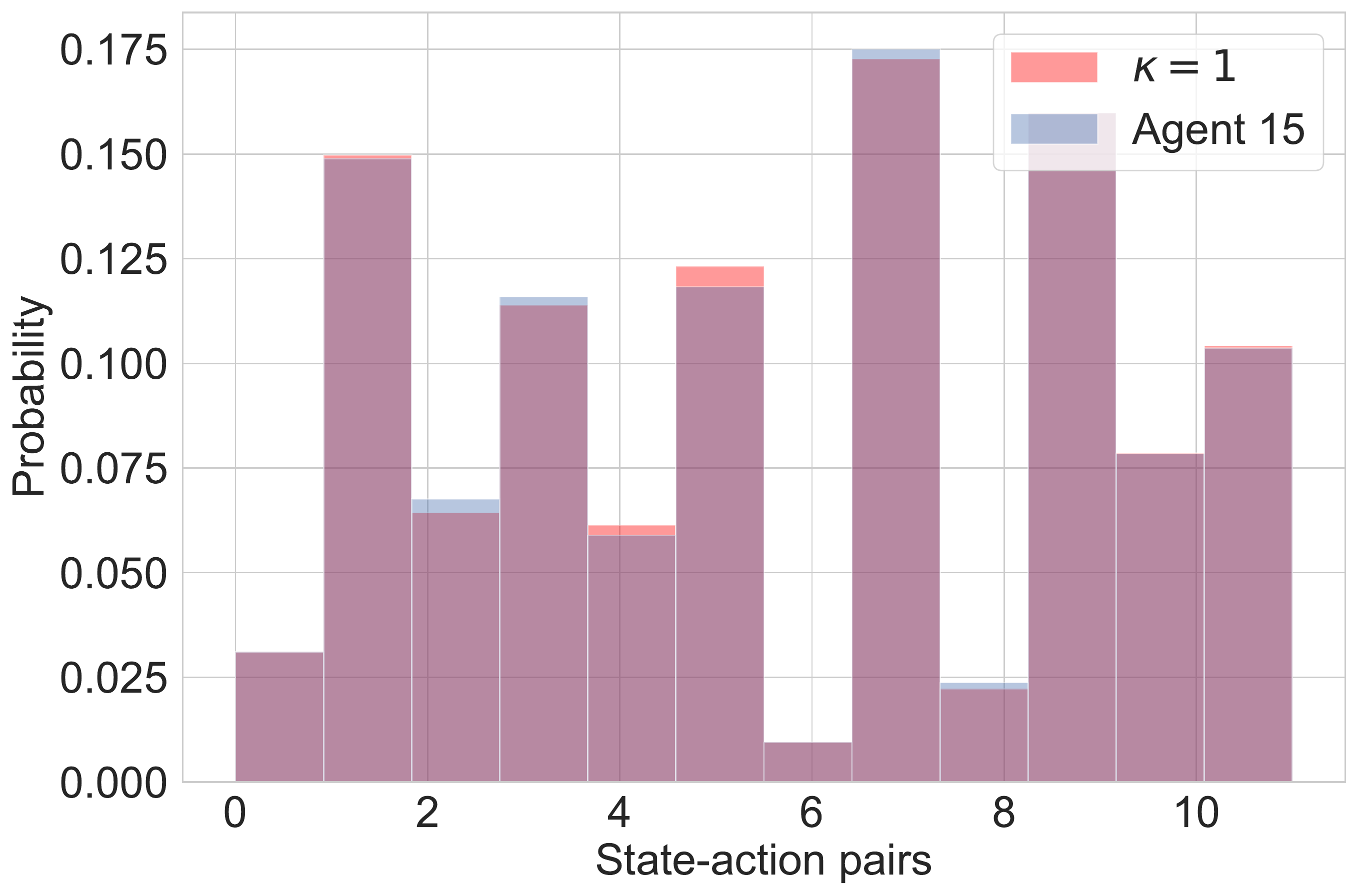}
	}
	\subfloat{
	\includegraphics[width=2.2in]{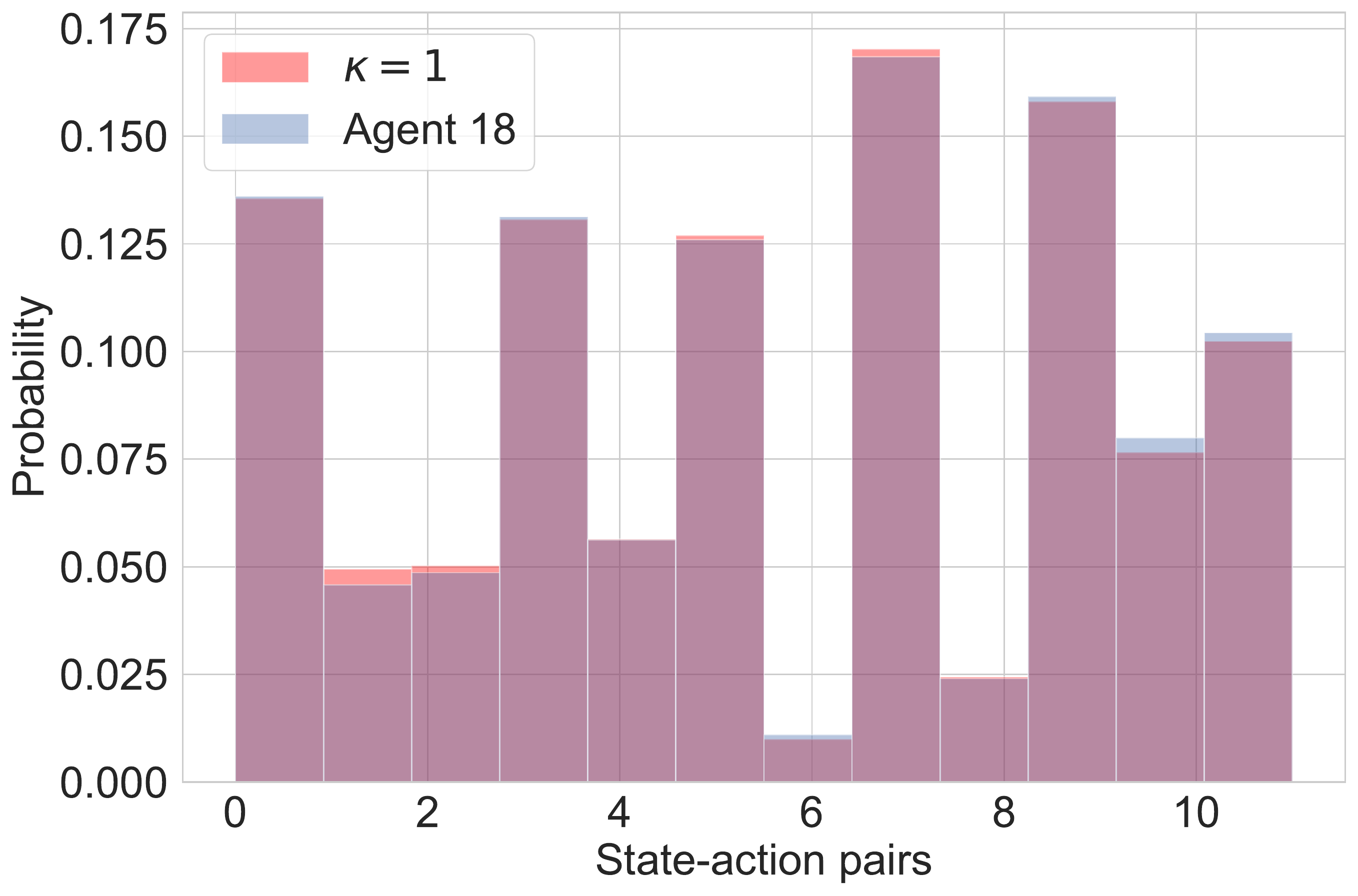}
	}
	\caption{The consensus analysis on the digital environment.}
	\label{fig_ql2}
\end{figure*}

\begin{figure*}[!tbp]
	\centering
% 	\hspace{-0.3cm}
    \subfloat{
	\includegraphics[width=2.2in]{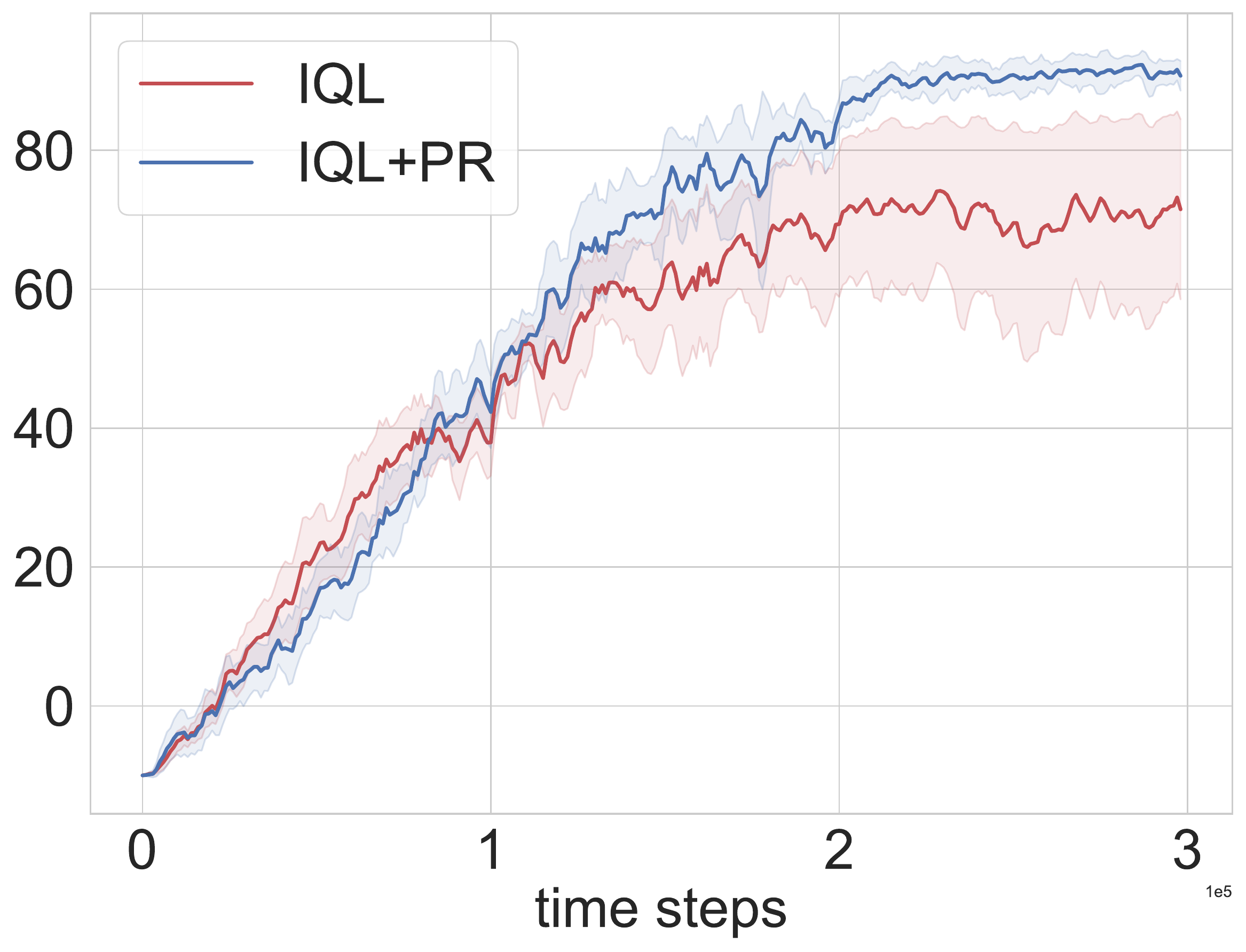}
	}
	\subfloat{
	\includegraphics[width=2.25in]{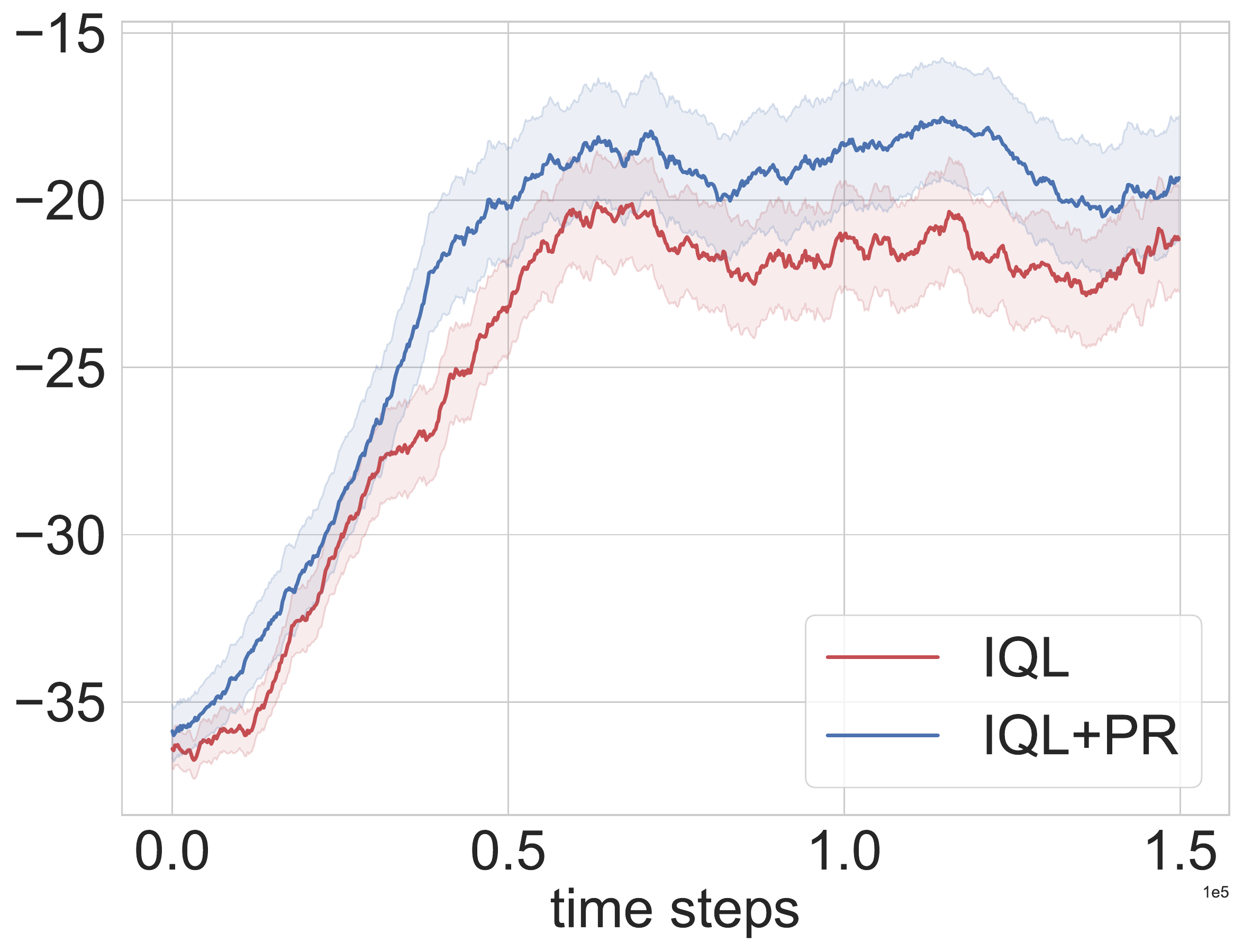}
	}
	\subfloat{
	\includegraphics[width=2.2in]{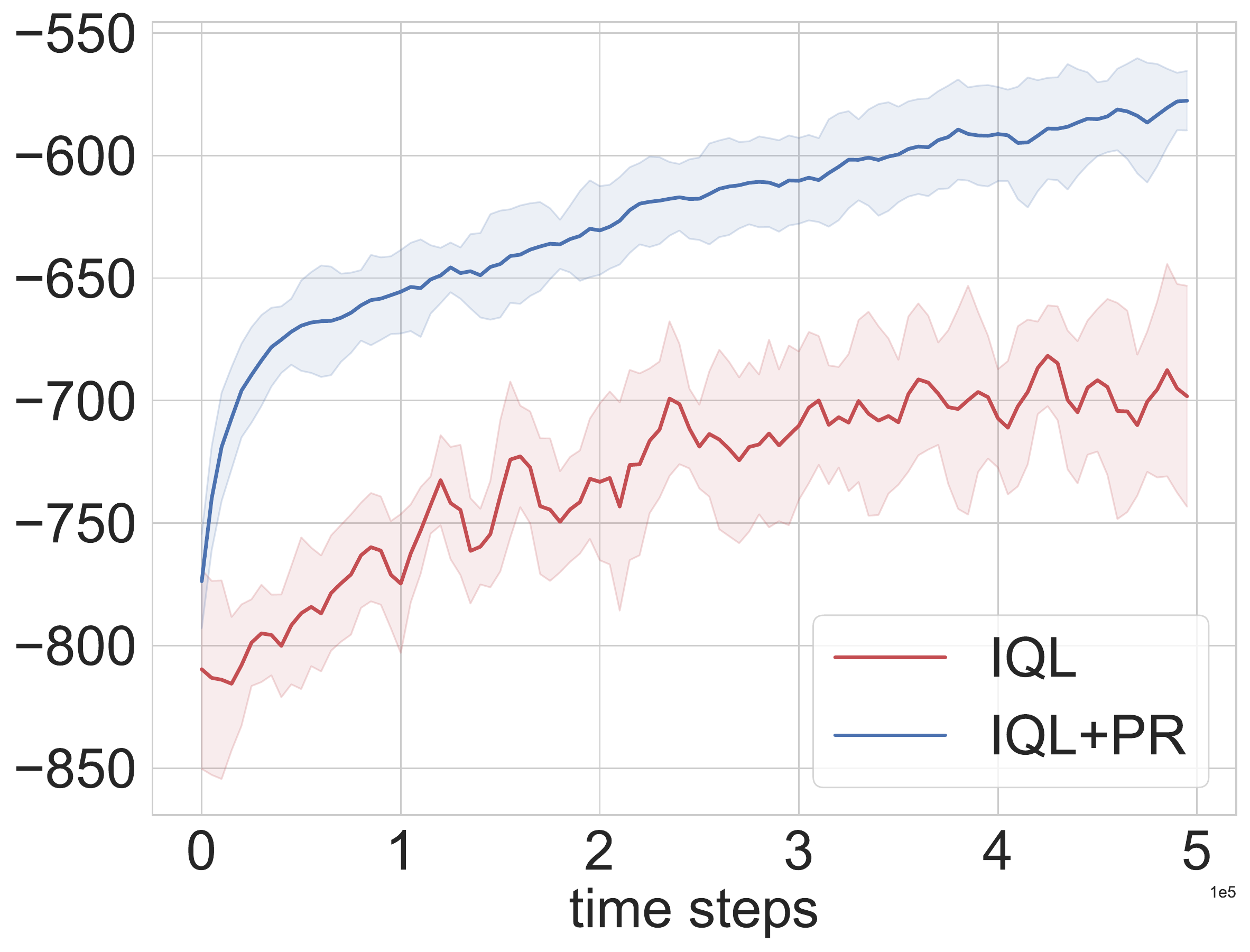}
	}

	\caption{Average return of different algorithms on discrete environments. All results is averaged by five seeds. \textbf{Left:} Landmarks, $|\mathcal{N}|=2$.  \textbf{Middle:} Blocker, $|\mathcal{N}|=3$. \textbf{Right:} Landmarks, $|\mathcal{N}|=10$}
	\label{fig_return}
\end{figure*}
Similar to \cite{dann2014policy, zhang2018fully}, we consider a Digital environment with $|\mathcal{S}| = 20$ states and $N=20$ agents.
Each agent has a binary-valued action space, i.e., $\mathcal{A}_i \in \left\{0,1\right\}$ and the elements in the transition probability matrix $p$ are uniformly sampled from $[0, 1]$ and normalized to be stochastic. For any state-action pair
$(s, a)$, the mean reward $r_i(s, a)$ of agent $i$ is sampled uniformly from $[0, 4]$, while the instantaneous rewards $r_i^t$ are sampled from the uniform distribution $[r_i(s, a)-0.5, r_i(s, a)+0.5]$. 
Moreover, we also consider a discrete Navigation environment, where each agent needs to reach landmarks \cite{lowe2017multi}, and a Blocker environment which requires agents to reach the bottom row by coordinating with its teammates to deceive blockers \cite{yang2020multi,heess2013actor}. 
For the adjacency state requirement, we choose the adjacent digital as the similar states on the digital environments. 
For Landmarks and Blockers, we make the observations of each agent one dimension less than the global state so that these states can be transformed into a set of adjacency states.
% As for Landmarks and Blocker, we make each agent's observations one dimension less than the global state, so that these states can be transformed to form sets of adjacency states. 
In this case, the adjacency level is $1$, i.e., $d_{s^{\sharp}}=d_{\sharp}=1$, then we use a generalized version of \eqref{eq_average} in the experiments, i.e., $Q_{\sharp,i}^t(s,a) = \frac{1}{\left|\mathcal{S}^{\sharp}\right|}\sum_{s^{\sharp} \in \mathcal{S}^{\sharp}}Q_i^t(s^{\sharp},a)$. More details can be found in the appendix.
We set the discount factor as 0.99 in Navigation, Blocker and Landmarks environments, and 0.8 in Digital task. Moreover, we set the learning rate $\alpha=0.5$ and $0.3$ for IQL and IQL+PR, respectively, the weight factor $\beta=0.3$ in Navigation, Digital and $\beta=0.4$ in Blocker, Landmarks, and set $\tau_1=0.65$ and $\tau_2=0.35$ for all environments.

Figure~\ref{fig_ql}-Upper shows the normalized policy distribution of different agents, which presents the consensus analysis of PR. 
We randomly select some state-action pairs and use the Boltzmann policy to compute the probability distribution \cite{sutton2018reinforcement}. 
% For most agents, such as agent 6, agent 9 and agent 12, they can reach consensus with the average policy. 
We can find that the policy probability distribution of each agent is very close in most states. Although some of them are biased, these agents still choose the same action, validating Theorem~\ref{Theorem_1_1}.
% We can find that in most states, each agent's policy probability distribution is very close. Although the probabilities of actions in some states are biased, the agent will still choose the same action based on similar observation with the average policy, which validates Theorem~\ref{Theorem_1_1}. 
We also show each agent's policy can reach consensus with enough iterations when $\kappa=1$, which is presented in the appendix.
Figure~\ref{fig_ql}-Bottom shows the mean squared error (MSE) between the probability distribution of all agent policies and the average policy, which decreases with the training process, implying an asymptotic consensus among agents. Figure~\ref{fig_ql2} shows the consensus analysis with the case $\kappa=1$, including agents $3,6,9,12,15$ and $18$.
We found that all the agents can reach consensus in most state-action pairs, which validates Corollary~\ref{Theorem_1_2}.

Figure~\ref{fig_return} shows the average return of different algorithms under multiple discrete environments.
As the statement of Theorem~\ref{Theorem_2}, our method can iterate to the optimal value gradually. In addition, this iterative method converges faster than IQL, confirming the effectiveness of adjacency states for decision-making.
Figure~\ref{fig_return}-Right shows the performance on a Landmark environment with $|\mathcal{N}|=10$ agents. 
Comparing with Figure~\ref{fig_return}-Left, we find that more performance gains can be obtained when each agent performs PR with more agents.
This may be due to the use of more information to obtain a more accurate estimate of the value function.
% Compare with Figure~\ref{fig_return}-Left, we find that when each agent performs PR with more agents, more performance advantages can be obtained. 
% This may be due to the use of more information to obtain a more accurate estimation of the value function. 

% \textcolor{red}{Figure~\ref{fig_return}-Right shows that when the agent $i$ performs PR with more agents, i.e., $|\mathcal{N}_i|=10$ , more performance advantages can be obtained. 
% This may be due to the use of more information to obtain a more accurate estimation of the value function.}
\begin{figure*}[!tbp]
	\centering
% 	\hspace{-0.3cm}
	\subfloat{
	\includegraphics[width=2.2in]{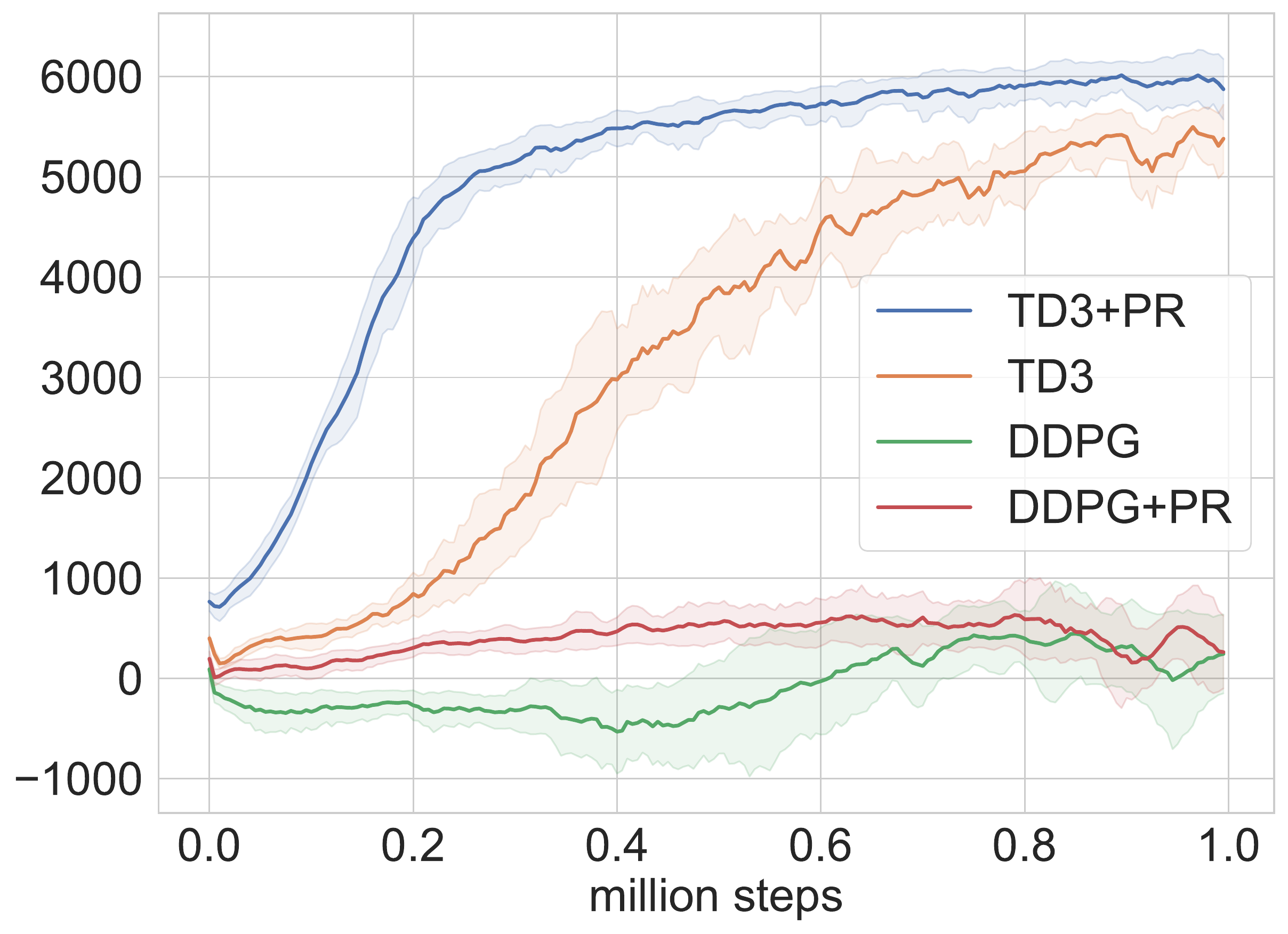}
	}
	\subfloat{
	\includegraphics[width=2.2in]{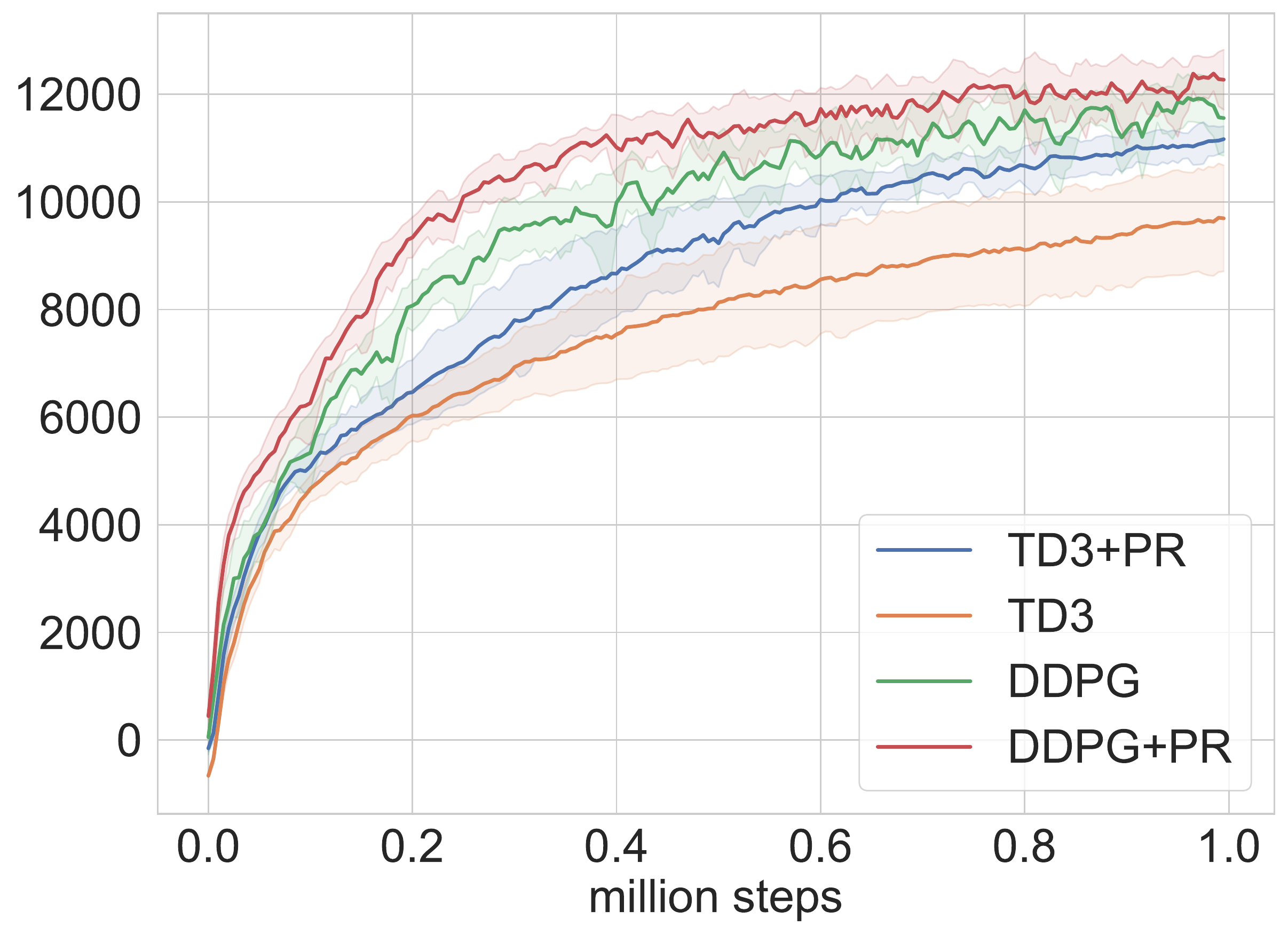}
	}
	\subfloat{
	\includegraphics[width=2.2in]{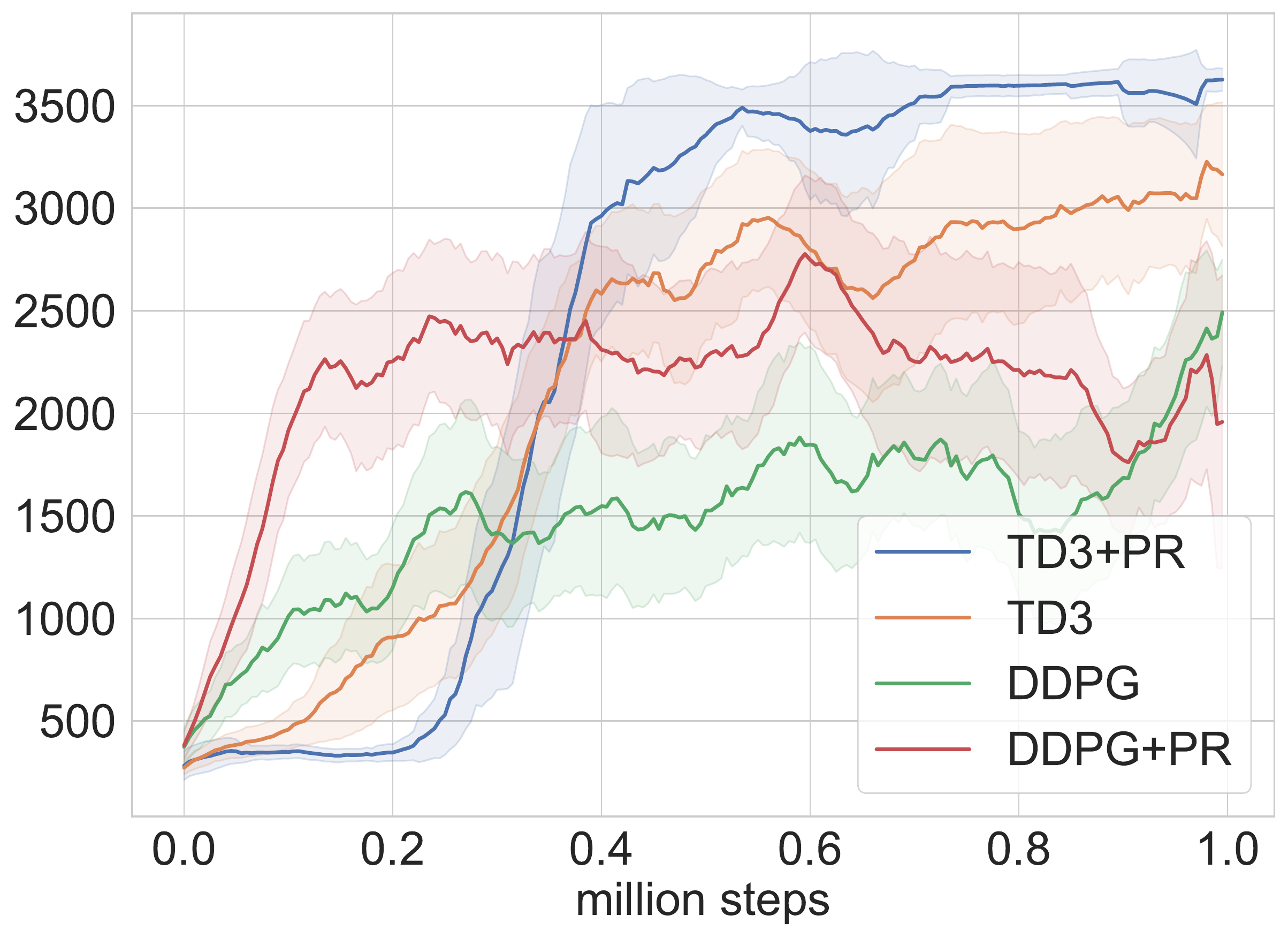}
	}

	\caption{Average return of different algorithms on MuJoCo environments. All results is averaged by five seeds. \textbf{Left:} Ant-v3. \textbf{Middle:} Half-Cheetah-v3. \textbf{Right:} Hopper-v3.}
	\label{fig_return_mujoco}
\end{figure*}

\begin{figure*}[!tbp]
	\centering
% 	\hspace{-0.3cm}
    \subfloat{
	\includegraphics[width=2.2in]{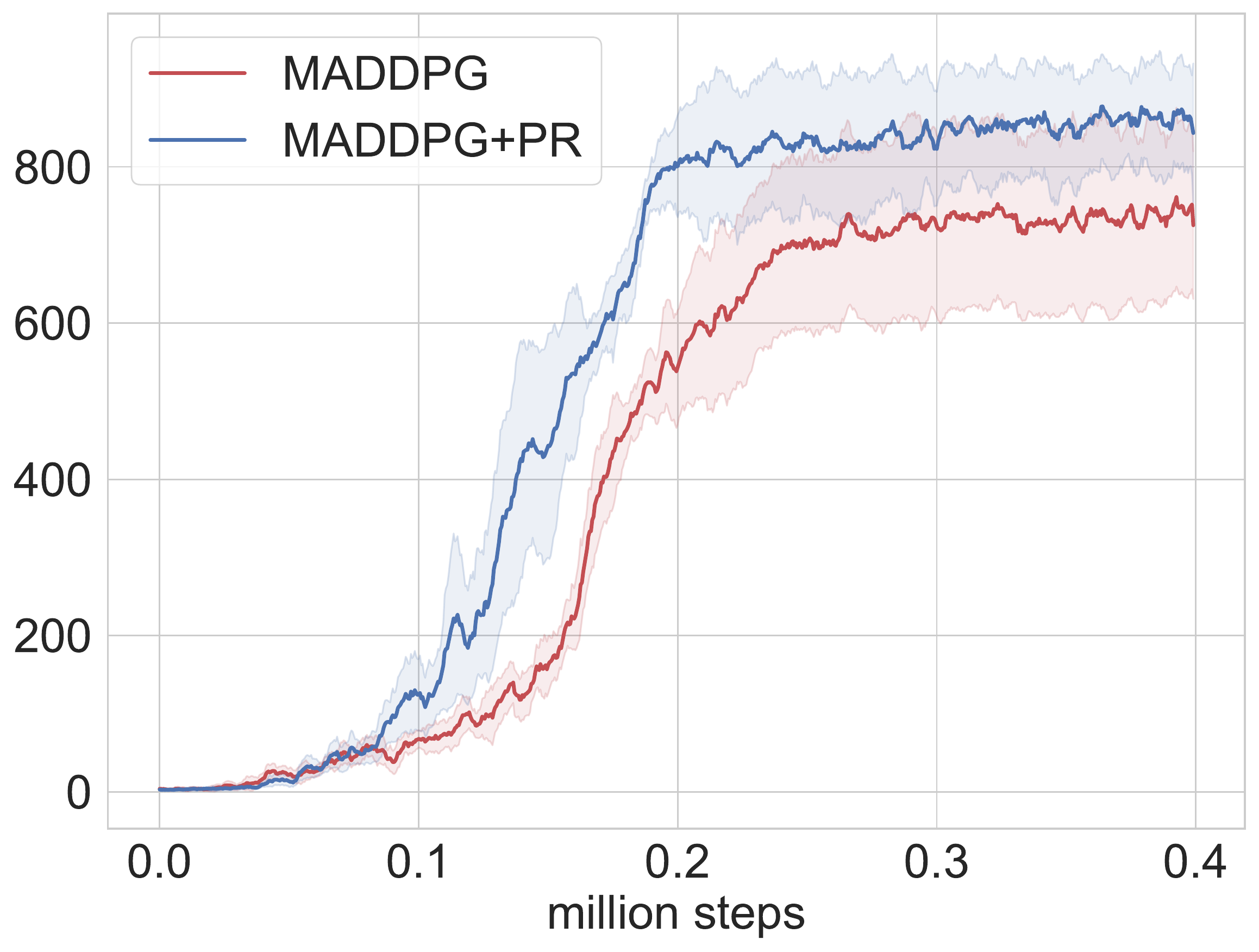}
	}
	\subfloat{
	\includegraphics[width=2.2in]{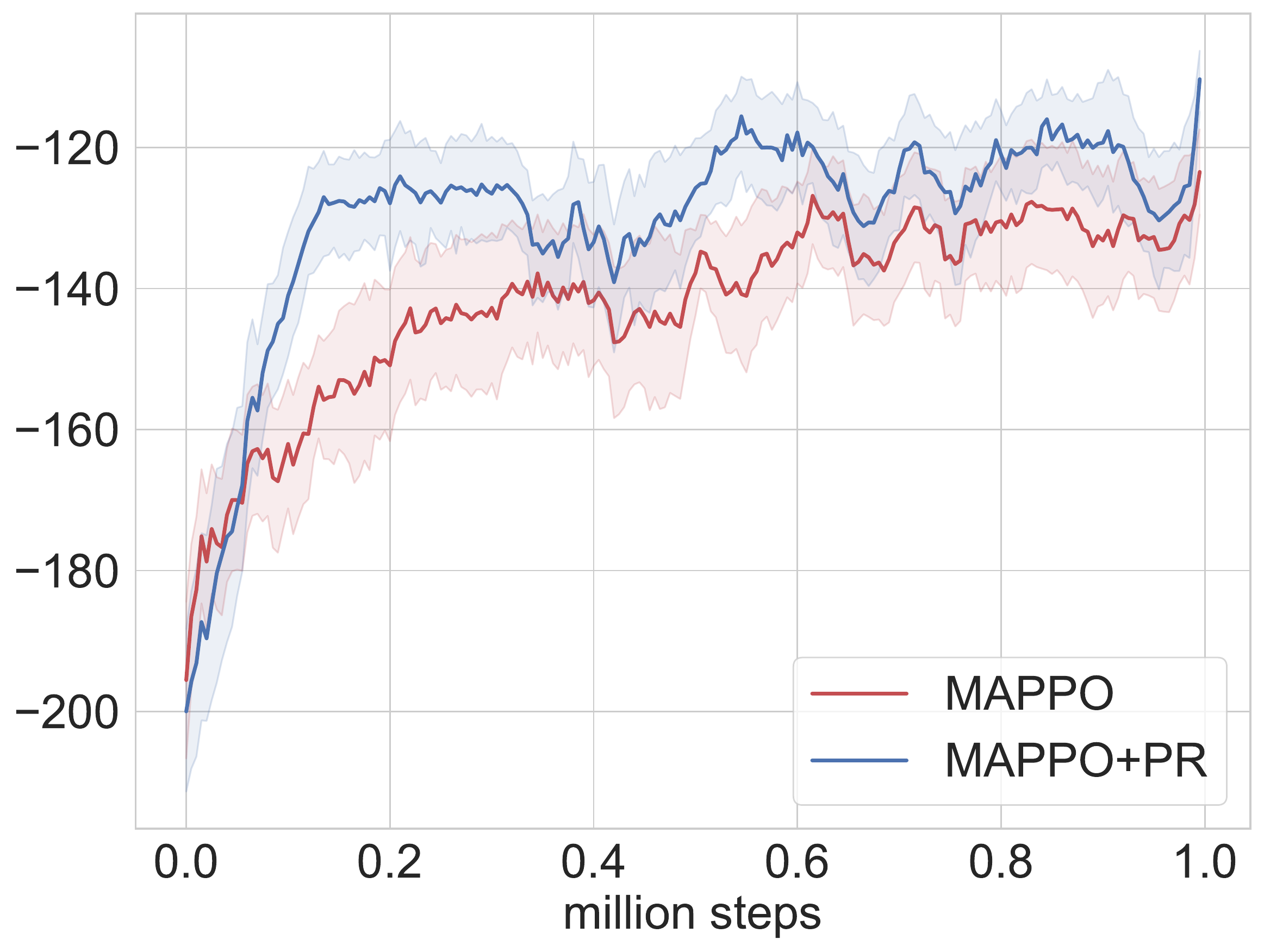}
	}
	\subfloat{
	\includegraphics[width=2.2in]{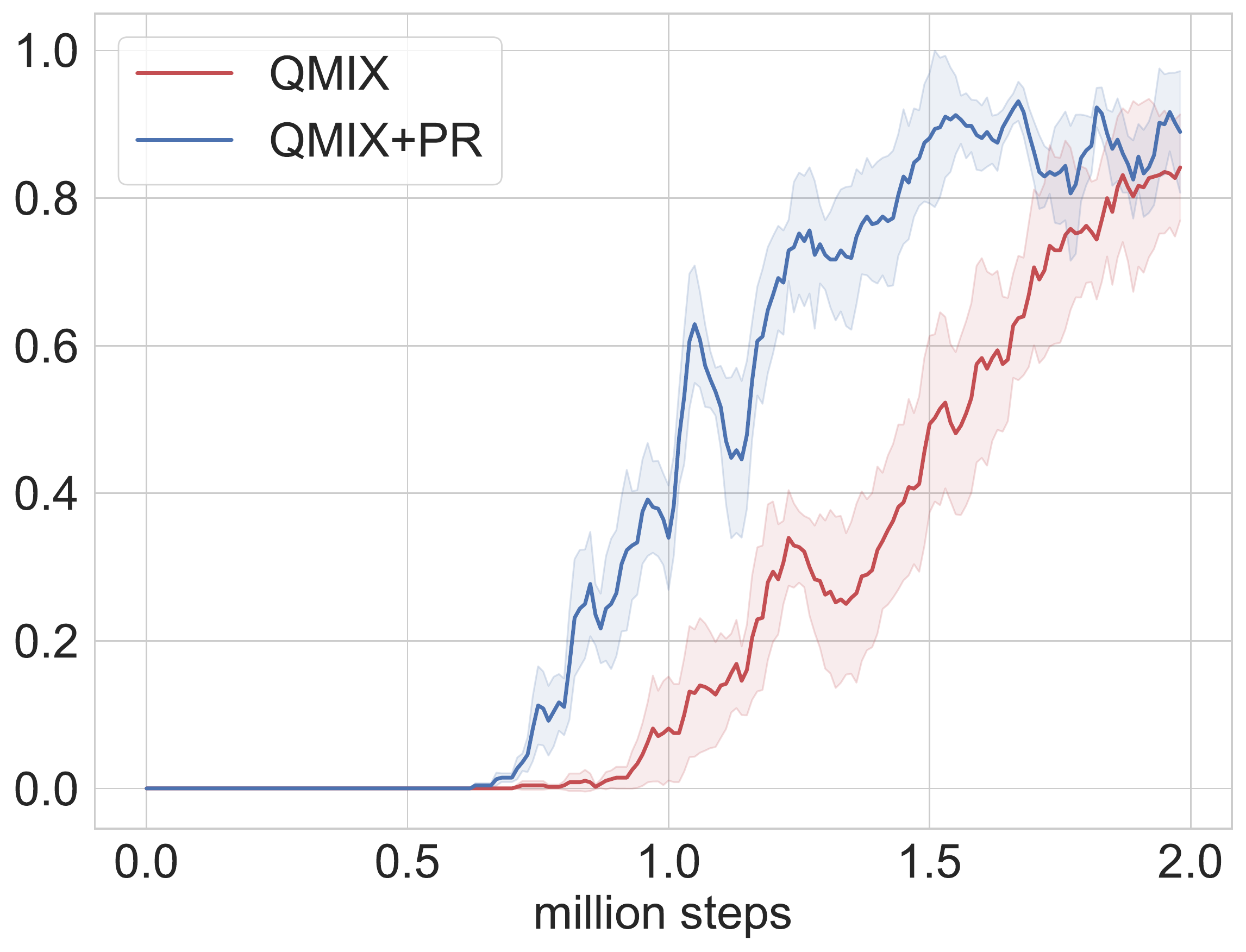}
	}
	
	\subfloat{
	\includegraphics[width=2.2in]{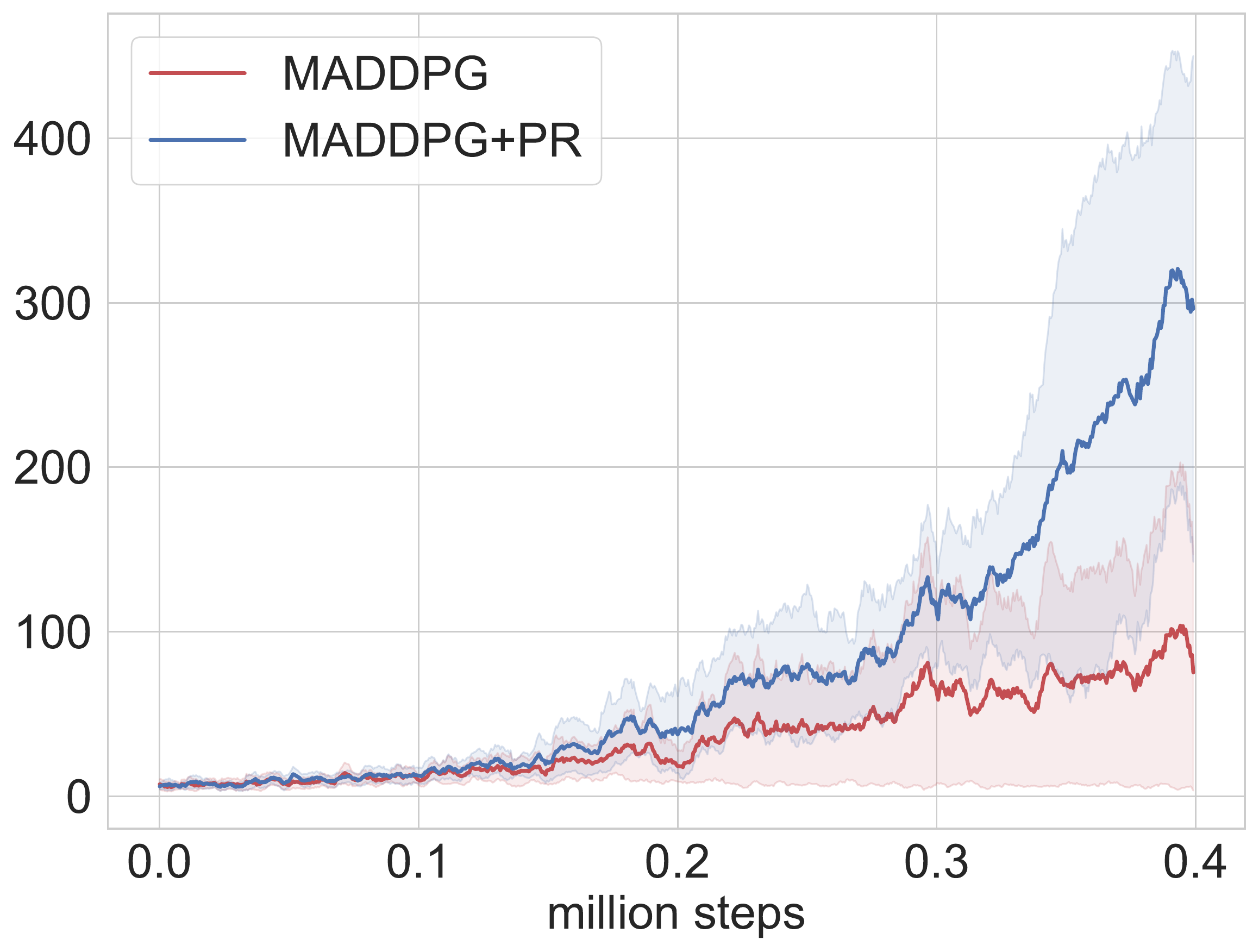}
	}
	\subfloat{
	\includegraphics[width=2.2in]{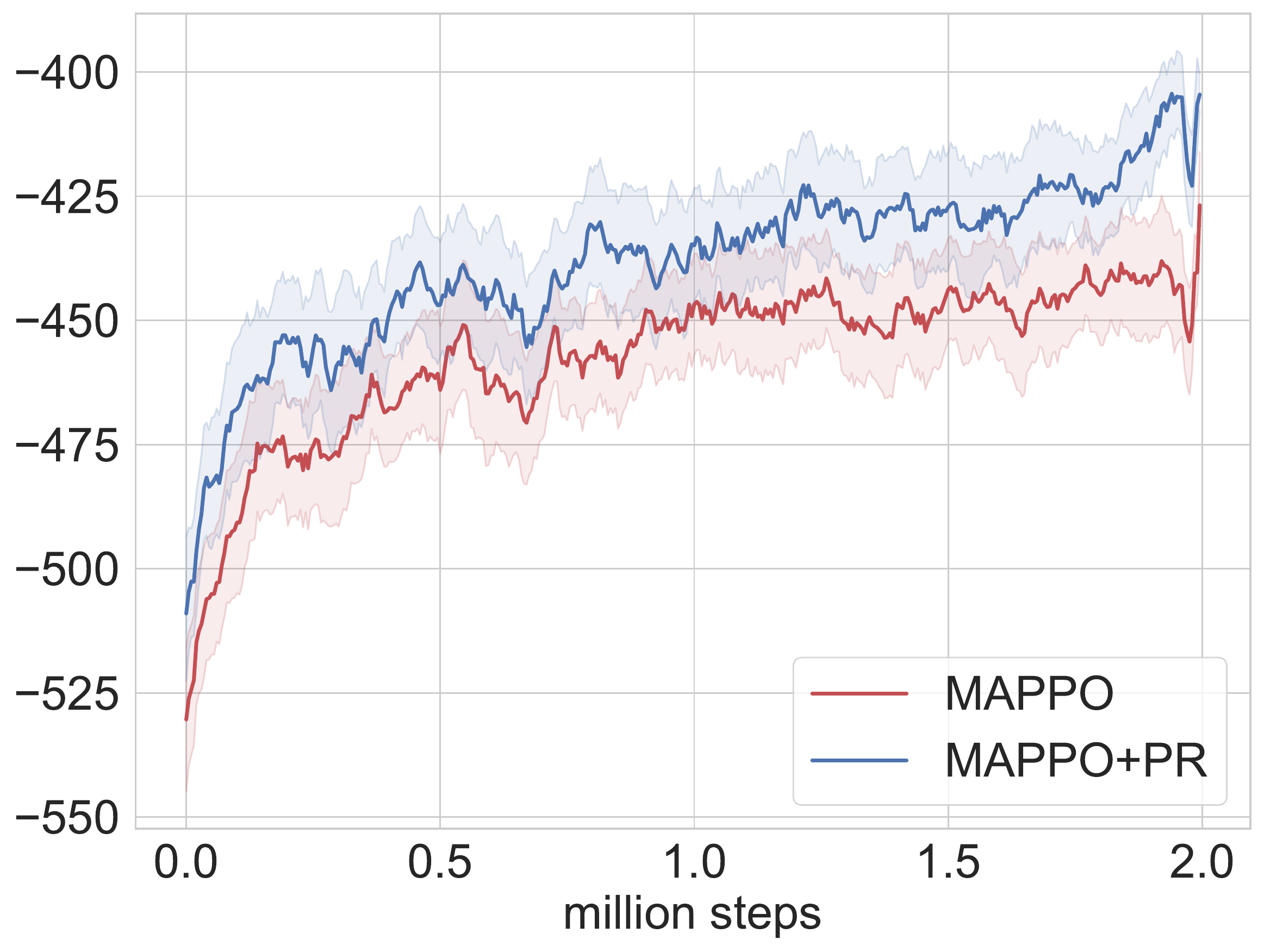}
	}
	\subfloat{
	\includegraphics[width=2.2in]{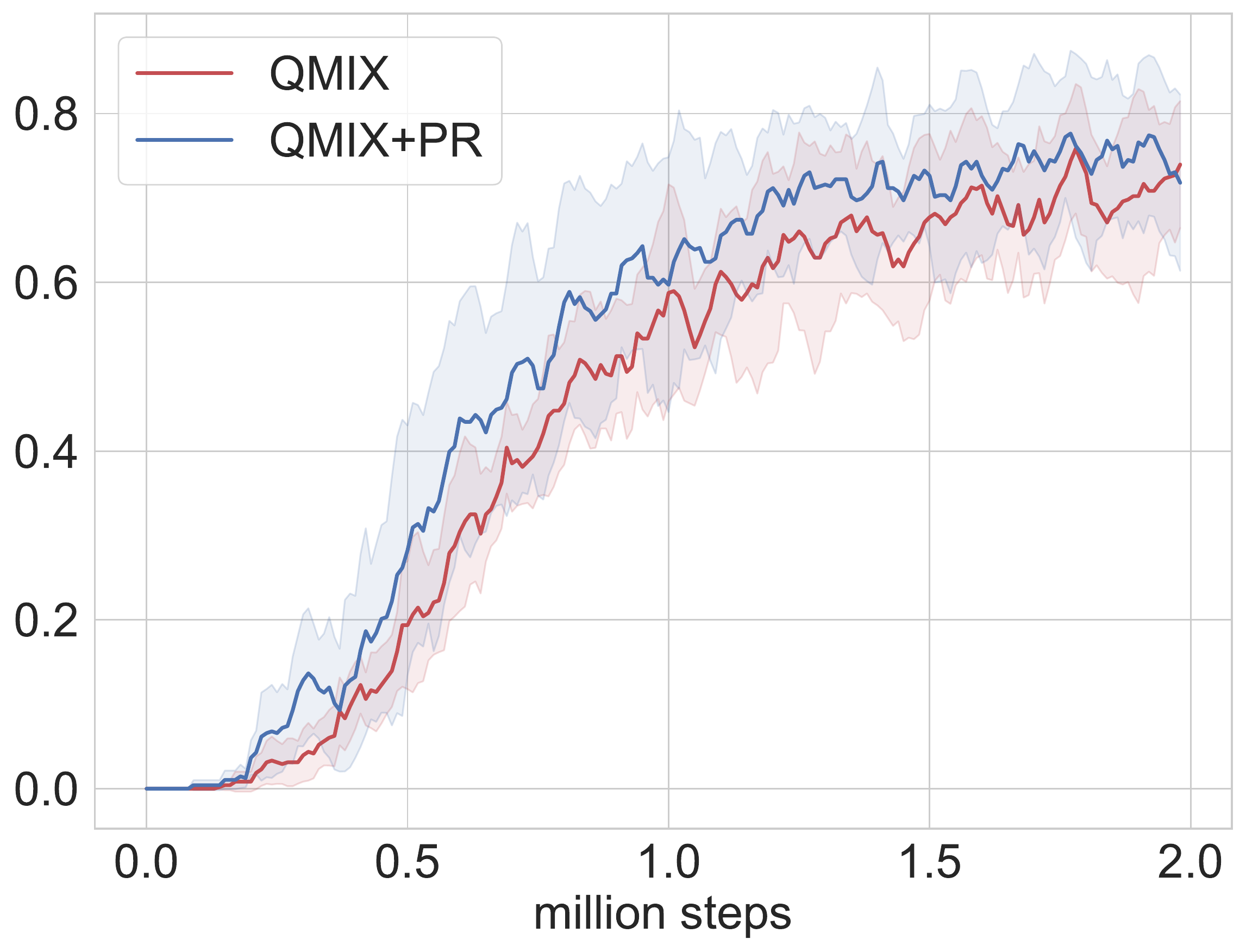}
	}
 	\caption{Average returns on different MPE environments with different algorithms. All results is averaged by five seeds. \textbf{Left:} Predator-prey $|\mathcal{N}|=3$ (top) and $|\mathcal{N}|=6$ (bottom). \textbf{Middle:} Cooperative Navigation $|\mathcal{N}|=3$ (top) and $|\mathcal{N}|=6$ (bottom). \textbf{Right:} StarCraft environments, 3s\_vs\_5z (top) and 5m\_vs\_6m (bottom).}
	\label{fig_deep}
\end{figure*}
\subsection{Deep PR Experiment Results}
We evaluate our deep PR on several continuous control tasks, such as MuJoCo control tasks \cite{todorov2012mujoco}, StarCraft \cite{samvelyan2019starcraft}, Cooperative Navigation and Predator-Prey with different number of agents in Multi-agent Practical Environments (MPE) \cite{lowe2017multi}. 
For the MuJoCo suite, we set the buffer size as $1e+6$, the discount factor as 0.99, the actor and critic learning rate as $3e-4$, the exploration noise as $0.1$, the batch size as $256$ and the actor update frequency as $2$. The soft update rate and weight factor $\kappa$ are set as $0.008$ and $0.3$ in Ant-V3, respectively, and are set as $0.005$ and $0.5$ in other environments.
For the MPE suite, we set the buffer size, the discount factor, the actor learning rate, the critic learning rate, the exploration noise and the soft update rate as $5e+5$, $0.95$, $1e-4$, $1e-3$, $0.1$, and $0.01$ for the MADDPG algorithm,
respectively. 
For MAPPO, we set the discount factor, the actor learning rate, the critic learning rate, the entropy coefficient and the GAE parameter as $0.99$, $1e-4$, $1e-4$, $0.01$, and $0.95$, respectively.
For QMIX, we set the discount factor, the learning rate, $\epsilon$-start, $\epsilon$-end and the $\epsilon$ anneal time as $0.99$, $5e-4$, $1.0$, $0.05$, and $50000$ in all environment, respectively. Moreover, we set the target update interval as 160 and 180 in $\text{3s\_vs\_5z}$ and $\text{5m\_vs\_6m}$, respectively.

% \begin{wrapfigure}[32]{r}{0.60\textwidth}
% \vspace{-14pt}
% 	\centering
% % 	\hspace{-0.3cm}
%     % \subcaption{Predator-prey $|\mathcal{N}|=3$ | $|\mathcal{N}|=6$]}
%     \subfloat{
% 	\includegraphics[width=1.6in]{fig_0512/tag_3_v6.pdf}
% 	}
% 	\subfloat{
% 	\includegraphics[width=1.6in]{fig_0512/tag_6_v6.pdf}
% 	}
	
%     \subfloat{
% 	\includegraphics[width=1.6in]{fig_0512/mappo_spread_3_v6.pdf}
% 	}
% 	\subfloat{
% 	\includegraphics[width=1.6in]{fig_0512/mappo_spread_6_v6.pdf}
% 	}

%     \subfloat{
% 	\includegraphics[width=1.6in]{fig_0512/3s_vs_5z.pdf}
% 	}
% 	\subfloat{
% 	\includegraphics[width=1.6in]{fig_0512/5m_vs_6m.pdf}
% 	}
%  	\caption{Average return on different environments with different algorithms. All results is averaged by five seeds. \textbf{Upper:} Predator-prey $|\mathcal{N}|=3$ and $|\mathcal{N}|=6$. \textbf{Lower:} Cooperative Navigation $|\mathcal{N}|=3$ and $|\mathcal{N}|=6$.}
% 	\label{fig_deep}
% \end{wrapfigure}

% python main.py -a maddpg -c maddpg_conf -g particle  -d particle_conf game_name=simple_tag num_adversaries=3 good_load_model=True good_load_model_path=source/simple_tag/tag4/model_30000

Figure~\ref{fig_return_mujoco} shows the average returns on Ant-v3, Half-Cheetah-v3 and Hopper-v3, which illustrates that our deep PR method can improve the performance of many vanilla RL algorithms. 
We found that the PR method can improve the convergence speed by up to almost 5 times, namely TD3+PR on Ant-v3 and DDPG+PR on Hopper-v3. In contrast, on Half-Cheetah-v3, the improvement of the proposed PR on TD3 and DDPG algorithms is smaller. Since this environment is relatively easy, and the baseline algorithms can also quickly learn the optimal policy. In addition, when the baseline performance is inherently poor, such as DDPG on Ant-v3, it is also difficult for PR to improve its performance.
% In addition, we also found that on Half-Cheetah-v3, the improvement of both TD3 and DDPG algorithms by PR is not obvious enough. 

% This may be because these baseline algorithms can quickly learn the optimal policy in this relatively easy environment, and the performance improvements of policy reciprocity on this basis are relatively few. 

% For some algorithms with poor baseline performance, i.e., DDPG on Ant-v3, PR is sometimes hard to improve the performance of these algorithms.

% \begin{figure*}[!htbp]
% 	\centering
% % 	\hspace{-0.3cm}
%     \subfloat{
% 	\includegraphics[width=2.2in]{kappa/maddpg_spread_n3.pdf}
% 	}
% 	\subfloat{
% 	\includegraphics[width=2.2in]{kappa/maddpg_spread_n6.pdf}
% 	}
% 	\subfloat{
% 	\includegraphics[width=2.2in]{kappa/qmix_spread_n3.pdf}
% 	}
%  	\caption{Average reward of different algorithms. All results is averaged by
% five seeds. \textbf{Left:} Cooperative Navigation $|\mathcal{N}|=3$ on MADDPG. \textbf{Middle:} Cooperative Navigation $|\mathcal{N}|=6$ on MADDPG. \textbf{Right:} Simple-spread on QMIX.}
% 	\label{fig_qmix}
% \end{figure*}

\begin{figure*}[!htbp]
	\centering
% 	\hspace{-0.3cm}
    \subfloat{
	\includegraphics[width=2.2in]{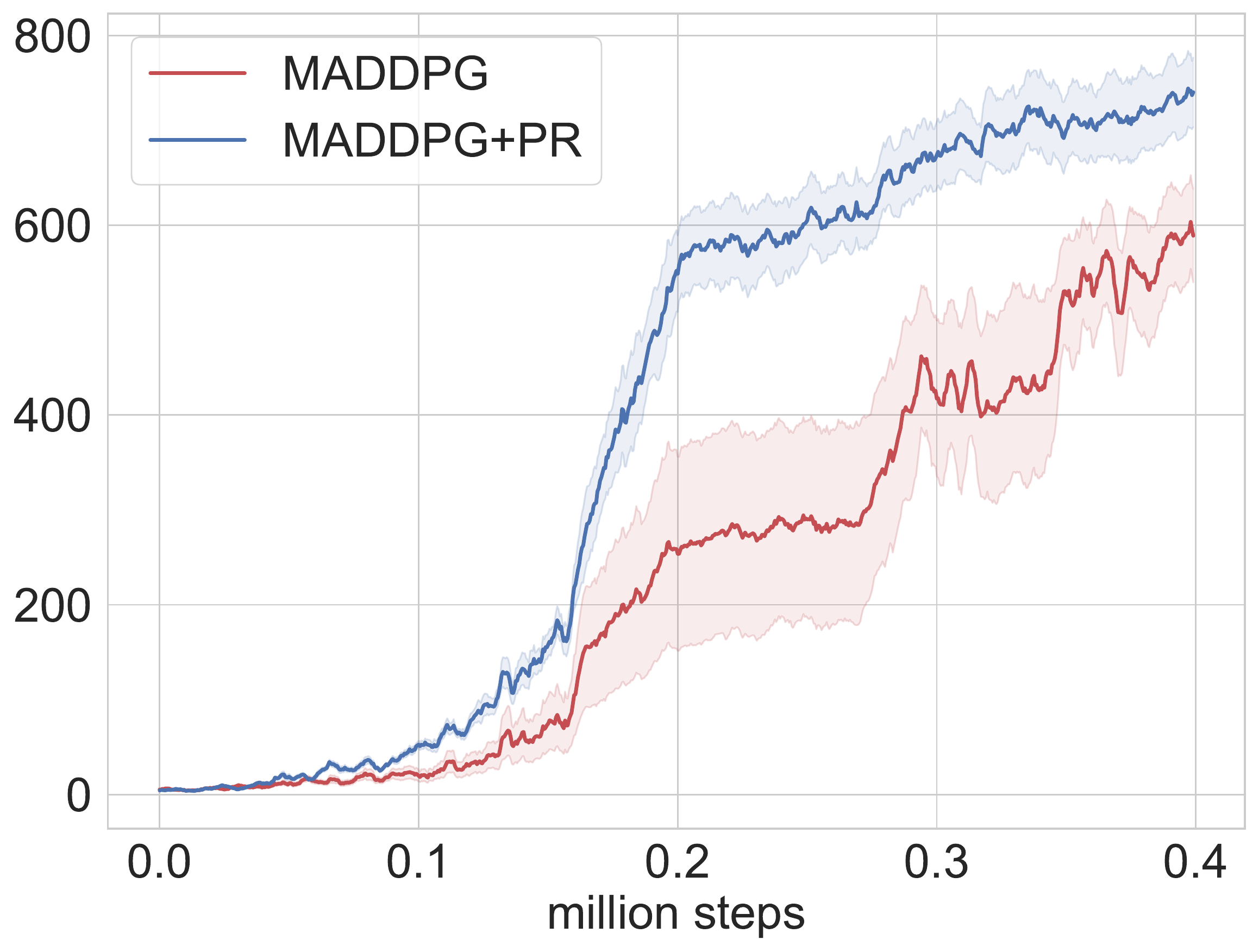}
	}
	\subfloat{
	\includegraphics[width=2.2in]{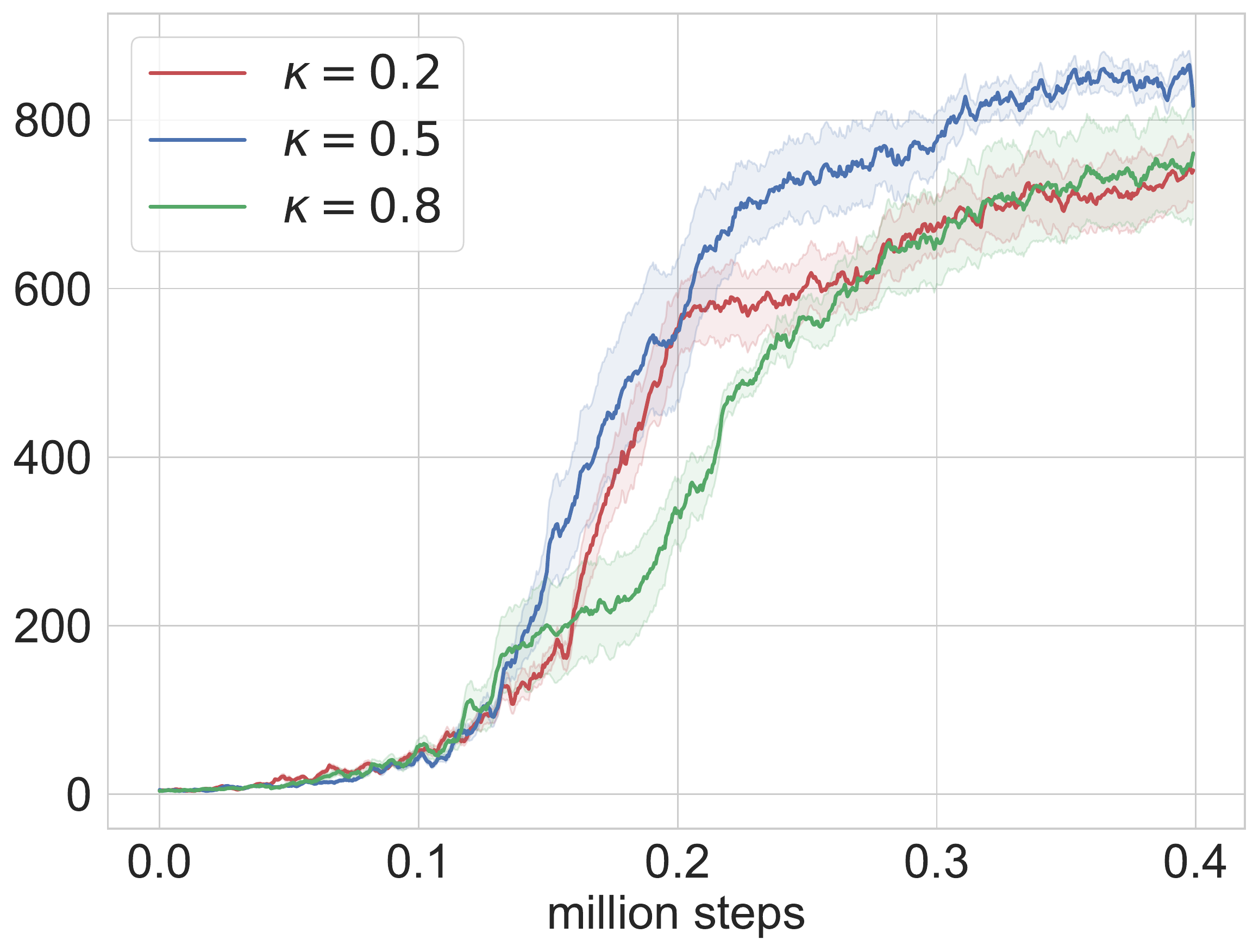}
	}
	\subfloat{
	\includegraphics[width=2.2in]{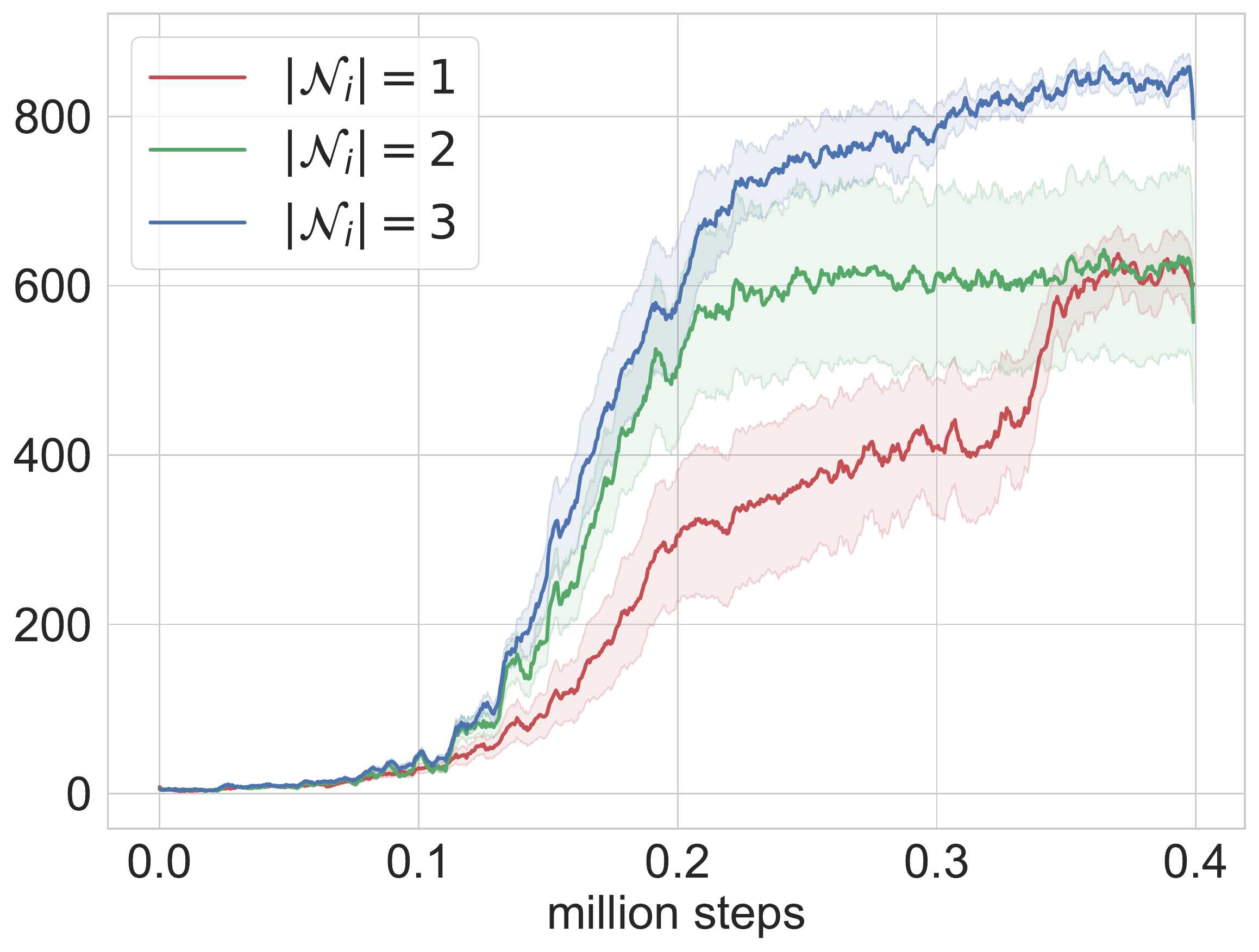}
	}
 	\caption{Effect of different Hyperparameters. All results is averaged by
five seeds. \textbf{Left:} Predator-prey $|\mathcal{N}|=4$. \textbf{Middle:} Effect of $\kappa$. \textbf{Right:} Effect of $|\mathcal{N}_i|$.}
	\label{fig_parameters}
\end{figure*}

\begin{figure}[!tbp]
	\centering
% 	\hspace{-0.3cm}
    \subfloat{
	\includegraphics[width=2.5in]{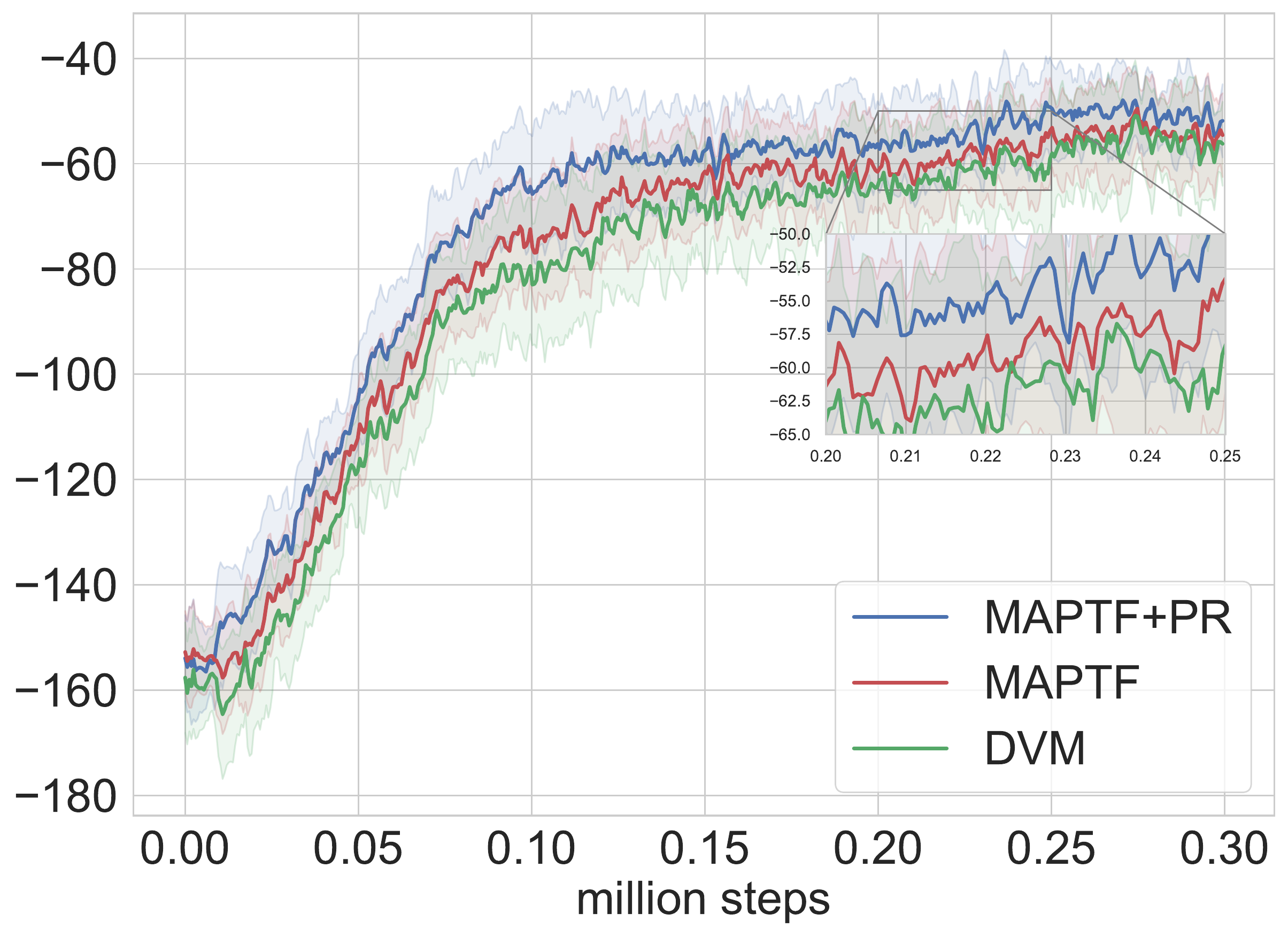}
	}
	
	\subfloat{
	\includegraphics[width=2.5in]{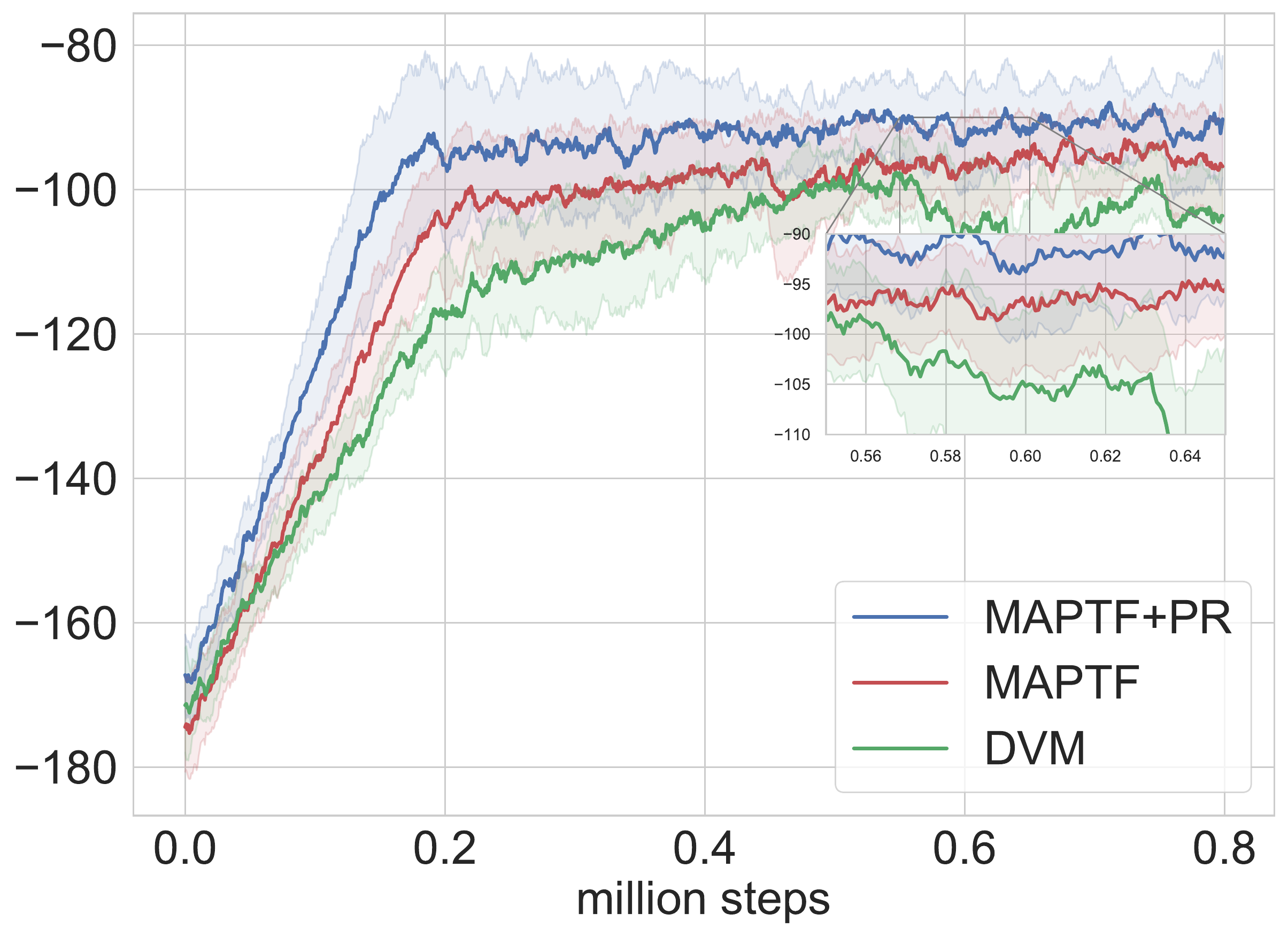}
	}

 	\caption{Comparison results with transfer RL algorithms on Simple-spread environment. All results is averaged by five seeds. \textbf{Upper:} $|\mathcal{N}|=3$. \textbf{Low:} $|\mathcal{N}|=6$. }
	\label{fig_tr}
\end{figure}

% \begin{wrapfigure}[12]{r}{0.59\textwidth}
% \vspace{-14pt}
% 	\centering
% % 	\hspace{-0.3cm}
%     % \subcaption{Predator-prey $|\mathcal{N}|=3$ | $|\mathcal{N}|=6$]}
%     % \subfloat[]{
% 	\includegraphics[width=1.6in]{vis_fig/maptf3.pdf}
% % 	}
%     \includegraphics[width=1.6in]{vis_fig/maptf6.pdf}

%  	\caption{Comparison results with transfer RL algorithms on Simple-spread environment. All results is averaged by five seeds. \textbf{Left:} $|\mathcal{N}|=3$. \textbf{Right:} $|\mathcal{N}|=6$. }
% 	\label{fig_tr}
% \end{wrapfigure}
Figure~\ref{fig_deep} validate the effectiveness of  our deep PR on several MPE and StarCraft environments. 
Compared with the vanilla MADDPG, MAPPO and QMIX algorithms, the deep PR method can significantly improve their performance. In the case of achieving the same performance, the sample complexity can be greatly reduced, and the final performance is significantly improved, especially in the Predator-prey $|\mathcal{N}|=6$ environment.

% Figure~\ref{fig_qmix} presents additional experiments, including the performance of different algorithms on MPE, further illustrating the effectiveness of the PR approach. 
% As shown in Figure~\ref{fig_qmix}-Middle, we found that when the episode length is set to 100, the average reward obtained is relatively unstable, which may be caused by the larger randomness in the early training stage due to the long horizon.

Figure~\ref{fig_parameters} shows the effect of Hyperparaments on the Predator-prey environment. As shown in Figure~\ref{fig_parameters}(a), we choose the number of agents as $4$ and found that PR outperforms the baseline.  
Figure~\ref{fig_parameters}(b) presents the effect of the weight factor $\kappa$. When $\kappa=0.5$, we can obtain the highest return. 
However, other cases, e.g., $\kappa=0.2$ or $\kappa=0.8$, will degrade performance. This means that $\kappa$ should be chosen reasonably; too large may cause the agent to ignore its perceived reward; on the contrary, the agent may fall into a local optimum. 
Figure~\ref{fig_parameters}(c) illustrates the effect of the number of agents used for PR. 
% We found that the higher the number of agents, the better the performance. As the number of agents participating interaction decreases, the performance gradually approaches the vanilla MADDPG in Figure~\ref{fig_parameters}(a).
We found that the higher the number of agents participating in the PR, the better the performance.
As the number of agents decreases, the performance gradually approaches the vanilla MADDPG in Figure~\ref{fig_parameters}(a).

Figure~\ref{fig_tr} shows the comparison results with MAPTF \cite{yang2021efficient} and DVM \cite{wadhwania2019policy} on Simple-spread environment, indicating that PR has the ability to further improve the performance of these transfer RL algorithms. We conduct experiments with different numbers of agents. When the number of agents is small, the information that agents can help each other is limited, hence the improvement of our PR algorithm is not obvious. However, when the number of agents increases, the performance improvement is more pronounced.
\appendix
\section{Proof of Main Results}

\subsection{Some Existing Results}
\begin{lemma}[Proposition 4.1 in \cite{kar2013cal}]\label{L2}
Let $\left\{z_t\right\}$ be real-valued and deterministic with 
\begin{equation}
    z_{t+1} \leq (1-\alpha_t)z_t + \alpha_t \varepsilon_t,
\end{equation}
where the deterministic sequences $\left\{\alpha_t\right\}$ and $\left\{\varepsilon_t\right\}$ satisfy $\alpha_t \in \left[0,1\right]$, $\sum_{t\geq 0} \alpha_t = \infty$, and there exists a constant $R>0$, such that
\begin{equation}
    \lim \sup_{t \rightarrow \infty}\varepsilon_t < R.
\end{equation}
Then $\lim \sup_{t \rightarrow \infty} z_t \leq R$.
\end{lemma}

\begin{lemma}[Lemma 4.3 in \cite{kar2013distributed}]\label{L5}
Let $\left\{z_t\right\}$ be an $\mathbb{R}_{+}$ valued $\left\{\mathcal{H}_t\right\}$ adapted process that satisfies 
\begin{equation}
    z_{t+1} \leq (1-r_1(t))z_t + r_2(t)(U_t+J_t).
\end{equation}
In the above, $\left\{r_1(t)\right\}$ is an $\left\{\mathcal{H}_t\right\}$ adapted process, such that, for all $t$, $r_1(t)$ satisfies $0 \leq r_1(t) \leq 1$ and
\begin{equation}
    \frac{a_1}{(t+1)^{\delta_1}} \leq \mathbb{E} \left[r_1(t)|\left\{\mathcal{H}_t\right\}\right] \leq 1
\end{equation}
with $a_1 > 0$ and $0 \leq \delta_1 \leq 1$, whereas, the sequence $\left\{r_2(t)\right\}$ is deterministic, $\mathbb{R}_{+}$ valued, and satisfies $r_2(t) \leq a_2/(t+1)^{\delta_2}$, with $a_2 >0$ and $\delta_2 > 0$. Further, let $\left\{U_t\right\}$ and $\left\{J_t\right\}$ be $\mathbb{R}_{+}$ valued $\left\{\mathcal{H}_t\right\}$ and $\left\{\mathcal{H}_{t+1}\right\}$ adapted processed respectively with $\sup_{t \geq 0}\left\|U_t\right\| < \infty$ and $\left\{J_t\right\}$ is i.i.d for each $t$ and satisfies the condition $\mathbb{E}\left[\left\|J_t\right\|^{2+\epsilon_1}\right]<\infty$. Then for every $\delta_0$ such that
\begin{equation}
    0 \leq \delta_0 <\delta_2-\delta_1-\frac{1}{2+\epsilon_1},
\end{equation}
we have $(t+1)^{\delta_0}z_t \rightarrow 0$ as $t \rightarrow \infty$.
\end{lemma}

\begin{lemma}[Lemma 4.4 in \cite{kar2013distributed}]\label{L7}
For positive integers $M$ and $N$, we define the consensus subspace $\mathcal{C}$ of $\mathbb{R}^{MN}$ as the following,
$$\mathcal{C}=\left\{\mathbf{y} \in \mathbb{R}^{MN}: \mathbf{y}=1_{N} \otimes \mathbf{y}' \ \emph{\text{for}} \ \mathbf{y}'\in \mathbb{R}^N \right\},$$
where $\otimes$ denotes the Kronecker product. 
Let $\mathcal{C}^{\perp}$ be the orthogonal complement of $\mathcal{C}$, such that for any $\mathbf{y} \in \mathbb{R}^{MN}$, $\mathbf{y}=\mathbf{y}_{\mathcal{C}}+\mathbf{y}_{\mathcal{C}^{\perp}}$, where $\mathbf{y}_{\mathcal{C}}$ is the consensus subspace projection of $\mathbf{y}$.

Let $\left\{\mathbf{z}_t\right\}$ be an $\mathbb{R}^{NP}$ valued $\left\{\mathcal{H}_t\right\}$ adapted process such that $\mathbf{z}_t \in \mathcal{C}^{\perp}$ for all $t$. Also, let $\left\{L_t\right\}$ be an i.i.d. sequence of graph Laplacian matrices that satisfies
\begin{equation}
    \lambda_{2}(\bar{L})=\lambda_{2}\left(\mathbb{E}\left[L_{t}\right]\right)>0,
\end{equation}
with $L_t$ being $\mathcal{H}_{t+1}$ adapted and independent of $\mathcal{H}_t$ for all $t$. Then, there exists a measurable $\left\{\mathcal{H}_{t+1}\right\}$ adapted process $\left\{r_t\right\}$ and a constant $c_r > 0$, such that $0 \leq r_t \leq 1$ a.s. and 
\begin{equation}
\begin{aligned}
\left\|\left(I_{N P}-\bar{r}_{t} L_{t} \otimes I_{P}\right) \mathbf{z}_{t}\right\| & \leq\left(1-r_{t}\right)\left\|\mathbf{z}_{t}\right\| \\
\mathbb{E}\left[r_{t} \mid \mathcal{H}_{t}\right] & \geq \frac{c_{r}}{(t+1)^{\delta}} \; \text{a.s.}
\end{aligned}
\end{equation}
for all $t$, where the weight sequence $\left\{\bar{r}_t\right\}$ satisfies $\bar{r}_t \leq \bar{r}/(t+1)^{\delta}$ for some $\bar{r}>0$ and $\delta \in \left(0,1\right]$.
\end{lemma}

\begin{lemma}[Lemma 5.4 in \cite{kar2013cal}]\label{L9}
For each state-action pair $(s,a)$, let $\left\{z_{s,a}(t)\right\}$ denote the $\left\{\mathcal{F}_t\right\}$ adapted read-valued process evolving as
\begin{equation}
 z^{t+1}(s, a)=\left(1-\alpha_t(s, a)\right) z^t(s, a)+\alpha_t(s, a)\left(\bar{\nu}^t(s, a)+\bar{\varepsilon}^t(s, a)\right),
\end{equation}
where the weight sequence $\{\alpha_t(s, a)\}$ is given in \eqref{eq_weight}, and $\left\{\bar{\nu}^t(s, a)\right\}$ is an $\left\{\mathcal{F}_{t+1}\right\}$ adapted process satisfying $\mathbb{E}\left[\overline{\nu}^t(s,a)\mid \mathcal{F}_{t}\right]=0$ for all $t$ and
\begin{equation}
    \mathbb{P}\left(\sup _{t \geq 0} \mathbb{E}\left[(\overline{\nu}^t)^2(s, a) \mid \mathcal{F}_{t}\right]<\infty\right)=1.
\end{equation}
Further, the process $\left\{\bar{\varepsilon}^t(s, a)\right\}$ is $\left\{\mathcal{F}_t\right\}$ adapted, such that, $\bar{\varepsilon}^t(s, a) \rightarrow 0$ as $t \rightarrow \infty$ a.s. Then, we have $z^t(s, a) \rightarrow 0$ as $t \rightarrow \infty$ a.s.
\end{lemma}

\section{Proof of Main Results}
\subsection{Proof of Theorem~\ref{Lemma_1}}
\begin{proof}
To proof the boundedness property, we first define some useful notations and properties. 

Define an operator as 
\begin{equation}\label{L1_5}
    \mathcal{G}_i(Q_i(s,a)) = \mathbb{E}\left[r_i(s,a)\right] + \gamma \sum_{s^{\prime} \in \mathcal{S}} p_{s,s'}^a \max_{a' \in \mathcal{A}_i}Q_i(s^{\prime},a^{\prime}),   
\end{equation}
and the residual term $v_i(s,a)$ as
\begin{equation}\label{L1_6}
v_i(s,a) = r_i(s,a) + \gamma \max_{a^{\prime}\in \mathcal{A}_i}Q_i^{t}(s^{\prime},a^{\prime})-\mathcal{G}_i(Q_i(s,a)),
\end{equation}
which plays the role of a martingale difference noise, i.e., $\mathbb{E}\left[v_i(s,a) | \mathcal{F}_t\right] = 0$ for all $t$ \cite{wei2006time}. And there exists positive constants $c_0, c_1$ and $c_2$, such that
\begin{equation}\label{L1_7}
\begin{split}
    \mathbb{E}\left[\left\|v_i(s,a)\right\|^2 | \mathcal{F}_t\right] & \leq (r_i(s,a)-\mathbb{E}[r_i(s,a)])^2 + c_0 \left\|Q_i^t\right\|_{\infty}^2 \\
    & \overset{(a)}{\leq} c_1 + c_2 \left\|\mathcal{Q}^t\right\|^2,
\end{split}
\end{equation}
where $\mathcal{Q}^t \in \mathbb{R}^{N \times |\mathcal{S}| \times  |\mathcal{A}|}$ includes all agents' $Q$-values, $(a)$ is based on Assumption~\ref{assumption_mm} and the fact $\|Q_i^t\|_{\infty}^2 \leq \|\mathcal{Q}^t\|^2$.

Since the one stage reward $r_i(s,a)$ is boundedness based on Assumption~\ref{assumption_mm}, for each agent $i$, there exists a positive constant $c_3$ such that the operator $\mathcal{G}_i(Q_i(s,a))$ in \eqref{L1_5} satisfies
\begin{equation}
    \left|\mathcal{G}_i(Q_i(s,a))\right| \leq c_3 + \gamma \left\|Q_i\right\|_{\infty},
\end{equation}
where $Q_i \in \mathbb{R}^{|\mathcal{S}_i|\times|\mathcal{A}_i|}$. 
Thus, there exists $\hat{\gamma} \in [0,1)$ and a constant $J > 0$, such that
\begin{equation}\label{e_J}
    \left|\mathcal{G}_i(Q_i(s,a))\right| \leq \hat{\gamma} \max \left(\left\|Q_i\right\|_{\infty}, J\right).
\end{equation}
Also, let $\hat{\varepsilon}$ be another constant such that $\hat{\gamma}(1+\hat{\varepsilon})=1$.

Based on Assumption~\ref{assumption_matrix}, $Q_{\star, i}$ includes all agents' information, i.e.,
\begin{equation}
\begin{split}
Q_{\star}^t(s,a) &= Q_{\star,i}^t(s,a) \\
&= \frac{\kappa}{|\mathcal{N}|} \sum_{i \in \mathcal{N}} Q_i^t(s,a) + \frac{1-\kappa}{|\mathcal{N}|} \sum_{i \in \mathcal{N}} Q_{\sharp,i}^t(s,a),
\end{split}
\end{equation}
therefore, 
we can rewrite \eqref{e2} as
\begin{equation}\label{L1_4}
\begin{split}
    Q_i^{t+1}(s,a) =&~ [1-\alpha_t(s,a)]Q_i^{t}(s,a) +\\ 
    & \alpha_t(s,a)\big[ r_i(s,a) + \gamma \max_{a^{\prime} \in \mathcal{A}_i} Q_i^{t}(s^{\prime},a^{\prime})\big] \\
    &+ \beta_t(s,a) (Q_{\star}^t(s,a)-Q_i^{t}(s,a)).
\end{split}
\end{equation}

Substitute \eqref{L1_5}, \eqref{L1_6} into \eqref{L1_4} and transform it as a vector form, we obtain
\begin{align}\label{L1_8}
    & \bm{q}^{t+1}(s,a) = \left(I_N-\frac{\beta_t(s,a)}{N} \kappa L_t-\alpha_t(s,a) I_N\right)\bm{q}^{t}(s,a) \nonumber \\
    &~~~ + \alpha_t(s,a) \left(\mathcal{G}(\bm{q}^t(s,a))+\bm{v}(s,a)\right) \nonumber \\
    &~~~ + \beta_t(s,a) (1-\kappa) \left(\frac{1}{N}1_{N \times N} \bm{q}_{\sharp}^t(s,a)-I_N\bm{q}^{t}(s,a)\right),
\end{align}
where $I_N$ is the identity matrix, $1_{N \times N}$ is all $1$ matrix with size $N \times N$, $L_t$ is a Laplacian matrix, and 
\begin{equation}
\begin{split}
\bm{q}^{t}(s,a) &= [Q_{1}^t(s,a),\cdots,Q_{N}^t(s,a)]^T \in \mathbb{R}^N \\
\bm{q}_{\sharp}^t(s,a) &= [Q_{\sharp, 1}^t(s,a), \cdots, Q_{\sharp, N}^t(s,a)]^T \in \mathbb{R}^N \\
\mathcal{G}(\bm{q}^t(s,a)) &= [\mathcal{G}_1(Q_1^t(s,a)),\cdots, \mathcal{G}_N(Q_{N}^t(s,a))]^T \in \mathbb{R}^N \\
\bm{v}(s,a) &= [v_1(s,a),\cdots,v_N(s,a)]^T \in \mathbb{R}^N.
\end{split}
\end{equation}

To bound $\mathcal{Q}^t$, we first construct an adapted process $\left\{M_t\right\}$, given by
\begin{equation}\label{L1_9}
    M_t=\max_{t^{\prime} \leq t} \Big\|\mathcal{Q}^{t'}\Big\|_{\infty},\ \forall t.
\end{equation}
Let $\left\{J_t\right\}$ be another adapted process, and for any $t>0$, $J_t=J_{t-1}$ if $M_t \leq (1+\hat{\varepsilon})J_{t-1}$; otherwise, 
if $M_t \geq (1+\hat{\varepsilon})J_{t-1}$, $J_t$ is defined by $J_0(1+\hat{\varepsilon})^k$, where $k>0$ satisfying
\begin{equation*}
   J_0(1+\hat{\varepsilon})^{k-1} \leq M_t \leq J_0(1+\hat{\varepsilon})^k.
\end{equation*}
The above construction makes the following is hold,
\begin{equation}\label{L1_10}
    \begin{split}
        M_t &\leq (1+\hat{\varepsilon})J_t, \  \forall t \geq 0 \\
        M_t &\leq J_t, \ \text{if} \  J_{t-1}< J_t.
    \end{split}
\end{equation}

Next we will use the proof by contradiction to show the boundedness of $\mathcal{Q}^t$. 
If $\mathcal{Q}^t$ is not bounded a.s., there exists an event $\mathcal{B}$ of positive measure, such that $M_t \rightarrow \infty$ as $t \rightarrow \infty$ on $\mathcal{B}$.
To set up a contradiction argument, for each state-action pair $(s,a)$, we define a process $\{\mathbf{z}^{t:t_0}(s,a)\}_{t \geq t_0}$ evolves as
\begin{equation}\label{L1_11}
\begin{split}
    \mathbf{z}^{t+1: t_{0}}(s, a)&=\left(I_{N}-\frac{\beta_{t}(s,a)}{N} \kappa L_{t}-\alpha_{t}(s,a) I_{N}\right) \mathbf{z}^{t:t_0}(s,a) \\
    &+ \beta_t(s,a)(1-\kappa) \bar{\Delta}_t + \alpha_t(s,a) \bar{\bm{v}}^t(s, a),
\end{split}
\end{equation}
the term $\bar{\Delta}_t$ in \eqref{L1_11} can be rewritten as  $\bar{\Delta}_t = \Delta_t/J_t= (\frac{1}{N}\mathbf{1}_{N \times N} \bm{q}_{\sharp}^t(s,a)-I_N\bm{q}^{t}(s,a))/J_t$, which satisfies 
\begin{equation*}
    \begin{split}
    \mathbb{E}\left[\left\|\bar{\Delta}_t\right\|^2\right] &= \frac{1}{J_t^2}\mathbb{E}\left[\left\|\Delta_t\right\|^2\right] \\ & \overset{(a)}{\leq}
        \frac{\|\frac{1}{N}1_{N\times N}\bm{q}_{\sharp, i}^t(s,a)\|^2+\|I_N\bm{q}^{t}(s,a)\|^2}{J_t^2} \\
        % \frac{2\left\|\mathcal{Q}_t\right\|^2}{J_t^2} 
        & \overset{(b)}{\leq} \frac{c_4M_t^2}{J_t^2} \leq c_4(1+\hat{\varepsilon})^2,
    \end{split}
\end{equation*}
where $(a)$ and $(b)$ is based on the triangle inequality and \eqref{L1_9}, respectively, $c_4$ is a positive constant.

The term $\bar{\bm{v}}^t(s, a) = \bm{v}^t(s, a) / J_t$, which satisfies $\mathbb{E}\left[\bar{\bm{v}}^t(s, a)|\mathcal{F}_t\right]=\mathbf{0}$ and
\begin{equation*}
\begin{split}
    \mathbb{E}\left[\left\|\bar{\bm{v}}^t(s, a)\right\|^2\Big|\mathcal{F}_t\right] &= \frac{1}{J_t^2} \mathbb{E}\left[\left\|\bm{v}^t(s, a)\right\|^2\Big|\mathcal{F}_t\right]
    \overset{(a)}{\leq} \frac{c_1}{J_t^2}+\frac{c_2\left\|\mathcal{Q}^t\right\|^2}{J_t^2} \\
    & \overset{(b)}{\leq} \frac{c_1}{J_0^2}+\frac{c_2 M_t^2}{J_t^2} \leq \frac{c_1}{J_0^2}+ c_2(1+\hat{\varepsilon})^2 \leq c_5,
\end{split}
\end{equation*}
where $(a)$ is based on \eqref{L1_7}, $(b)$ is based on $J_0 \leq J_t$ and \eqref{L1_9}, and $c_5$ is an another positive constant.

Our goal is to bound $\bm{q}^{t}(s,a)$ by $\{\mathbf{z}^{t:t_0}(s,a)\}$. Before that, we first give the following lemmas.

\begin{lemma}\label{L3}
For each state-action pair $(s,a)$, let $\left\{\mathbf{z}^t(s,a)\right\}$ denote the $\left\{\mathcal{F}_t\right\}$ adapted process evolving as
\begin{equation}
    \mathbf{z}^{t+1}(s,a) = \left(I_N-\beta_t L_t-\alpha_t I_N\right)\mathbf{z}^t(s,a)+ \tilde{\beta}_t\bar{\Delta}_t + \alpha_t \bar{\bm{v}}^t(s,a), 
\end{equation}
where $\{\bar{\bm{v}}^t(s,a)\}$ satisfies $\mathbb{E}[\bar{\bm{v}}^t(s,a)|\mathcal{F}_t]=0$, for all $t$ and
\begin{equation}
    \sup_{t\geq 0} \mathbb{E}\left[\left\|\bar{\bm{v}}^t(s,a)\right\||\mathcal{F}_t\right] < R,
\end{equation}
where $R$ is a constant. The weight sequence $\{\beta_t\}$, $\{\tilde{\beta}_t\}$ and $\{\alpha_t\}$ satisfy the condition of $\{\beta_t\}$ and $\{\alpha_t\}$ in \eqref{eq_weight}, respectively, and $\bar{\Delta}_t$ is an another random process, also bounded by $R$,
then, for any constant $\varepsilon>0$, there exists a random time $t_{\varepsilon}$, such that $\mathbf{z}^{t_{\varepsilon}}(s,a) \leq \varepsilon$.
\end{lemma}

\begin{lemma}\label{L4}
For each state-action pair $(s,a)$ and $t_0 \geq 0$, consider the process $\left\{\mathbf{z}^{t:t_0}(s,a)\right\}$ that evolves as 
\begin{equation}
\mathbf{z}^{t+1: t_{0}}(s, a)=\big(I_{N}-\beta_{t} L_{t}-\alpha_{t} I_{N}\big) \mathbf{z}^{t: t_{0}}(s, a)
+\tilde{\beta}_t \bar{\Delta}_t + \alpha_t \bar{\bm{v}}^t(s, a)
\end{equation}
with $\mathbf{z}^{t_{0}: t_{0}}(s, a) = \mathbf{0}$, $\bar{\Delta}_t$ and $\bar{\bm{v}}^t(s, a)$ satisfy the condition in Lemma~\ref{L3}, the weight sequence $\{\beta_t\}$, $\{\tilde{\beta}_t\}$ and $\{\alpha_t\}$ satisfy the condition in Lemma~\ref{L3}. Then, for any constant $\varepsilon>0$, there exists a random time $t_{\varepsilon}$, such that, $\left\|\mathbf{z}^{t:t_0}(s,a)\right\| \leq \varepsilon$, for all $t_{\varepsilon} \leq t_0 \leq t $.
\end{lemma}

Therefore, the evolution process $\{\mathbf{z}^{t:t_0}(s,a)\}_{t \geq t_0}$ falls the purview of Lemma~\ref{L4}, i.e., there exists $t_{\hat \varepsilon}$, such that
\begin{equation*}
    \left\|\mathbf{z}^{t:t_0}(s,a)\right\| \leq \hat{\varepsilon}
\end{equation*}
for all $t_{\hat{\varepsilon}} \leq t_0 \leq t$ and state-action pairs $(s,a)$.

In order to obtain a contradiction, we show that the following hold a.s. on $\mathcal{B}$ for all state-action pairs $(s,a)$ and $t \geq t_1$,
\begin{equation}\label{L1_12}
\bm{q}^{t}(s,a) \eqslantless  J_{t_1}\left(\mathbf{z}^{t:t_1}(s,a)+1_N\right), \ J_{t} = J_{t_1},
\end{equation}
where $\eqslantless$ denotes the pathwise inequality and $1_N$ is all $1$ vector.

\eqref{L1_12} is established by induction and holds for $t = t_1$ by construct that $\mathbf{z}^{t_1:t_1}(s,a) = \bm{0}$ and $\left\|\bm{q}^{t_1}(s,a)\right\|_{\infty} \leq M_{t_1} \leq J_{t_1}$ for all state-action pairs $(s,a)$. 
To obtain \eqref{L1_12} at the $t+1$ time slot, we first consider
\begin{equation}\label{L1_13}
\begin{split}
& \left(I_N-\frac{\beta_t(s,a)}{N} \kappa L_t-\alpha_t(s,a) I_N \right) \bm{q}^{t}(s,a) \\ 
\eqslantless &~\left(I_N-\frac{\beta_t(s,a)}{N} \kappa L_t-\alpha_t(s,a) I_N\right) \left(J_{t_1}\mathbf{z}^{t:t_1}(s,a)+J_{t_1} 1_N\right) \\
 = &~ J_{t_1} \left (I_N-\frac{\beta_t(s,a)}{N} \kappa L_t-\alpha_t(s,a) I_N \right) \mathbf{z}^{t:t_1}(s,a) \\
 &~~~~~~~~~~~~~~~~+\left(1-\alpha_t(s,a) \right)J_{t_1}1_N,
\end{split}        
\end{equation}
where the last equation use the property of the Laplacian matrix that $L_t 1_N = \mathbf{0}$. 
Next, we substitute \eqref{L1_13} into \eqref{L1_8} to obtain
\begin{equation}
\begin{split}
    & \bm{q}^{t+1}(s,a) =  \big(I_N-\frac{\beta_t(s,a)}{N} \kappa L_t-\alpha_t(s,a) I_N\big)\bm{q}^{t}(s,a) \\
    & ~~ + \beta_t(s,a) (1-\kappa) \Delta_t  + \alpha_t(s,a) \left(\mathcal{G}(\bm{q}^t(s,a))+\bm{v}_i(s,a)\right) \\
    & \eqslantless J_{t_1} \big(I_N-\frac{\beta_t(s,a)}{N} \kappa L_t-\alpha_t(s,a) I_N\big)\mathbf{z}^{t:t_1}(s,a) \\
    &  ~~ +\left(1-\alpha_t(s,a)\right)J_{t_1} 1_N + \beta_t(s,a) (1-\kappa) \Delta_t  \\
    &  ~~ + \alpha_t(s,a) \left(\mathcal{G}(\bm{q}^t(s,a))+\bm{v}^t(s,a)\right) \\
    & \overset{(a)}{\eqslantless} J_{t_1} \big(I_N-\frac{\beta_t(s,a)}{N} \kappa L_t-\alpha_t(s,a)  I_N\big)\mathbf{z}^{t:t_1}(s,a) \\
    & ~~ +\left(1-\alpha_t(s,a) \right)J_{t_1}1_N + \alpha_t(s,a) \hat{\gamma}(1+\hat{\varepsilon})J_{t_1} 1_N \\
    & ~~ +\beta_t(s,a) (1-\kappa) J_{t_1} \bar{\Delta}_t + \alpha_t(s,a) J_{t_1} \bar{\bm{v}}^t(s,a)\\
    &=J_{t_1}\mathbf{1}_N+ J_{t_1}\big[\big(I_N-\frac{\beta_t(s,a)}{N} \kappa L_t-\alpha_t(s,a)  I_N\big)\mathbf{z}^{t:t_1}(s,a)\\
    &~~~ +\beta_t(s,a) (1-\kappa) \bar{\Delta}_t + \alpha_t(s,a) \bar{\mathbf{v}}^t(s,a) \big],
\end{split}
\end{equation}
where $(a)$ is based on $\eqref{e_J}$.
Therefore, 
\begin{equation}
    \bm{q}^{t+1}(s,a) \leq J_{t_1} 1_N + J_{t_1} \mathbf{z}^{\left(t+1:t_1\right)}{(s,a)}.
\end{equation}
Note that the process $\{\mathbf{z}^{t:t_0}(s,a)\}$ is boundedness, then we can conclude that the $\bm{q}$ is boundedness. Then we complete the proof.
\end{proof}

\subsection{Proof of Theorem~\ref{Theorem_1_1}}

\begin{proof}
Recall that the process $\{\bm{q}^t(s,a)\}$ evolves as 
\begin{align}\label{t1_1}
    &\bm{q}^{t+1}(s,a) = \left(I_N-\frac{\beta_t(s,a)}{N} \kappa L_t-\alpha_t(s,a) I_N \right)\bm{q}^{t}(s,a) \nonumber \\
    &+ \alpha_t(s,a) \left(\mathcal{G}(\bm{q}^t(s,a))+\bm{v}(s,a)\right) \nonumber \\
    &+ \beta_t(s,a) (1-\kappa) \left(\frac{1}{N}\mathbf{1}_{N \times N} \bm{q}_{\sharp}^t(s,a)-I_N\bm{q}^{t}(s,a)\right), 
\end{align}
which can be rewritten as 
\begin{equation}\label{t1_2}
    \begin{split}
        & \bm{q}^{t+1}(s,a) = \left(I_N-\frac{\beta_t(s,a)}{N} \kappa L_t-\alpha_t(s,a) I_N\right)\bm{q}^{t}(s,a) \\ 
        &~~~+ \alpha_t(s,a) \left(\bm{U}_t+\bm{J}_t\right) \\
        &~~~+ \beta_t(s,a) (1-\kappa) \left(\frac{1}{N}\mathbf{1}_{N \times N} \bm{q}_{\sharp}^t(s,a)-I_N\bm{q}^{t}(s,a)\right) \\
        &= \left(I_N-\frac{\beta_t(s,a)}{N} \kappa L_t-\alpha_t(s,a) I_N\right)\bm{q}^{t}(s,a) \\ 
        &~~~+ \beta_t(s,a) (1-\kappa) \Delta_t + \alpha_t(s,a) \left(\bm{U}_t+\bm{J}_t\right),
    \end{split}
\end{equation}
where $\left\{\bm{U}_t\right\}$ and $\left\{\bm{J}_t\right\}$ are $\mathbb{R}^N$-valued processed whose $i$-th components are given by
\begin{equation}\label{t1_3}
    U_{t,i} = \gamma \max_{a \in \mathcal{A}_i}Q_i^t(s,a) \quad \text{and}  \quad J_{t,i} = r_i(s,a).
\end{equation}

Note that the process $\left\{\bm{q}^t(s,a)\right\}$ may only change at $T_{s,a}(k)$, therefore, define another process $\left\{\mathbf{z}_k\right\}$ is the randomly sampled version of $\left\{\bm{q}^t(s,a)\right\}$, which evolve as 
\begin{equation}\label{t1_4}
\begin{split}
    \mathbf{z}_{k+1} &= \left(I_N-\frac{\beta_k}{N} \kappa L_k-\alpha_k I_N\right)\mathbf{z}_{k} \\ 
    &+\beta_k (1-\kappa) \Delta_k + \alpha_k \left(\bm{U}_k+\bm{J}_k\right)
\end{split}
\end{equation}
where $\alpha_k, \beta_k, \Delta_k, \bm{U}_k $ and $\bm{J}_k$ are the sampled versions at time $T_{s,a}(k)$, respectively.

Let $\bar{\mathbf{z}}_k = \mathbf{1}_N^T\mathbf{z}_k / N$ denote the average of the components of $\mathbf{z}_k$. Therefore, by using the properties of the Laplacian matrix $L_k$ and all $1$ matrix $I_{N \times N}$, i.e., $(1/N)\mathbf{1}_N^T 1_{N \times N}\bm{q}_{\sharp, i}^t(s,a) = 1_{N \times N}\bm{q}_{\sharp, i}^t(s,a) $, the residual $\hat{\mathbf{z}}_k=\mathbf{z}_k - \bar{\mathbf{z}}_k$ evolves as 
\begin{equation}\label{t1_5}
\begin{split}
    \hat{\mathbf{z}}_{k+1} &= \left(I_N-\frac{\beta_k}{N}\kappa L_k-\alpha_k I_N\right)\hat{\mathbf{z}}_k \\ 
    &- \beta_k(1-\kappa)\hat{\mathbf{z}}_k + \alpha_k \big(\hat{\bm{U}}_k+\hat{\bm{J}}_k\big),
\end{split}
\end{equation}
where 
\begin{equation}\label{t1_6}
    \begin{split}
        \hat{\bm{U}}_k &= \left(I_N-(1/N)1_N1_N^T\right)\bm{U}_k \\
        \hat{\bm{J}}_k &= \left(I_N-(1/N)1_N1_N^T\right)\bm{J}_k.
    \end{split}
\end{equation}

Define $\mathcal{H}_k = \mathcal{F}_{T_{s,a}(k)}$ be the $\sigma$-algebra associated with $T_{s,a}(k)$, then by Lemma \ref{L7}, there exists a measurable $\left\{\mathcal{H}_{k+1}\right\}$ adapted process $\left\{\zeta_k\right\}$ and constant $c_{\zeta} > 0$, such that $0 \leq \zeta_k \leq 1$, we have
\begin{equation}\label{t1_7}
\begin{split}
    \left\|\left(I_N-\frac{\beta_k}{N} \kappa L_k - \alpha_k I_N\right)\hat{\mathbf{z}}_k\right\| &\leq \left\|\left(I_N-\frac{\beta_k}{N} \kappa L_k\right)\hat{\mathbf{z}}_k\right\| + \alpha_k  \hat{\mathbf{z}}_k \\
    &\leq (1-\zeta_k)\left\|\hat{\mathbf{z}}_k\right\| + \alpha_k \left\|\hat{\mathbf{z}}_k\right\|,
\end{split}
\end{equation}
where the process $\left\{\zeta_k\right\}$ satisfies $\mathbb{E}\left[\zeta_k|\mathcal{H}_k\right] \geq \frac{c_{\zeta}}{(k+1)^{\tau_2}}$ with a constant $c_{\zeta}>0$. There exists $k_0$ and another constant $c_6$, such that
\begin{equation}\label{t1_8}
    \left\|\left(I_N-\frac{\beta_k}{N}\kappa L_k - \alpha_k I_N \right)\hat{\mathbf{z}}_k\right\| \leq (1-c_6 \zeta_k) \left\|\hat{\mathbf{z}}_k\right\|
\end{equation}
for $k \geq k_0$.
Substitute \eqref{t1_8} into \eqref{t1_7}, we obtain
\begin{equation}\label{t1_9}
\begin{split}
    \left\|\hat{\mathbf{z}}_{k+1}\right\| &\leq (1-c_6 \zeta_k) \left\|\hat{\mathbf{z}}_k\right\| + \beta_k (1-\kappa) \left\|\hat{\mathbf{z}}_k\right\| + \alpha_k \big(\big\|\hat{\bm{U}}_k\big\|+\big\|\hat{\bm{J}}_k\big\|\big) \\ 
    & \leq (1-c_7 \zeta_k) \left\|\hat{\mathbf{z}}_k\right\| + \alpha_k \big(\big\|\hat{\bm{U}}_k\big\|+\big\|\hat{\bm{J}}_k\big\|\big),
\end{split}
\end{equation}
where $c_7$ is another constant. The process $\{\|\hat{\bm{U}}_k\|\}$ is pathwise bounded according to Theorem~\ref{Lemma_1} and $\{\|\hat{\bm{J}}_k\|\}$ is i.i.d satisfying the condition $\mathbb{E}[\|\hat{\bm{J}}_k\|^{2+\varepsilon_1}] < \infty$ from Assumption \ref{assumption_mm}.

Hence, this update process of $\left\|\hat{\mathbf{z}}_k\right\|$ falls the purview of Lemma \ref{L5}, we can conclude that $(k+1)^{\tau}\hat{\mathbf{z}}_k \rightarrow \mathbf{0}$ as $k \rightarrow \infty$ for all $\tau \in (0,\tau_1-\tau_2-1/(2+\varepsilon_1))$. In particular, $\hat{\mathbf{z}}_k \rightarrow \mathbf{0}$ as $k \rightarrow \infty$, we obtain
\begin{equation}\label{t1_10}
    \mathbb{P}\left(\lim_{t \rightarrow \infty}\left|Q_i^t(s,a) - \bar{Q}^t(s,a)\right|=0\right) = 1, \forall i, s, a,
\end{equation}
where $\bar{Q}^t(s,a) = (1/N)\mathbf{1}_N^T \bm{q}^t(s,a) = \frac{1}{N}\sum_i Q_i^t(s,a)$. Then we finish the proof.
\end{proof}

\subsection{Proof of Corollary~\ref{Theorem_1_2}}

\begin{proof}
Consider only the same state-action interaction case, i.e., $\kappa=1$, which is denoted by $\tilde{Q}$.

In this case, $\tilde{Q}_{\star}(s,a)$ update as
\begin{equation}\label{t1_11}
    \tilde{Q}_{\star}^{t+1}(s,a) = \frac{1}{N}\sum_i \tilde{Q}_{i}^t(s,a),
\end{equation}
and then the agent updates can be written as 
\begin{align}\label{t1_12}
    \tilde{\bm{q}}^{t+1}(s,a) &= \left(I_N-\frac{\beta_t(s,a)}{N} L_t-\alpha_t(s,a)I_N\right)\tilde{\bm{q}}^t(s,a) \nonumber \\ 
    & ~~~ +\alpha_t(s,a) \left(\mathcal{G}(\tilde{\bm{q}}^t(s,a))+\tilde{\bm{v}}^t(s,a)\right) \nonumber \\
    &= \left(I_N-\frac{\beta_t(s,a)}{N} L_t-\alpha_t(s,a)I_N\right)\tilde{\bm{q}}^t(s,a) \nonumber \\ 
    & ~~~ +\alpha_t(s,a) \big(\tilde{\bm{U}}_t+\tilde{\bm{J}}_t\big)
\end{align}
where $\tilde{\bm{q}}^{t}(s,a) = [\tilde{Q}_{1}^t(s,a),\cdots,\tilde{Q}_{N}^t(s,a)]^T$, 
$\tilde{\bm{U}}_t$ and $\tilde{\bm{J}}_t$ are the corresponding version of $\bm{U}_t$ and $\bm{J}_t$ when $\kappa=1$, respectively.

Let $\left\{\mathbf{z}_k\right\}$ be a sampled version of sequence $\bm{q}^{t}(s,a) - \tilde{\bm{q}}^{t}(s,a)$, combine (42) and \eqref{t1_12} we have
\begin{equation}\label{t1_13}
\begin{split}
    & \mathbf{z}_{k+1} = \left(I_N-\frac{\beta_k}{N}\kappa L_k - \alpha_k I_N\right)\mathbf{z}_{k} + \alpha_k \left( \hat{\bm{U}}_k+ \hat{\bm{J}}_k\right) \\
    & +\beta_k(1-\kappa)\underbrace{\left[\frac{1}{N}\mathbf{1}_{N\times N} \bm{q}_{\sharp}^k(s,a)-I_N \bm{q}^k(s,a)+\frac{1}{N}L_k \tilde{\bm{q}}^k(s,a) \right] }_{\bm{\Xi}^k} ,
\end{split}
\end{equation}
where $\hat{\bm{U}}_k = \bm{U}_{k} - \tilde{\bm{U}}_{k}$ and $\hat{\bm{J}}_k = \bm{J}_{k} - \tilde{\bm{J}}_{k}$. 
Now we focus on the last term $\bm{\Xi}^k$ in~\eqref{t1_13}, after pathwise analysis, each component of this term can be rewritten as
\begin{equation}\label{t1_14}
    \Xi^k_i = \frac{1}{N}\sum_i Q_{\sharp,i}^k(s,a)-Q_{i}^k(s,a)-\frac{1}{N}\sum_i \tilde{Q}^k_{i}(s,a)+\tilde{Q}_{i}^k(s,a).
\end{equation}

Substitute \eqref{t1_14} into \eqref{t1_13}, we obtain that
\begin{equation}\label{t1_15}
\begin{split}
    &~~ \mathbf{z}_{k+1} = \left(I_N-\frac{\beta_k}{N}\kappa L_k - \alpha_k I_N\right)\mathbf{z}_{k} \\
    &+\frac{\beta_k(1-\kappa)}{N}\left(I_{N \times N}\bm{q}_{\sharp}^k(s,a)-I_{N \times N}\tilde{\bm{q}}^t(s,a)\right) \\ 
    &- \beta_k(1-\kappa)\mathbf{z}_{k} + \alpha_k \left(\hat{\bm{U}}_k+\hat{\mathbf{J}}_k\right).
\end{split}
\end{equation}

Similar with \eqref{t1_8}, there exists $k_0$
\begin{equation}
    \left\| \left (I_N-\frac{\beta_k}{N}\kappa L_k - \alpha_k I_N \right ) \mathbf{z}_k \right \| \leq (1-c_{8} \zeta_k) \left\|\mathbf{z}_k\right\|
\end{equation}
for $k \geq k_0$, where $\left\{\zeta_k\right\}$ is an adapted process satisfying $\mathbb{E}\left[\zeta_k|\mathcal{H}_k\right] \geq \frac{c_{\zeta}}{(k+1)^{\tau_2}}$ with a constant $c_{\zeta}>0$ and $c_{8}$ is another constant.

For the third term in \eqref{t1_15}, there exists another constant $c_{9}$, such that
\begin{align}
    & \beta_k(1-\kappa)\left\|I_{N \times N}\bm{q}_{\sharp}^k(s,a)-I_{N \times N}\tilde{\bm{q}}^k(s,a)\right\| \nonumber \\
    &\leq \beta_k(1-\kappa)\big(\left\|I_{N \times N}\bm{q}_{\sharp}^k(s,a)-I_{N \times N}\bm{q}^k(s,a)\right\| \nonumber \\
    &~~~~+ \left\|I_{N \times N}\bm{q}^k(s,a)-I_{N \times N}\tilde{\bm{q}}^k(s,a)\right\| \big) \nonumber \\
    &\overset{(a)}{\leq} \beta_k(1-\kappa) \Delta_k + \frac{\beta_k(1-\kappa)c_{9}}{N} \left\|\mathbf{z}_k\right\|,
\end{align}
where $(a)$ uses that $\Delta_k = \|I_{N \times N}\bm{q}_{\sharp}^k(s,a)-I_{N \times N}\bm{q}^k(s,a)\|$ and Jensen's inequality.
Above all, we have
\begin{align}
    &\left\|\mathbf{z}_{k+1}\right\| \leq (1-c_{8} \zeta_k)\left\|\mathbf{z}_k\right\|+\beta_k(1-\kappa)(c_{9}+1)\left\|\mathbf{z}_k\right\| \nonumber  \\ 
    & ~~~~~~ + \beta_k(1-\kappa)\Delta_k+ \alpha_k\left(\big\|\hat{ \bm{U}}_k\big\|+\big\|\hat{\bm{J}}_k\big\|\right) \\
    &\leq \left(1-c_{10}\zeta_k\right)\left\|\mathbf{z}_k\right\|+ \beta_k(1-\kappa)\Delta_k +\alpha_k\left(\big\| \hat{\bm{U}}_k\big\|+\big\|\hat{\bm{J}}_k\big\|\right), \nonumber 
\end{align}
there exists $c_{10}<c_{8}-(1-\kappa)(c_9+1)$
such that the last inequality holds. 
Note that the process $\{\|\hat{\bm{U}}_k\|\}$ is pathwise bounded according to Theorem~\ref{Lemma_1} and $\{\|\hat{\bm{J}}_k\|\}$ is i.i.d. and satisfies the condition $\mathbb{E}[\|\hat{\bm{J}}_k\|^{2+\varepsilon_1}] < \infty$ from Assumption \ref{assumption_mm}. 
Hence, this update process falls under the purview of the following Lemma \ref{L8}.
\begin{lemma}\label{L8}
Let $\left\{z_t\right\}$ be an $\mathbb{R}_{+}$ valued $\left\{\mathcal{H}_t\right\}$ adapted process that satisfies 
\begin{equation}
    z_{t+1} \leq (1-r_1(t))z_t + r_1(t)\Delta_t + r_2(t)(U_t+J_t).
\end{equation}
In the above, $\left\{r_1(t)\right\}$ is an $\left\{\mathcal{H}_t\right\}$ adapted process, such that, for all $t$, $r_1(t)$ satisfies $0 \leq r_1(t) \leq 1$ and
\begin{equation}
    \frac{a_1}{(t+1)^{\delta_1}} \leq \mathbb{E} \left[r_1(t)|\left\{\mathcal{H}_t\right\}\right] \leq 1
\end{equation}
with $a_1 > 0$, $0 \leq \delta_1 \leq 1$ and $2\delta_1 > \delta_2-\frac{1}{2+\varepsilon_1}$, whereas, the sequence $\left\{r_2(t)\right\}$ is deterministic, $\mathbb{R}_{+}$ valued, and satisfies $r_2(t) \leq a_2/(t+1)^{\delta_2}$, with $a_2 >0$ and $\delta_2 > 0$. 
$\Delta_t$ is a bounded process.
Further, let $\left\{U_t\right\}$ and $\left\{J_t\right\}$ be $\mathbb{R}_{+}$ valued $\left\{\mathcal{H}_t\right\}$ and $\left\{\mathcal{H}_{t+1}\right\}$ adapted processed respectively with $\sup_{t \geq 0}\left\|U_t\right\| < \infty$ and $\left\{J_t\right\}$ is i.i.d for each $t$ and satisfies the condition $\mathbb{E}\left[\left\|J_t\right\|^{2+\epsilon_1}\right]<\infty$. Then for every $\delta_0$ such that
\begin{equation}
    0 \leq \delta_0 <\delta_2-\delta_1-\frac{1}{2+\varepsilon_1},
\end{equation}
we have $(t+1)^{\delta_0}z_t \rightarrow 0$ as $t \rightarrow \infty$.
\end{lemma}
Then we can obtain that $(k+1)^{\tau}\mathbf{z}_k \rightarrow \mathbf{0}$ as $k \rightarrow \infty$ for all $\tau \in \left(0, \tau_1-\tau_2-1/(2+\varepsilon_1)\right)$. In particular, $\mathbf{z}_k \rightarrow \infty$ as $k \rightarrow \infty$, we have
\begin{equation}
    \mathbb{P}\left(\lim_{t\rightarrow \infty}\left|Q_{i}^t(s,a)-\tilde{Q}_{i}^t(s,a)\right|\right), \forall i, s, a.
\end{equation}
Then we compete the proof.
\end{proof}

\subsection{Proof of Theorem~\ref{Theorem_2}}

\begin{proof}
To prove the main result of this paper, we first introduce some definitions.
Define an operator $\bar{\mathcal{G}}$, which can be written as
\begin{equation}
    \bar{\mathcal{G}}(Q_i(s,a)) = \frac{1}{N}\sum_{i=1}^{N} \mathbb{E}\left[r_i(s,a)\right] + \gamma \sum_{s^{\prime} \in \mathcal{S}} p_{s,s'}^a \max_{a' \in \mathcal{A}_i}Q_i(s^{\prime},a^{\prime}) 
\end{equation}

For this operator $\bar{\mathcal{G}}$, it is a contraction in each agent, which satisfies
\begin{equation}\label{p1}
    \left\|\bar{\mathcal{G}}({Q})-\bar{\mathcal{G}}(Q^{\prime})\right\|_{\infty} \leq \gamma \left\|Q-Q^{\prime}\right\|_{\infty}, \forall Q, Q^{\prime}.
\end{equation}
This means that there exists a unique fixed point of $Q^{*}$ of $\bar{\mathcal{G}}$, satisfying the $\bar{\mathcal{G}}(Q^{*}) = Q^{*}$.

Noting that $1_N^T L_t=\mathbf{0}$, for each state-action pair $(s,a)$, the average state-action value can be written as 
\begin{equation}
\begin{split}
    & \bar{Q}^{t+1}(s,a) = (1-\alpha_t(s,a))\bar{Q}^{t}(s,a) \\ 
    &~~ + \beta_t(s,a)(1-\kappa)\left(\bar{Q}^{t}_{\sharp}(s,a)-\bar{Q}^{t}(s,a)\right) \\ 
    &~~ + \alpha_t(s,a) \left(\bar{\mathcal{G}}(\bar{Q}^t(s,a))+\bar{v}^t(s,a)+\bar{\varepsilon}^t(s,a)\right),
\end{split}
\end{equation}
where $\bar{Q}^{t}_{\sharp}(s,a)=(1/N)1_N^T\bm{q}^t_{\sharp}(s,a)$, $\bar{v}^t(s,a) = (1/N) \mathbf{1}_N^T\bm{v}^t(s,a)$ and 
\begin{equation}
    \bar{\varepsilon}^t(s,a) = \frac{1}{N} \sum_{i=1}^{N}\left(\mathcal{G}(Q^t_i(s,a))-\mathcal{G}(\bar{Q}^t(s,a))\right).
\end{equation}

According to \eqref{p1}, there exists a constant $c_{11}$, such that
\begin{equation}
    \left|\bar{\varepsilon}^t(s,a)\right| \leq c_{11} \sum_{i=1}^N \left\|Q^t_i-\bar{Q}^t\right\|
\end{equation}
for all $t$. By Theorem~\ref{Theorem_1_1}, we have $\bar{\varepsilon}^t(s,a) \rightarrow 0$ as $t \rightarrow \infty$. 

Define an auxiliary process $\left\{z^t(s,a)\right\}$ for each state-action pair $(s,a)$, such that for all $t$,
\begin{equation}
    z^{t+1}(s,a) = (1-\alpha_t(s,a))z^t(s,a)+\alpha_t(s,a)(\bar{v}^t(s,a)+\bar{\varepsilon}^t(s,a)).
\end{equation}
% Based on the properties of $\left\{\bar{v}^t(s,a)\right\}$ and $\left\{\bar{\varepsilon}^t(s,a)\right\}$, i.e.,
Since $\mathbb{E}\left[\bar{v}^t(s,a)|\mathcal{F}_t\right]=0$ and $\mathbb{P}(\sup _{t \geq 0} \mathbb{E}[(\overline{v}^t(s, a))^{2} \mid \mathcal{F}_{t}]<\infty)=1$, we have $z^{t}(s,a) \rightarrow 0$ as $t \rightarrow \infty$ by Lemma \ref{L9}.
Due to the fact that the process $\bar{Q}^t$ is bounded and hence there exists an finite random variable $R$, such that 
\begin{equation}
    R = \lim \sup_{t \rightarrow \infty}\left\|\bar{Q}^t-Q^{*}\right\|_{\infty}.
\end{equation}
Assuming that there always exists an event $\mathcal{B}$ of positive measure such that $R>0$, in order to give a counter example,
we consider a process $\{\hat{Q}^t(s,a)\}$, for each state-action pair $(s,a)$, such that $\hat{Q}^t(s,a)=\bar{Q}^t(s,a)-z^t(s,a)-Q^{*}(s,a)$ for all $t$, which evolves as 
\begin{align}\label{T3_100}
    & \hat{Q}^{t+1}(s,a) = (1-\alpha_t(s,a))\hat{Q}^t(s,a) \nonumber \\
    & ~~~ + \beta_t(s,a)(1-\kappa)\left(\bar{Q}^{t}_{\sharp}(s,a)-\bar{Q}^{t}(s,a)\right) \nonumber \\
    & ~~~ + \alpha_t(s,a) \left(\bar{\mathcal{G}}(\bar{Q}(s,a))-\bar{\mathcal{G}}(Q^{*}(s,a))\right) \nonumber\\
    &\overset{(a)}{\leq} (1-\alpha_t(s,a))\hat{Q}^t(s,a) +\alpha_t(s,a) \gamma \left\|\bar{Q}^{t}-Q^{*}\right\|_{\infty}   \nonumber\\ 
    & ~~~+\beta_t(s,a)(1-\kappa) \left(\bar{Q}^{t}_{\sharp}(s,a)-\bar{Q}^{t}(s,a)\right)   \nonumber \\
    & \overset{(b)}{\leq} (1-\alpha_t(s,a)-\beta_t(s,a)(1-\kappa))\hat{Q}^t(s,a)  \nonumber \\
    & ~~~ +\alpha_t(s,a) \gamma \left\|\bar{Q}^{t}-Q^{*}\right\|_{\infty} \nonumber \\ 
    & ~~~ + \beta_t(s,a)(1-\kappa) \left(\bar{Q}^{t}_{\sharp}(s,a)- {Q}^{*}(s,a)-z^{t}(s,a)\right),
\end{align}
where $(a)$ is based on \eqref{p1} and $(b)$ is obtained by re-arranging the terms.

Since that $\lim_{t \rightarrow \infty}z^t(s,a) = 0$, there exists $t'$ such that $z^t(s,a) \leq R'$ for $t>t^{\prime}$. Therefore, \eqref{T3_100} can be written as
\begin{equation}
\begin{split}
    & ~~ \hat{Q}^{t+1}(s,a) \leq (1-\alpha_t(s,a)-\beta_t(s,a)(1-\kappa))\hat{Q}^t(s,a) \\ 
    & ~~~~~ +\alpha_t(s,a) \gamma \left\|\bar{Q}^{t}-Q^{*}\right\|_{\infty} \\
    &  + \beta_t(s,a)(1-\kappa) \left(\bar{Q}^{t}_{\sharp}(s,a)- {Q}^{*}(s,a)\right) + \beta_t(s,a)(1-\kappa) R' \\
    & \overset{(a)}{\leq} (1-\alpha_t(s,a)-\beta_t(s,a)(1-\kappa))\hat{Q}^t(s,a) \\
    & ~~~~~ +\alpha_t(s,a) \gamma \left\|\bar{Q}^{t}-Q^{*}\right\|_{\infty} \\
    & ~ + \beta_t(s,a)(1-\kappa) \left\|\bar{Q}^{t}-Q^{*}\right\|_{\infty} + \beta_t(s,a)(1-\kappa) R',
\end{split}
\end{equation}
for $t>t^{\prime}$, where $(a)$ is based on \eqref{p1}.
Let $\delta$ be a positive constant and such that $\gamma(1+\delta)<1$, we have
\begin{equation}
    \left\|\bar{Q}^{t}-Q^{*}\right\|_{\infty} \leq R(1+\delta).
\end{equation}
Then,
\begin{equation}
\begin{split}
    &~\hat{Q}^{t+1}(s,a) \leq \left(1-\alpha_t(s,a)-\beta_t(s,a)(1-\kappa)\right)\hat{Q}^t(s,a) \\ &+ \left(\alpha_t(s,a)+\beta_t(s,a) \frac{1-\kappa}{\gamma}+\beta_t(s,a) \frac{(1-\kappa)R'}{\gamma(1+\delta)R}\right)\gamma(1+\delta)R.
\end{split}
\end{equation}

\begin{lemma}\label{L6}
Let $\left\{z_t\right\}$ be real-valued and deterministic with
\begin{equation}
    z_{t+1} \leq (1-\alpha_t)z_t + (\alpha_t+\beta_t) \varepsilon_t,
\end{equation}
where the deterministic sequence $\{\alpha_t\}$, $\{\beta_t\}$ and $\{\varepsilon_t\}$ satisfy $\alpha_t, \beta_t \in \left[0,1\right]$, for all t, $\sum_{t\geq 0}\alpha_t \leq \infty$, $\beta_t \rightarrow 0$ as $t \rightarrow \infty$ and there exists a constant $R>0$, such that,
\begin{equation}
    \lim\sup_{t \rightarrow \infty} \varepsilon_t \leq R.
\end{equation}
Then we have $\lim \sup_{t\rightarrow \infty}z_t \leq R$.
\end{lemma}

This iterative process falls down the purview of Lemma \ref{L6}, which yields
\begin{equation}
    \mathbb{P}\big(\lim \sup_{t \rightarrow \infty} \big|\hat{Q}^t(s,a)\leq \gamma(1+\delta)R\big|\big) \geq \mathbb{P}(\mathcal{B}) > 0.
\end{equation}
Since, the above holds for each state-action pair $(s,a)$ and $\gamma(1+\delta)<1$, we conclude that
\begin{equation}
    \lim \sup_{t \rightarrow \infty} \left\|\bar{Q}^{t}-Q^{*}\right\|_{\infty} < R,
\end{equation}
on the event $\mathcal{B}$. Since $\mathcal{B}$ has positive measure, this contradicts the above hypothesis, hence, $R=0$, therefore, we complete the proof.
\end{proof}

\section{Proof of Lemmas}
\subsection{Proof of Lemma~\ref{def3}}

\begin{proof}
Due to the fact $\mathcal{P}(s^{t+1}=s^{\prime}|\mathcal{F}_t) = p_{s,s^{\prime}}^a$, we have $$\mathbb{E}_{s^{\prime}}[\max_{a^{\prime} \in \mathcal{A}_i}Q_i(s^{\prime},a^{\prime})|\mathcal{F}_t]=\sum_{s^{\prime}}p_{s,s^{\prime}}^a\max_{a^{\prime} \in \mathcal{A}_i}Q_i(s^{\prime}, a^{\prime}).$$
For each state-action pair, its $Q$-value is update by temporal difference in $\eqref{eq_q_learning}$. Then the expectation of the iteration process for any $s^{\sharp}$ can be written as
\begin{equation*}
\begin{split}
    & \mathbb{E}[Q^{t+1}_i(s^{\sharp},a)|\mathcal{F}_{t+1}] = (1-\alpha) \mathbb{E}[Q_i^t(s^{\sharp},a)|\mathcal{F}_t] \\
    &~~~~~ + \alpha \Big[\mathbb{E}[r_i(s^{\sharp},a)]+\gamma \sum_{s^{\prime}}p_{s^{\sharp},s^{\prime}}^a\max_{a' \in \mathcal{A}_i}Q_i^t(s^{\prime}, a^{\prime})\Big].
\end{split}
\end{equation*}
We found that the term $\mathbb{E}[r_i(s^{\sharp},a)]+\gamma \sum_{s^{\prime}}p_{s^{\sharp},s^{\prime}}^a\max_{a' \in \mathcal{A}_i}Q_i^t(s^{\prime}, a^{\prime})$ is not conditioned on $\mathcal{F}_t$.
Since $Q_i^0$ is randomly initialized, we have $\mathbb{E}\left[Q_i(s^{\sharp},a)|\mathcal{F}_t\right] = \mathbb{E}\left[Q_i(s^{\sharp},a)|s^{\sharp},a\right]$.
Then we complete the proof.
\end{proof}

\subsection{Proof of Lemma~\ref{L3}}
\begin{proof}
First, we review some useful properties of Laplacian matrix $L_t$ for each $\mathbf{y} \in \mathbb{R}^N$:
\begin{equation}
\begin{split}
\|\big(I_N-\beta_t L_t-\alpha_t I_N\big)\mathbf{y}\| & \leq (1-c \zeta_t)\|\mathbf{y}\|  + \alpha_t \|\mathbf{y}\| \\ 
& \leq (1-c'\zeta_t) \|\mathbf{y}\|,
\label{L_laplacian_condition}
\end{split}
\end{equation}
where the first inequality is based on Lemma~\ref{L7}, $\mathbb{E}[\zeta_t|\mathcal{F}]\geq \frac{c_{\zeta}}{(t+1)^{\tau_2}}$, and $c, c'$ are positive constants.

Define a process $\left\{V^t\right\}$, such that $V^t = \left\|\mathbf{z}^t(s,a)\right\|^2$, we have
\begin{equation}\label{Lemma7_1}
\begin{split}
    \mathbb{E}\left[V^{t+1}|\mathcal{F}\right] &\overset{(a)}{=} \mathbb{E}\left\|\big(I_N-\beta_t L_t-\alpha_t I_N\big)\mathbf{z}^t(s,a)\right\|^2 + \tilde{\beta}_t^2\left\|\bar{\Delta}_t\right\|^2 \\ 
    &+2\mathbb{E}\left[\big(I_N-\beta_t L_t-\alpha_t I_N\big)\tilde{\beta}_t\mathbf{z}^t(s,a) \bar{\Delta}_t^{T}\right] \\ 
    &~~~~~~~~~~~~~~~~~~~~~~ + \alpha_t^2\mathbb{E}\left[\left\|\bar{\bm{v}}^t(s,a)\right\|^2\right].
\end{split}
\end{equation}
For the third term in \eqref{Lemma7_1}, we have
\begin{equation}
\begin{split}
    &~~~~~2\mathbb{E}\left[\big(I_N-\beta_t L_t-\alpha_t I_N\big)\tilde{\beta}_t\mathbf{z}^t(s,a) \Delta_t^{T}\right]  \\
    & \overset{(a)}{\leq} 2 \mathbb{E}\left[\|\big(I_N-\beta_t L_t-\alpha_t I_N\big)\tilde{\beta}_t^{\frac{1}{2}}\mathbf{z}^t(s,a)\|\cdot\|\tilde{\beta}_t^{\frac{1}{2}}\bar{\Delta}_t^{T}\| \right]  \\
    & \leq 2 \mathbb{E}\left[\|\big(1-c_{12}\zeta_t\big)\tilde{\beta}_t^{\frac{1}{2}} \mathbf{z}^t(s,a)\|\cdot\|\tilde{\beta}_t^{\frac{1}{2}}\bar{\Delta}_t^{T}\| \right] \\
    & \leq \mathbb{E}\left[\big(1-c_{12}\zeta_t\big)\tilde{\beta}_t\|\mathbf{z}^t(s,a)\|^2+\tilde{\beta}_t\|\bar{\Delta}_t^{T}\|^2 \right],
\end{split}
\end{equation}
where $(a)$ is based on Cauchy-Schwarz inequality and  $\mathbb{E}[\zeta_t|\mathcal{F}]\geq \frac{c_{\zeta}}{(t+1)^{\tau_2}}$.

Similarly, 
\begin{equation*}
\begin{split}
    \mathbb{E}\left\|\big(I_N-\beta_t L_t-\alpha_t I_N\big)\mathbf{z}^t(s,a)\right\|^2 & \leq \mathbb{E}(1-c_{13}\zeta_t)^2\|\mathbf{z}^t(s,a)\|^2 \\ 
    & \leq \mathbb{E}(1-c'_{13}\zeta_t)\|\mathbf{z}^t(s,a)\|^2.
\end{split}
\end{equation*}
Therefore, there exists $c_{14}$, such that 
\begin{equation}
    \mathbb{E}\left[V^{t+1}|\mathcal{F}\right] \leq \big(1-c_{14}\zeta_t\big)\mathbb{E}\left[V^{t}|\mathcal{F}\right]+\alpha_t^2 c_2  + (\tilde{\beta}_t^2+\tilde{\beta}_t)\left\|\bar{\Delta}_t\right\|^2, 
\end{equation}
due to that $\mathbb{E}[\zeta_t|\mathcal{F}]\geq \frac{c_{\zeta}}{(t+1)^{\tau_2}} > \tilde{\beta}_t$, 
for any $\varepsilon$, there exists a random time $t_{\varepsilon}$, such that $V^{t} \leq \varepsilon$.
Then we complete the proof.
\end{proof}

\subsection{Proof of Lemma~\ref{L4}}
\begin{proof}
Note that, for each $t \geq t_0$,
\begin{equation}
\begin{split}
    \left\|\mathbf{z}^{t:t_0}(s,a)\right\| & \leq \Big\|\mathbf{z}^{t:0}(s,a)-\Big(\prod_{t^{\prime}=t_0}^{t-1}\big(I_N-\beta_{t'}(s,a)L_{t'} \\ 
    &~~~~~~~~~~~~~~~-\alpha_{t'}(s,a)I_N\big)\Big)\mathbf{z}^{t_0,0}(s,a)\Big\| \\
    &\leq \left\|\mathbf{z}^{t:0}(s,a)\right\| + \left\|\mathbf{z}^{t_0,0}(s,a)\right\|,
\end{split}
\end{equation}
where the last inequality using the fact that 
\begin{equation}
    \left\|I_N-\beta_t(s,a)L_t-\alpha_t(s,a)I_N\right\| \leq 1, \forall t \geq 0.
\end{equation}
By Lemma 3, there exists a constant $\varepsilon$ makes that 
\begin{equation}
    \left\|\mathbf{z}^{t:0}(s,a)\right\| \leq \varepsilon/2, 
\end{equation}
as $t \rightarrow \infty$. Hence, There exists $t_{\varepsilon}$, such that $\left\|\mathbf{z}^{t:t_0}(s,a)\right\| \leq \varepsilon, \forall t \geq t_{\varepsilon}$. Then we complete the proof.
\end{proof}

\subsection{Proof of Lemma~\ref{L8}}

\begin{proof}
To proof Lemma~\ref{L8}, for any scalar $a$, we first define its truncation $(a)_C$ at level $C > 0$ by
\begin{equation}
    (a)_{C}= \begin{cases}\frac{a}{|a|} \min (|a|, C) & \text { if } a \neq 0 \\ 0 & \text { if } a=0\end{cases}.
\end{equation}
Then, according to the equation A.13 in \cite{kar2013distributed}, we can obtain the truncation sequence $\left\{\hat{z}_C(t)\right\}$ evolving as
\begin{equation}\label{L8_1}
    \hat{z}_C(t+1) \leq (1-r_1(t))\hat{z}_C(t) + r_1(t) \hat{\Delta}_C(t) + \tilde{r}_2(t), 
\end{equation}
where 
\begin{equation}
    \tilde{r}_2(t) \leq \frac{k_1}{(t+1)^{\delta_2-\delta_1-\frac{1}{2+\varepsilon_1}}},
\end{equation}
for some constant $k_1>0$. Note that $\delta_1 > \delta_2-\delta_1-\frac{1}{2+\varepsilon_1}$ and $\Delta_t$ is a bounded process, hence there exists another constant $k_2$ such that
$$r_1(t) \hat{\Delta}_C(t) + \tilde{r}_2(t) \leq \hat{r}_2(t) \leq \frac{k_2}{(t+1)^{\delta_2-\delta_1-\frac{1}{2+\varepsilon_1}}}.$$
Then \eqref{L8_1} can be rewritten as
\begin{equation}
    \hat{z}_C(t+1) \leq (1-r_1(t))\hat{z}_C(t) + \hat{r}_2(t), 
\end{equation}
the remaining portion of this proof is same as the process in \cite{kar2013distributed}.
\end{proof}

\subsection{Proof of Lemma~\ref{L6}}
\begin{proof}
Consider $\delta > 0$ and note that, there exists $t_{\delta} > 0$, such that $\varepsilon_t \leq R+\delta$ and $\beta_t < 0 + \delta$ for all $t \geq t_{\delta}$. Hence, for $t \geq t_{\delta}$, we have 
\begin{equation}
    z_{t+1} \leq (1-\alpha_t)z_t + (\alpha_t+\beta_t) \left(R+\delta \right).
\end{equation}
Let $\hat{z}_t = z_t-(R+\delta)$ for all $t$, we have, for $t \geq t_{\delta}$,
\begin{equation}
\begin{split}
    \hat{z}_{t+1} \leq (1-\alpha_t)\hat{z}_{t} + \delta\left(R+\delta\right).
\end{split}
\end{equation}
When invoking the above equation recursively, we have
\begin{equation}
    \hat{z}_{t+1} \leq \prod_{j=t_{\delta}}^t(1-\alpha_j) \hat{z}_{t_{\delta}}+\sum_{i=t_{\delta}}^t\delta \prod_{j=i}^t(1-\alpha_j)(R+\delta).
\end{equation}
Thus we have $\sum_{t\geq t_{\delta}}\alpha_t = \infty$, then we conclude that 
\begin{equation}
    \lim \sup_{t\rightarrow \infty} \prod_{j=t_{\delta}}^t(1-\alpha_j) \leq \lim \sup_{t \rightarrow \infty} \exp \left( {-\sum_{j=t_{\delta}}^t\alpha_j} \right)=0.
\end{equation}
Hence, we have
\begin{equation}
    \hat{z}_{t+1} \leq \sum_{i=t_{\delta}}^t\delta \prod_{j=i}^t(1-\alpha_j)(R+\delta),
\end{equation}
while 
\begin{equation}
    \lim \sup_{t \rightarrow \infty} z_t \leq R+\delta + \sum_{i=t_{\delta}}^{t-1}\delta \prod_{j=i}^{t-1}(1-\alpha_j)(R+\delta)
\end{equation}
from which the desired assertion follows by taking $\delta$ to zero. Then we finish the proof.
\end{proof}

\section{Conclusions}
In this paper, we propose a personalization approach in meta-RL to solve the gradient conflict problem, which learns a meta-policy and personalized policies for all tasks and specific tasks, respectively. By adopting a personalization constrain in the objective function, our algorithm encourages each task to pursue its personalized policy around the meta-policy under the tabular and deep network settings. We introduce an auxiliary policy to decouple the personalized and meta-policy learning process and propose an alternating minimization method for policy improvement.
Moreover, theoretical analysis shows that our algorithm converges linearly with the iteration number and gives an upper bound on the difference between the personalized policies and meta-policy. Experimental results demonstrate that pMeta-RL outperforms many advanced meta-RL algorithms on the continuous control tasks.


\bibliographystyle{IEEEtran}
\bibliography{cite}

\end{document}